\documentclass{article} 
\usepackage{iclr2027_conference,times}

\usepackage{amsmath,amsfonts,bm}

\def\eqref#1{equation~\ref{#1}}

\def\1{\bm{1}}

\DeclareMathAlphabet{\mathsfit}{\encodingdefault}{\sfdefault}{m}{sl}
\SetMathAlphabet{\mathsfit}{bold}{\encodingdefault}{\sfdefault}{bx}{n}

\usepackage{xcolor}
\usepackage{xspace}
\usepackage{booktabs}
\usepackage{graphicx,multirow,colortbl}
\usepackage[listings]{tcolorbox}

\usepackage{hyperref}
\usepackage{url}

\usepackage{microtype}
\usepackage{inconsolata}
\usepackage{graphicx}

\usepackage{ragged2e}
\usepackage{amsmath}
\usepackage{subcaption}
\usepackage{booktabs}
\usepackage{multirow}
\usepackage{tabularx}
\usepackage{makecell}
\usepackage{siunitx}
\usepackage{xcolor}
\usepackage{pifont} 

\usepackage{array}
\usepackage{amssymb}   
\usepackage{wrapfig}
\usepackage{float}
\usepackage{placeins}
\usepackage{needspace}

\usepackage{rotating}
\definecolor{rowgray}{gray}{0.95} 
\usepackage{pifont} 
\usepackage{colortbl} 

\usepackage{tcolorbox}
\tcbuselibrary{breakable}
\usepackage{xcolor}
\usepackage{enumitem}
\usepackage{listings}
\usepackage{newunicodechar}
\usepackage{fontawesome5} 
\tcbuselibrary{listings,skins}

\definecolor{textgray}{gray}{0.5}      
\newcolumntype{C}[1]{>{\centering\arraybackslash}p{#1}}
\usepackage{array}
\newcolumntype{M}{>{\centering\arraybackslash}p{2.6em}}

\title{MASBench: Benchmarking LLM-based\\
Multi-Agent Collaboration\\
under Partial Observability}
\iclrfinalcopy

\author{
\parbox{0.95\textwidth}{
\centering
\normalsize\normalfont
Qizhi Chu$^{1}$ \quad
Zekai Yu$^{1}$ \quad
Sijie Wen$^{1}$ \quad
Yang Liu$^{1}$\\[0.35em]
Chen Qian$^{2}$ \quad
Cheng Yang$^{1}$\thanks{Correspondence to: Cheng Yang
\textless\href{mailto:yangcheng@bupt.edu.cn}
{\nolinkurl{yangcheng@bupt.edu.cn}}\textgreater.} \quad
Chuan Shi$^{1}$ \quad
Zhiyuan Liu$^{3}$\\[0.8em]
{\small
$^{1}$Beijing University of Posts and Telecommunications\\[0.15em]
$^{2}$Shanghai Jiao Tong University
\quad
$^{3}$Tsinghua University
}
}
}
\newtcblisting{promptbox}{
  listing only,
  colback=gray!6,
  colframe=gray!40,
  boxrule=0.4pt,
  arc=0.8mm,
  left=8pt, right=8pt, top=8pt, bottom=8pt,
  listing options={
    basicstyle=\small\ttfamily,
    columns=fullflexible,
    keepspaces=true,
    showstringspaces=false,
  },
}
\begin{document}

\maketitle
\lhead{Preprint} 

\begin{abstract}
Large language models (LLMs) have progressively evolved into the core of autonomous agents. Building on this progress, LLM-based multi-agent systems (MAS) coordinate multiple agents into a synergistic team to accomplish complex tasks that exceed the capabilities of individual agents. The effectiveness of such systems depends not only on the agents themselves, but also on how collaboration mechanisms are designed and organized. Note that real-world collaboration is typically \textit{partially observable}, where each agent can only access partial information about the environment due to physical or privacy-related constraints. However, many existing multi-agent benchmarks assume global observability, and leave limited support for systematically evaluating collaboration mechanisms. 
To bridge this gap, we introduce \textbf{MASBench}, a multi-agent collaboration benchmark designed under \textit{partially observable} constraints. It is organized into three progressive task categories: \textit{Reasoning}, \textit{Scheduling}, and \textit{Game}. Through this structure, we progressively evaluate three representative collaboration mechanisms: \textit{Protocol}, \textit{Memory}, and \textit{Routing}. MASBench further provides deterministic evaluation metrics, including performance score, communication cost, and communication efficiency, to characterize both collaboration outcomes and communication overhead. Experiments across diverse LLM backbones and mechanism configurations offer empirical guidance for effective MAS design. Code is available at: \url{https://github.com/BUPT-GAMMA/MASBench}.

\end{abstract}

\section{Introduction}


Large language model-based (LLM-based) agents can decompose complex objectives into executable steps, invoke external tools, and perform iterative corrections based on environmental feedback, laying the foundation for complex planning and long-horizon task execution~\cite{yao2022react,schick2023toolformer,shinn2023reflexion}. Building upon this, multi-agent systems (MAS)~\cite{hong2023metagpt,li2023camel,xi2025rise} organize multiple autonomous agents via collaboration mechanisms~\cite{packer2023memgpt,liu2023dynamic,marro2024scalable,liang2024encouraging,shen2025understanding}, enabling them to work synergistically as a team. This paradigm significantly expands the boundaries of agent applications, handling intricate scenarios ranging from software development~\cite{hong2023metagpt,qian2024chatdev} to strategic simulations~\cite{meta2022human,park2023generative,xu2023exploring}. 



It is worth emphasizing that the performance gains in multi-agent systems do not merely stem from increasing the number of agents but fundamentally depend on the design and organization of collaboration mechanisms~\cite{li2024survey,gridach2025agentic}. Synthesizing existing literature and practice, here we focus on three representative mechanisms: \textit{memory} improves long-term collaboration by recording, summarizing, and maintaining key information across multi-turn interactions; \textit{routing}
facilitates efficient information exchange by determining the communication topology; and \textit{protocol} organizes communication through message schemas and interaction rules.

As shown in Figure~\ref{fig:observability}, real-world group collaboration often operates under strict constraints of being \textit{partially observable}~\cite{bernstein2002complexity,stanton2006distributed,butko2008pomdp,Oliehoek2016ACI}, where the participants lack direct access to the global state. However, many existing benchmarks for evaluating LLM-based agents assume largely global observability in

\begingroup
\setlength{\intextsep}{4pt}
\setlength{\columnsep}{10pt}

\begin{wrapfigure}{r}{0.44\textwidth}
    \centering
    \includegraphics[width=\linewidth]{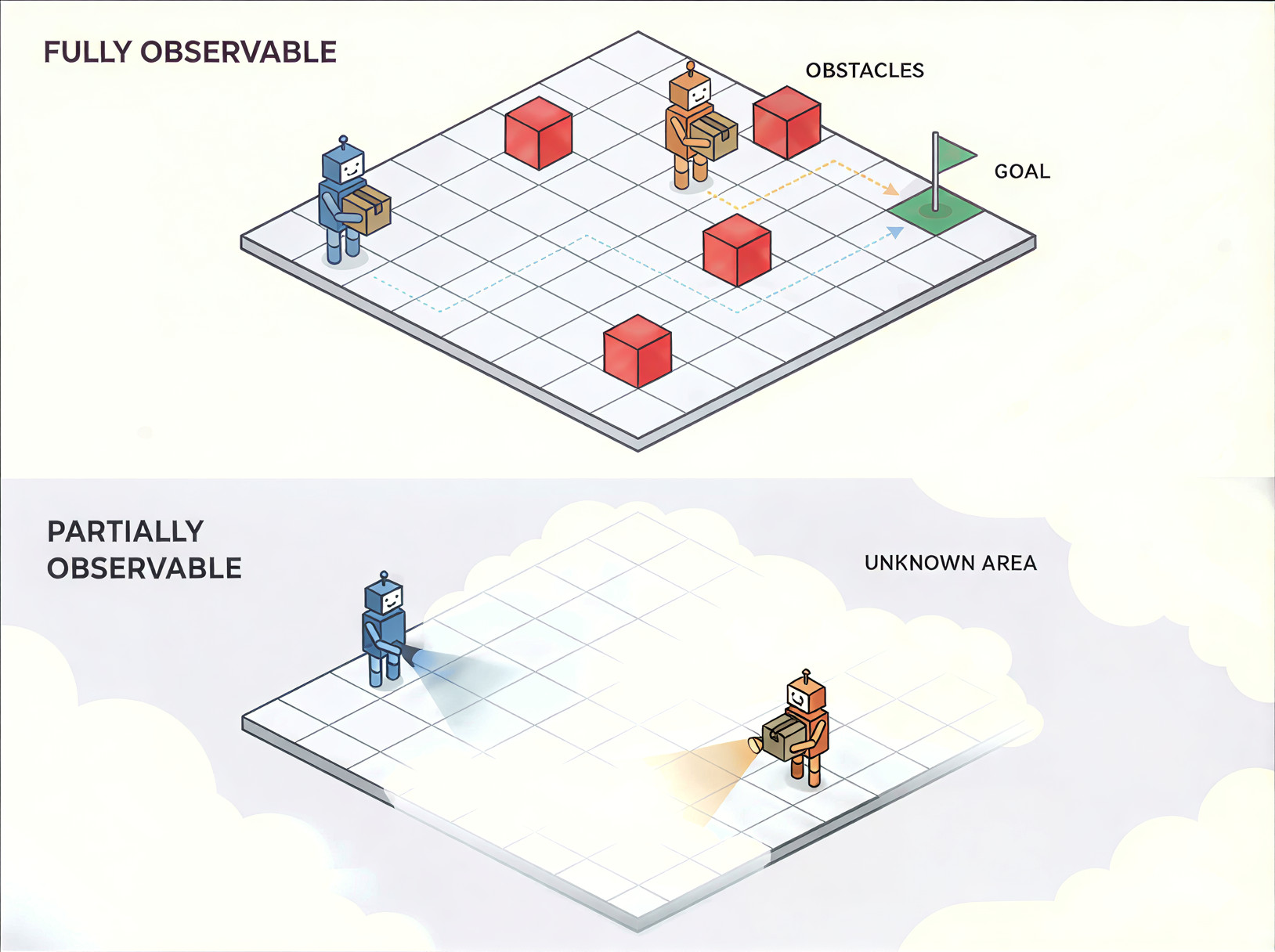}
    \vspace{-0.4\baselineskip}
    \captionsetup{font=small,skip=2pt}
    \caption{Fully versus partially observable collaboration scenarios.}
    \label{fig:observability}
\end{wrapfigure}

practice~\cite{liu2023agentbench,mialon2023gaia,qin2023toolllm,sun2025collab}, thereby trivializing the challenges faced in realistic collaboration. Although recent works~\cite{chang2024agentboard,zhu2025multiagentbench,ruan2025benchmarking} have attempted to introduce partially observable settings, their primary focus is often end-to-end task performance rather than mechanism-level analysis. Moreover, many benchmarks rely on LLM-as-a-judge~\cite{zhu2023judgelm} for at least part of their assessment, due to the absence of clear quantitative metrics, or concentrate on domain-specific scenarios. The LLM-based evaluation has been shown to be highly vulnerable to systematic biases (\textit{e.g.}, verbosity bias) and evaluator versions~\cite{shi2025judging}. Meanwhile, domain-specific benchmarks~\cite{gong2024mindagent,li2025legalagentbench,fan2025ai} tend to prioritize evaluating an individual agent's adaptability to a particular environment, rather than assessing the collaboration mechanisms.

\par
\endgroup



To bridge this gap, we propose \textbf{MASBench}, a multi-agent collaboration benchmark under \textit{partially observable constraints}. MASBench is organized into three progressive task categories based on mechanism requirements: 

\textbf{Reasoning (Focus on Protocol):} We distribute the evidence chain for a given problem between two agents, ensuring that each receives only partial information and is unable to independently complete the reasoning. Agents must share key clues and reach a consensus through multi-turn dialogue.

\textbf{Scheduling (Focus on Memory \& Protocol):} This category simulates privacy-restricted meeting scheduling. Each agent can only access its own schedule and is prohibited from publicly disclosing the complete calendar. Agents must discuss candidate time slots and verify feasibility via negotiation.

\textbf{Game (Focus on Memory, Routing \& Protocol):} This category describes a cooperative grid-world task, where multiple agents are required to collect and transport different types of resources under local observation. 

This classification is not merely a stacking of difficulty but implies a targeted verification of specific collaboration mechanisms through scenario construction. We also design three evaluation metrics, including performance score, communication cost and communication efficiency, to quantitatively characterize the collaboration. The main contributions are summarized as follows:


$\bullet$ We propose \textbf{MASBench}, a multi-agent collaboration benchmark under \textit{partially observable} constraints. The three task categories (\textit{i.e.,} Reasoning, Scheduling, and Game) progressively evaluate key collaboration mechanisms of LLM-based agents.
    
    
$\bullet$ Additionally, the collaboration process can be quantified without LLM-as-a-judge based on the proposed performance score, communication cost and communication efficiency metrics, ensuring reproducibility and fairness.
    
$\bullet$ We conduct extensive experiments across diverse combinations of LLM backbones and collaboration mechanisms. The results provide empirical guidance for designing more efficient and effective multi-agent systems in the future. 

\section{Related Work}

\subsection{LLM-Based MAS}
LLM-based multi-agent systems have demonstrated remarkable effectiveness in tackling complex tasks~\cite{li2023camel,li2024survey,guo2024large}. Prior studies suggest MAS effectiveness is strongly influenced by \emph{collaboration mechanisms}: \textit{Memory} maintains key state across turns~\cite{packer2023memgpt,park2023generative,wang2023voyager,zhong2024memorybank}. For example, Generative Agents~\citep{park2023generative} utilizes memory streams to maintain behavioral consistency. \textit{Routing} shapes information flow by deciding \emph{who communicates with whom}~\cite{liu2023dynamic,wu2024autogen,zhuge2024gptswarm,shen2025understanding}. For example, DyLAN~\citep{liu2023dynamic} dynamically optimizes interaction graphs based on context. \textit{Protocol} standardizes interaction via structured stages and message schemas~\cite{song2023restgpt,IOA,marro2024scalable,chen2025optima}. For example, OPTIMA explicitly optimizes inter-agent communication by training on trajectories that jointly reward task performance and token efficiency~\citep{chen2025optima}. 
Beyond these mechanisms, the community has explored decision-optimization strategies, such as multi-agent debate~\citep{liang2024encouraging}, reflexion~\citep{shinn2023reflexion}, and voting~\citep{wang2023self} to enhance decision quality.


\subsection{Benchmarks for LLM-based MAS}

Prior studies often adopt general-purpose benchmarks, emphasizing outcome-oriented task completion under relatively loose interaction conditions (\textit{e.g.,} GAIA, AgentBench)~\citep{liu2023agentbench,mialon2023gaia}. 
With the rise of LLM-based multi-agent systems, benchmarks have expanded toward collaborative task solving~\citep{zhu2025multiagentbench,sun2025collab} and simulation~\citep{zhou2023sotopia}. Recent benchmarks further introduce partially observable settings~\citep{zhou2023sotopia,yu2024mineland,zhu2025multiagentbench,ruan2025benchmarking}.

Despite these advances, existing benchmarks are primarily designed for end-to-end agent capability evaluation, rather than controlled analysis of how different collaboration mechanisms contribute under partial observability. 
Moreover, many multi-agent benchmarks rely on LLM-as-a-judge for at least part of their assessment, which is prompt-sensitive and vulnerable to systematic biases, undermining reproducibility~\citep{zheng2023judging, shi2025judging}.
By contrast, MASBench focuses on mechanism-level evaluation: it organizes tasks to progressively stress protocol, memory, and routing, and reports deterministic task-grounded performance, communication cost and communication efficiency.

Besides, there are also domain-specific multi-agent benchmarks that simulate structured role interactions in realistic workflows~\citep{dong2024villageragent,li2025legalagentbench,fan2025ai,long2024teamcraft,gong2024mindagent}. These benchmarks are complementary to MASBench, but they typically prioritize environment-specific adaptation of individual agents rather than assessing collaboration mechanisms.

\section{Preliminary}
\label{sec:preliminary}

\subsection{Problem Formulation}
\label{sec:problem}



In this paper, we formalize the multi-agent system as a tuple
$\langle \mathcal{G}, \mathcal{A}, \mathcal{E}\rangle$, where $\mathcal{G}$
is the overall goal, $\mathcal{A}$ is a set of LLM-based agents, and
$\mathcal{E}$ is the shared environment. Now we describe the process of a
multi-round collaboration. Specifically, each agent $\mathcal{A}_i \in
\mathcal{A}$ is defined as a tuple
$\langle \mathcal{O}_i,m_i,\mathcal{M}_i,\mathcal{D}_i,\mathcal{C}_i\rangle$:

\begin{itemize}[
    leftmargin=1.5em,
    labelsep=0.45em,
    itemsep=0pt,
    parsep=0pt,
    topsep=2pt,
    partopsep=0pt
]
    \item $\mathcal{O}_i$ is the observation function that perceives the current environment and receives messages sent to $\mathcal{A}_i$ in the last round;
    \item $m_i$ is the memory of agent $\mathcal{A}_i$;
    \item $\mathcal{M}_i$ is the memory management function that updates memory based on the current observation as well as $\mathcal{A}_i$'s memory and actions from last round;
    \item $\mathcal{D}_i$ is the decision function that outputs the action interacting with the environment based on the current memory and observations;
    \item $\mathcal{C}_i$ is the communication function that outputs the action interacting with other agents based on the current memory and observations.
\end{itemize}

The global state of environment $\mathcal{E}$ will transit given the actions
of all agents. Note that, under the partial observability setting, each agent
has a different observation function that can only perceive a local view of
the global state and receive the messages sent to it. Finally, the system-level
output is usually derived from the last global state to match the overall goal
$\mathcal{G}$.

\subsection{Key Collaboration Mechanisms}
Synthesizing existing literature and practice~\cite{packer2023memgpt,shen2025understanding,liu2023dynamic,marro2024scalable,IOA}, we focus on three key mechanisms that regulate the collaboration process: 

\textbf{Memory.} The memory mechanism supports the consistency of long-term collaboration. By dynamically recording and summarizing past observations and interactions, memory maintains a historical representation to assist an agent's decision making.




\textbf{Routing.} The routing mechanism determines the communication topology, namely \emph{who communicates with whom and when}. 
By dynamically selecting the most suitable agent to communicate with, routing can facilitate more effective and timely information exchange among the agents.

\textbf{Protocol.} The protocol mechanism regulates how agent communication and interaction are organized. By specifying message schemas and interaction rules, protocols can help reduce interaction ambiguity and promote the stability of the collaboration process.

Routing and protocol primarily regulate the communication function, while task-specific interaction rules may also constrain or inform the decision function.

\begin{figure*}[!t] 
    \centering
    \includegraphics[width=1\linewidth]{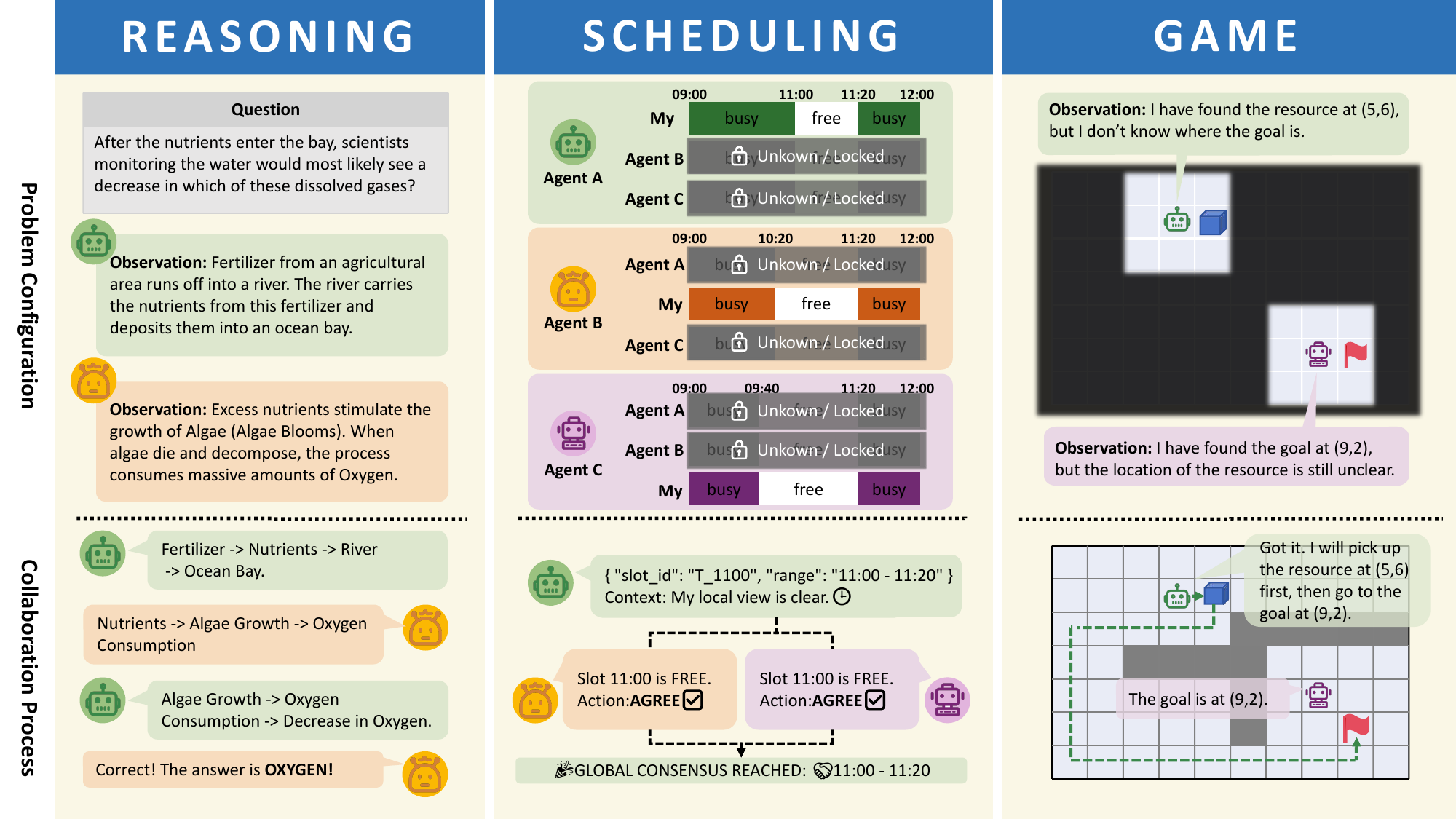} 
    
    \caption{\textbf{Illustration of the three task categories in MASBench: Reasoning (left), Scheduling (middle) and Game (right).} }
    \label{fig:benchmark_overview} 
    \vspace{-10pt}
\end{figure*}

\section{MASBench}

\subsection{Benchmark Overview}
\label{sec:benchmark_overview}

As shown in Figure~\ref{fig:benchmark_overview}, MASBench provides a comprehensive evaluation framework for multi-agent collaboration under partial observability. 
We adopt a mechanism-driven perspective to evaluate how agents leverage memory, protocols, and routing to bridge the information gaps from partial observability. MASBench comprises three tasks (reasoning, scheduling, and game), designed to progressively increase collaboration mechanism demands. 

\subsection{Benchmark Design}
The statistics of the three tasks are shown in Table~\ref{tab:statistics}. Each of the tasks targets on the evaluation of specific mechanisms.

\textbf{Reasoning.}
Reasoning provides the most compact collaboration setting, where two agents are required to answer a question based on the distinct portions of information assigned to each. 
Specifically, We adapt multi-hop instances from HotpotQA~\cite{yang2018hotpotqa} to form a collaborative inference task under partial observability. For each problem, the evidence chain is decomposed and distributed across two agents, so that no single agent has sufficient information to solve the task independently. Agents must exchange key clues and reach a final answer through multi-turn dialogue. We split instances into \textit{Simple} and \textit{Complex} subsets based on supporting-fact count, context length, and answer length. The detailed information are provided in Appendix~\ref{app:reasoning_details}.


\begingroup
\setlength{\intextsep}{3pt}
\setlength{\columnsep}{8pt}

\begin{wraptable}[9]{r}{0.5\textwidth}
    \vspace{-0.5\baselineskip}
    \centering
    \captionsetup{skip=2pt}
    \caption{Benchmark statistics.}
    \label{tab:statistics}

    \footnotesize
    \setlength{\tabcolsep}{1.5pt}
    \renewcommand{\arraystretch}{1.05}

    \begin{tabular}{@{}lccMMM@{}}
        \toprule
        \multirow{2}{*}{\textbf{Category}} &
        \multirow{2}{*}{\textbf{Agents}} &
        \multirow{2}{*}{\textbf{Instances}} &
        \multicolumn{3}{c}{\textbf{Mechanisms}} \\
        \cmidrule(lr){4-6}
        & & &
        \textbf{Prot.} & \textbf{Mem.} & \textbf{Route.} \\
        \midrule
        Reasoning  & 2    & 300 (200/100) & \checkmark & & \\
        Scheduling & 5    & 200 (100/100) & \checkmark & \checkmark & \\
        Game       & 4--8 & 100 (50/50)   & \checkmark & \checkmark & \checkmark \\
        \bottomrule
    \end{tabular}

    \vspace{-0.3\baselineskip}
\end{wraptable}
\textbf{Scheduling.} 
Scheduling task simulates a privacy-constrained meeting coordination scenario, where multiple agents need to negotiate a meeting time feasible for everyone. Compared with Reasoning, each agent faces longer private inputs and more complex multi-round negotiation, making memory needed to be considered in this task. Since communication follows a broadcast-style chat, routing is not included.

Specifically, Calendars are randomly generated with 10-minute granularity, and the group must agree on a continuous 20-minute meeting window. We quantify difficulty by the number of feasible 20-minute windows $N_{\mathrm{solution}}$: \textit{Simple} instances have $N_{\mathrm{solution}} \in \{4,5\}$, while \textit{Complex} instances have $N_{\mathrm{solution}} \in \{2,3\}$. The detailed calendar generation procedure is provided in Appendix~\ref{app:scheduling_details}.

\textbf{Game.} Game poses the richest collaboration demands in the benchmark as a cooperative Dec-POMDP-style grid-world task that requires spatio-temporal coordination under partial observability. Multiple agents must explore the map, collect resources, and deposit them into warehouses to earn scores within limited rounds. Since each agent only observes a local window and communication is point-to-point, all three mechanisms (\textit{i.e.,} memory, protocol and routing) are crucial.

The environment contains terrain cells, resources, and warehouses with resource-specific quotas. Each agent observes a limited local window centered on itself. At each step, an agent takes an action to interact with the environment, including movement and item operations. Notably, each agent can carry only one type of resource at a time, subject to a maximum capacity. After that, agents may send point-to-point messages to interact with other agents. The messages will be received in the beginning of next round. Detailed action rules and scoring constraints are provided in Appendix~\ref{app:game_details}. Problem complexity is primarily controlled by map scale: \textit{Simple} instances use $15\times15$ maps, while \textit{Complex} instances use $18\times18$ maps with more resources and higher coordination demands.

\subsection{Evaluation Metrics}
\label{sec:metrics}
We evaluate the tasks with the following three metrics for task performance and communication cost.

\textbf{1) Performance Score.} 
We report the \textbf{average normalized score} on a $[0,100]$ scale across all test instances. The scoring criteria for each task are defined as follows:

\begin{itemize}[leftmargin=1.0em, labelsep=0.35em, itemsep=1pt, topsep=2pt, parsep=0pt, partopsep=0pt]
    \item \textbf{Reasoning:} We compute word-level F1 between predicted and ground-truth answers for each instance, and report the mean F1 scaled by $100$.
    
    \item \textbf{Scheduling:} We use the success rate, where an instance is counted as successful only if the agents output a meeting window feasible for \emph{all} agents. The rate is scaled by $100$.
    
    \item \textbf{Game:} We normalize the final collected-resource score by the map-specific reference score and scale it by $100$, and report the average normalized game score. Note that the complex subset does not necessarily indicate a lower performance score, since the number of agents and available resources also increases.
\end{itemize}


\textbf{2) Communication Cost.} 
We measure the communication cost as the total tokens generated for \emph{inter-agent messages}. Note that the token costs for the local reasoning of each agent are excluded.




\textbf{3) Communication Efficiency.} Combining the two metrics above, we define the communication efficiency as the performance score achieved per $1{,}000$ tokens of communication cost.


\section{Experiments}
\subsection{Experimental Settings}

\begin{table*}[!t]
\centering
\vspace{-0.15\baselineskip}
\caption{\textbf{Experimental results on Reasoning.} Here Score$\uparrow$, Tok$\downarrow$ and Eff$\uparrow$ denote the performance score, communication cost, and communication efficiency defined in Sec.~\ref{sec:metrics}. \textbf{Bold} denotes the best results per backbone.}
\label{tab:reasoning_results}
\vspace{-0.05\baselineskip}
\setlength{\tabcolsep}{5pt}
\renewcommand{\arraystretch}{1.07}
\resizebox{0.85\textwidth}{!}{
    \begin{tabular}{l l
        S[table-format=2.2] S[table-format=5.0] S[table-format=3.2]
        S[table-format=2.2] S[table-format=5.0] S[table-format=3.2]
        S[table-format=2.2] S[table-format=5.0] S[table-format=3.2]}
    \toprule
    \multirow{2}{*}{Model} & \multirow{2}{*}{Setting} &
    \multicolumn{3}{c}{Simple} &
    \multicolumn{3}{c}{Complex} &
    \multicolumn{3}{c}{Overall} \\
    \cmidrule(lr){3-5}\cmidrule(lr){6-8}\cmidrule(lr){9-11}
    & & {Score$\uparrow$} & {Tok$\downarrow$} & {Eff$\uparrow$} &
    {Score$\uparrow$} & {Tok$\downarrow$} & {Eff$\uparrow$} &
    {Score$\uparrow$} & {Tok$\downarrow$} & {Eff$\uparrow$} \\
    \midrule
    
    \multirow{2}{*}{Llama-3.1-8B-Instruct}
    & Base         & {29.25} & {1,224} & {23.89} & {19.69} & {1,334} & {14.76} & {26.06} & {1,261} & {20.67} \\
    & Protocol   & {\textbf{36.54}} & {\textbf{529}} & {\textbf{69.11}} & {\textbf{27.28}} & {\textbf{533}} & {\textbf{51.16}} & {\textbf{33.45}} & {\textbf{530}} & {\textbf{63.09}} \\
    \midrule
    
    \multirow{2}{*}{GPT-5-nano}
    & Base         & {45.14} & {447} & {100.97} & {39.81} & {807} & {49.35} & {43.36} & {567} & {76.48} \\
    & Protocol   & {\textbf{51.65}} & {\textbf{150}} & {\textbf{344.52}} & {\textbf{45.68}} & {\textbf{166}} & {\textbf{275.56}} & {\textbf{49.66}} & {\textbf{155}} & {\textbf{319.97}} \\
    \midrule
    
    \multirow{2}{*}{Gemini-2.5-Flash-Lite}
    & Base         & {\textbf{59.47}} & {533} & {111.50} & {33.82} & {702} & {48.21} & {50.92} & {589} & {86.39} \\
    & Protocol   & {59.03} & {\textbf{278}} & {\textbf{212.27}} & {\textbf{48.54}} & {\textbf{417}} & {\textbf{116.45}} & {\textbf{55.53}} & {\textbf{324}} & {\textbf{171.21}} \\
    \midrule
    
    \multirow{2}{*}{Deepseek-v3.2}
    & Base         & {58.20} & {678} & {85.90} & {46.83} & {792} & {59.16} & {54.41} & {716} & {76.04} \\
    & Protocol   & {\textbf{58.36}} & {\textbf{198}} & {\textbf{295.42}} & {\textbf{48.97}} & {\textbf{229}} & {\textbf{214.10}} & {\textbf{55.23}} & {\textbf{208}} & {\textbf{265.61}} \\
    \midrule
    
    \multirow{2}{*}{Qwen-Plus}
    & Base         & {59.39} & {353} & {168.36} & {48.64} & {442} & {110.06} & {55.81} & {382} & {145.91} \\
    & Protocol   & {\textbf{61.37}} & {\textbf{149}} & {\textbf{411.51}} & {\textbf{54.98}} & {\textbf{185}} & {\textbf{297.11}} & {\textbf{59.24}} & {\textbf{161}} & {\textbf{367.71}} \\

    \bottomrule
    \end{tabular}
}
\vspace{-0.05\baselineskip}
\end{table*}

\begin{table*}[!t]
\centering
\caption{\textbf{Experimental results on Scheduling.} Score$\uparrow$, Tok$\downarrow$ and Eff$\uparrow$ denote the performance score, communication cost, and communication efficiency from Sec.~\ref{sec:metrics}. \textbf{Bold} and \underline{underline} denote the best and second-best results per backbone.}
\label{tab:scheduling_results}

\vspace{-0.05\baselineskip}

\setlength{\tabcolsep}{5pt}
\renewcommand{\arraystretch}{1.07}
\resizebox{0.85\textwidth}{!}{
    \begin{tabular}{l l
        S[table-format=2.2] S[table-format=5.0] S[table-format=3.2]
        S[table-format=2.2] S[table-format=5.0] S[table-format=3.2]
        S[table-format=2.2] S[table-format=5.0] S[table-format=3.2]}
    \toprule
    \multirow{2}{*}{Model} & \multirow{2}{*}{Setting} &
    \multicolumn{3}{c}{Simple} &
    \multicolumn{3}{c}{Complex} &
    \multicolumn{3}{c}{Overall} \\
    \cmidrule(lr){3-5}\cmidrule(lr){6-8}\cmidrule(lr){9-11}
        & &
        {Score$\uparrow$} & {Tok$\downarrow$} & {Eff$\uparrow$} &
        {Score$\uparrow$} & {Tok$\downarrow$} & {Eff$\uparrow$} &
        {Score$\uparrow$} & {Tok$\downarrow$} & {Eff$\uparrow$} \\
    \midrule
    
\multirow{3}{*}{GPT-5-nano}
    & Base                 & {\underline{1}} & {\textbf{1,131}} & {\underline{0.88}} & {1} & {\textbf{1,101}} & {0.91} & {1} & {\textbf{1,116}} & {0.90} \\
    & Memory             & {\underline{1}} & {\underline{1,842}} & {0.54} & {\underline{3}} & {\underline{1,683}} & {\underline{1.78}} & {\underline{2}} & {\underline{1,763}} & {\underline{1.13}} \\
    & Memory + Protocol  & {\textbf{52}} & {2,089} & {\textbf{24.89}} & {\textbf{32}} & {3,794} & {\textbf{8.43}} & {\textbf{42}} & {2,942} & {\textbf{14.28}} \\
    \midrule
    
    \multirow{3}{*}{Gemini-2.5-Flash-Lite}
    & Base          & {19} & {\underline{996}} & {19.07} & {15} & {\textbf{1,001}} & {14.98} & {17} & {\underline{999}} & {17.02} \\
    & Memory         & {\underline{40}} & {1,816} & {\underline{22.03}} & {\underline{33}} & {1,795} & {\underline{18.39}} & {\underline{36.5}} & {1,805} & {\underline{20.22}} \\
    & Memory + Protocol  & {\textbf{78}} & {\textbf{847}} & {\textbf{92.07}} & {\textbf{47}} & {\underline{1,019}} & {\textbf{46.12}} & {\textbf{62.5}} & {\textbf{933}} & {\textbf{66.98}} \\
    \midrule
    
    \multirow{3}{*}{Deepseek-v3.2}
    & Base          & {29} & {\textbf{388}} & {74.66} & {16} & {\textbf{448}} & {35.74} & {22.5} & {\textbf{418}} & {53.82} \\
    & Memory         & {\underline{45}} & {\underline{507}} & {\textbf{88.77}} & {\underline{31}} & {\underline{638}} & {\underline{48.61}} & {\underline{38}} & {\underline{572}} & {\underline{66.39}} \\
    & Memory + Protocol  & {\textbf{74}} & {840} & {\underline{88.10}} & {\textbf{52}} & {969} & {\textbf{53.64}} & {\textbf{63}} & {905} & {\textbf{69.64}} \\
    \midrule
    
    \multirow{3}{*}{Qwen-Plus}
    & Base          & {16} & {\textbf{506}} & {31.65} & {11} & {\textbf{525}} & {20.94} & {13.5} & {\textbf{515}} & {26.19} \\
    & Memory         & {\underline{51}} & {1,063} & {\underline{47.96}} & {\underline{29}} & {1,273} & {\underline{22.77}} & {\underline{40}} & {1,168} & {\underline{34.23}} \\
    & Memory + Protocol  & {\textbf{69}} & {\underline{993}} & {\textbf{69.48}} & {\textbf{58}} & {\underline{1,095}} & {\textbf{52.97}} & {\textbf{63.5}} & {\underline{1,044}} & {\textbf{60.82}} \\
    
    \bottomrule
    \end{tabular}
}
\vspace{-0.1\baselineskip}
\end{table*}
\paragraph{LLM Backbones.}
We conduct evaluations on five LLM backbones covering diverse capability--cost regimes: Llama-3.1-8B~\cite{dubey2024llama}, GPT-5-nano~\cite{openai_gpt5nano_docs}, Gemini-2.5-Flash-Lite~\cite{comanici2025gemini}, DeepSeek-v3.2~\cite{liu2025deepseek}, and Qwen-Plus~\cite{alibaba_modelstudio_models}. Due to the higher long-horizon planning and action-validity requirements in \textsc{Scheduling} and \textsc{Game}, Llama-3.1-8B exhibits a high failure rate in these two categories. Thus, we include Llama-3.1-8B only in the \textsc{Reasoning} category.

\paragraph{Mechanism Configurations.}
Our benchmark focuses on three representative coordination mechanisms in LLM-based MAS: \textsc{Memory}, \textsc{Protocol}, and \textsc{Routing}. For each task category, we evaluate only the subset of mechanisms that is operationally meaningful given the specific interaction structure and constraints. Concrete operationalizations are task-dependent and specified accordingly for each task category below, with implementation details provided in Appendix~\ref{app:mechanisms}.

\textbf{Reasoning.}
\textsc{Base} uses free-form dialogue, while \textsc{Protocol} structures the interaction using typed messages for information requests, evidence sharing, and final answers.

\textbf{Scheduling.} 
\textsc{Base} uses unconstrained natural-language negotiation based on their private schedule and the recent dialogue context. \textsc{Memory} maintains structured availability and negotiation states, including candidate and rejected meeting slots across rounds. \textsc{Protocol} further organizes the interaction into a multi-phase protocol (\textit{e.g.}, coarse search $\rightarrow$ candidate proposal $\rightarrow$ voting phase).

\textbf{Game.} 
\textsc{Base} disables inter-agent communication, yielding a baseline driven solely by local observations. \textsc{Memory} stores discovered world states from local observations and received messages. \textsc{Routing} enables point-to-point communication with two recipient-selection modes: \textsc{Random} samples a teammate, while \textsc{Gate} first decides whether communication is needed and then selects the recipient. \textsc{Protocol} further structures transmitted messages using grounded task-relevant fields, including agent state, observations, and current plans.


\subsection{Main Results}

\begin{table}[!ht]
\centering
\caption{\textbf{Experimental results on Game.} Score$\uparrow$, Tok$\downarrow$ and Eff$\uparrow$ denote performance score, communication cost, and communication efficiency from Sec.~\ref{sec:metrics}. Results are averaged over five runs per map and configuration. \textbf{Bold} and \underline{underline} denote the best and second-best results per backbone.}
\label{tab:game_results}

\resizebox{\textwidth}{!}{%
    \begin{tabular}{l l
                    S[table-format=3.1] S[table-format=6.0] S[table-format=3.2]
                    S[table-format=3.1] S[table-format=6.0] S[table-format=3.2]
                    S[table-format=3.1] S[table-format=6.0] S[table-format=3.2]}
    \toprule
    \multirow{2}{*}{Model} & \multirow{2}{*}{Setting} &
    \multicolumn{3}{c}{Simple} &
    \multicolumn{3}{c}{Complex} &
    \multicolumn{3}{c}{Overall} \\
    \cmidrule(lr){3-5}\cmidrule(lr){6-8}\cmidrule(lr){9-11}
    & &
    {Score$\uparrow$} & {Tok$\downarrow$} & {Eff$\uparrow$} &
    {Score$\uparrow$} & {Tok$\downarrow$} & {Eff$\uparrow$} &
    {Score$\uparrow$} & {Tok$\downarrow$} & {Eff$\uparrow$} \\
    \midrule

     \multirow{6}{*}{GPT-5-nano} 
     & Base & {37.27} & \multicolumn{1}{c}{---} & \multicolumn{1}{c}{---} & {47.95} & \multicolumn{1}{c}{---} & \multicolumn{1}{c}{---} & {42.61} & \multicolumn{1}{c}{---} & \multicolumn{1}{c}{---} \\
     & Memory & {46.73} & \multicolumn{1}{c}{---} & \multicolumn{1}{c}{---} & {54.57} & \multicolumn{1}{c}{---} & \multicolumn{1}{c}{---} & {50.65} & \multicolumn{1}{c}{---} & \multicolumn{1}{c}{---} \\
     & Memory + Random & {49.36} & {5,161} & {9.56} & {58.35} & {6,633} & {8.80} & {53.85} & {5,897} & {9.13} \\
     & Memory + Gate & {\underline{55.72}} & {3,531} & {15.78} & {\underline{62.23}} & {4,302} & {14.47} & {\underline{58.98}} & {3,916} & {15.06} \\
     & Memory + Random + Protocol & {53.53} & {\underline{2,917}} & {\underline{18.35}} & {59.29} & {\underline{3,916}} & {\underline{15.14}} & {56.41} & {\underline{3,417}} & {\underline{16.51}} \\
     & Memory + Gate + Protocol & {\textbf{58.20}} & {\textbf{1,851}} & {\textbf{31.44}} & {\textbf{63.21}} & {\textbf{2,359}} & {\textbf{26.79}} & {\textbf{60.71}} & {\textbf{2,105}} & {\textbf{28.84}} \\
    \midrule

     \multirow{6}{*}{Gemini-2.5-Flash-Lite} 
     & Base & {23.57} & \multicolumn{1}{c}{---} & \multicolumn{1}{c}{---} & {24.14} & \multicolumn{1}{c}{---} & \multicolumn{1}{c}{---} & {23.85} & \multicolumn{1}{c}{---} & \multicolumn{1}{c}{---} \\
     & Memory & {24.89} & \multicolumn{1}{c}{---} & \multicolumn{1}{c}{---} & {26.61} & \multicolumn{1}{c}{---} & \multicolumn{1}{c}{---} & {25.75} & \multicolumn{1}{c}{---} & \multicolumn{1}{c}{---} \\
     & Memory + Random & {24.82} & {2,652} & {9.36} & {27.77} & {3,510} & {7.91} & {26.29} & {3,081} & {8.53} \\
     & Memory + Gate & {\underline{26.39}} & {\underline{1,667}} & {\underline{15.83}} & {\underline{30.90}} & {\underline{2,302}} & {\underline{13.42}} & {\underline{28.64}} & {\underline{1,984}} & {\underline{14.43}} \\
     & Memory + Random + Protocol & {26.09} & {2,364} & {11.04} & {28.87} & {3,002} & {9.62} & {27.48} & {2,683} & {10.24} \\
     & Memory + Gate + Protocol & {\textbf{28.87}} & {\textbf{1,437}} & {\textbf{20.09}} & {\textbf{31.95}} & {\textbf{1,746}} & {\textbf{18.30}} & {\textbf{30.41}} & {\textbf{1,592}} & {\textbf{19.11}} \\
    \midrule

     \multirow{6}{*}{Deepseek-v3.2} 
     & Base & {17.91} & \multicolumn{1}{c}{---} & \multicolumn{1}{c}{---} & {19.26} & \multicolumn{1}{c}{---} & \multicolumn{1}{c}{---} & {18.59} & \multicolumn{1}{c}{---} & \multicolumn{1}{c}{---} \\
     & Memory & {44.69} & \multicolumn{1}{c}{---} & \multicolumn{1}{c}{---} & {55.56} & \multicolumn{1}{c}{---} & \multicolumn{1}{c}{---} & {50.12} & \multicolumn{1}{c}{---} & \multicolumn{1}{c}{---} \\
     & Memory + Random & {51.93} & {3,182} & {16.32} & {56.53} & {3,981} & {14.20} & {54.23} & {3,581} & {15.14} \\
     & Memory + Gate & {\underline{56.44}} & {\underline{2,211}} & {\underline{25.52}} & {\underline{61.47}} & {\underline{2,843}} & {\underline{21.62}} & {\underline{58.95}} & {\underline{2,527}} & {\underline{23.33}} \\
     & Memory + Random + Protocol & {54.49} & {2,953} & {18.45} & {58.32} & {3,670} & {15.89} & {56.40} & {3,312} & {17.03} \\
     & Memory + Gate + Protocol & {\textbf{57.67}} & {\textbf{1,687}} & {\textbf{34.19}} & {\textbf{62.47}} & {\textbf{2,027}} & {\textbf{30.82}} & {\textbf{60.07}} & {\textbf{1,857}} & {\textbf{32.35}} \\
    \midrule

     \multirow{6}{*}{Qwen-Plus} 
     & Base & {34.47} & \multicolumn{1}{c}{---} & \multicolumn{1}{c}{---} & {43.00} & \multicolumn{1}{c}{---} & \multicolumn{1}{c}{---} & {38.73} & \multicolumn{1}{c}{---} & \multicolumn{1}{c}{---} \\
     & Memory & {49.05} & \multicolumn{1}{c}{---} & \multicolumn{1}{c}{---} & {57.73} & \multicolumn{1}{c}{---} & \multicolumn{1}{c}{---} & {53.39} & \multicolumn{1}{c}{---} & \multicolumn{1}{c}{---} \\
     & Memory + Random & {49.37} & {4,353} & {11.34} & {57.03} & {5,501} & {10.37} & {53.20} & {4,927} & {10.80} \\
     & Memory + Gate & {\underline{56.52}} & {\underline{3,196}} & {\underline{17.69}} & {\underline{63.01}} & {\underline{3,900}} & {\underline{16.16}} & {\underline{59.77}} & {\underline{3,548}} & {\underline{16.85}} \\
     & Memory + Random + Protocol & {56.45} & {3,335} & {16.93} & {58.41} & {4,240} & {13.78} & {57.43} & {3,787} & {15.16} \\
     & Memory + Gate + Protocol & {\textbf{59.26}} & {\textbf{1,937}} & {\textbf{30.60}} & {\textbf{64.85}} & {\textbf{2,398}} & {\textbf{27.04}} & {\textbf{62.05}} & {\textbf{2,167}} & {\textbf{28.63}} \\

    \bottomrule
    \end{tabular}%
} 

\end{table}

Tables~\ref{tab:reasoning_results}, \ref{tab:scheduling_results}, and \ref{tab:game_results} present the main results; we highlight the main empirical trends below.

\textbf{Strong backbone capability is decisive for task utility.} 
Across all three task categories, Qwen-Plus achieves the highest \emph{Score}, outperforming the other backbones. This performance gap indicates that backbone capability remains a decisive factor for reaching higher task utility under partial observability, even when agents are equipped with collaboration mechanisms. Importantly, these results also demonstrate that \textbf{MASBench} can serve as a meaningful evaluation bedrock for \emph{backbone models} themselves, rather than merely comparing coordination mechanisms.

However, superior raw performance does not necessarily translate to better cost-effectiveness: although Qwen-Plus often leads on \emph{Score}, it is frequently surpassed by other backbones on \emph{Eff}, suggesting that higher utility may come with disproportionate communication overhead.

\textbf{Mechanism gains are consistent across backbones and tasks.} 
Across backbones, enabling task-relevant mechanisms generally improves collaboration effectiveness, indicating that the observed gains are not tied to a particular model.

In \textbf{Reasoning}, \textsc{Protocol} improves coordination efficiency across backbones: \emph{Score} increases modestly, ranging from $+1.5\%$ to $+28.4\%$, while \emph{Tok} drops substantially, by $45.0\%$ to $72.7\%$, resulting in a $2.0\times$ to $4.2\times$ gain in \emph{Eff}.

In \textbf{Scheduling}, \textsc{Memory} improves \emph{Score} over \textsc{Base}, with a cross-round ablation (Appendix~\ref{app:cross_round_memory}) showing gains from state retention even without availability preprocessing. \textsc{Protocol} further improves success rate and efficiency.

In \textbf{Game}, \textsc{Memory} improves long-horizon coordination over \textsc{Base} by retaining discovered states. With routing enabled, \textsc{Gate} outperforms \textsc{Random}, improving \emph{Score} by 2.35 to 6.57 points and reducing \emph{Tok} by $28.0\%$ to $35.6\%$ across backbones. \textsc{Protocol} further improves both routing modes, yielding a $1.12\times$--$1.91\times$ gain in \emph{Eff} over routing-only variants. Yet, \textsc{Memory+Random+Protocol} still underperforms \textsc{Memory+Gate} in \emph{Score}, highlighting the importance of routing quality.

\subsection{Framework-Level Evaluation}
\label{sec:framework_eval}

\begin{table}[!ht]
\centering
\vspace{-0.15\baselineskip}
\caption{\textbf{Framework-level evaluation on a shared 30\% subset.}
Score$\uparrow$ and Tok$\downarrow$ denote performance and mean communicated tokens per instance.
Reference follows the task-specific configurations in Sec.~\ref{sec:framework_eval}.
\textbf{Bold} marks the highest Score and lowest Tok per backbone and task.}
\label{tab:framework_results}
\vspace{-0.05\baselineskip}

\setlength{\tabcolsep}{5pt}
\renewcommand{\arraystretch}{1.07}
\sisetup{
    mode=text,
    detect-weight=true,
    detect-family=true,
    group-separator={,},
    group-minimum-digits=4
}

\resizebox{0.8\textwidth}{!}{%
\begin{tabular}{l l
    S[table-format=2.2] S[table-format=4.1]
    S[table-format=2.2] S[table-format=4.1]
    S[table-format=2.2] S[table-format=4.1]}
\toprule
\multirow{2}{*}{Model} & \multirow{2}{*}{System} &
\multicolumn{2}{c}{Reasoning} &
\multicolumn{2}{c}{Scheduling} &
\multicolumn{2}{c}{Game} \\
\cmidrule(lr){3-4}\cmidrule(lr){5-6}\cmidrule(lr){7-8}
& & {Score$\uparrow$} & {Tok$\downarrow$} &
    {Score$\uparrow$} & {Tok$\downarrow$} &
    {Score$\uparrow$} & {Tok$\downarrow$} \\
\midrule

\multirow{3}{*}{GPT-5-nano}
& CAMEL
& 38.15 & 255.5
& 6.67 & 688.2
& 46.47 & 4971.9 \\
& AgentVerse
& 43.31 & 353.4
& 16.67 & \bfseries 533.2
& 42.02 & 4023.8 \\
& Reference
& \bfseries 52.07 & \bfseries 185.4
& \bfseries 41.67 & 1150.2
& \bfseries 57.74 & \bfseries 2033.0 \\
\midrule

\multirow{3}{*}{Gemini-2.5-Flash-Lite}
& CAMEL
& 53.97 & 544.3
& 6.67 & 457.4
& 29.82 & 1527.0 \\
& AgentVerse
& 47.87 & 575.5
& 13.33 & \bfseries 291.8
& 18.97 & \bfseries 1430.5 \\
& Reference
& \bfseries 54.08 & \bfseries 374.5
& \bfseries 66.67 & 933.1
& \bfseries 32.25 & 1517.0 \\
\midrule

\multirow{3}{*}{DeepSeek-v3.2}
& CAMEL
& 57.92 & 543.1
& 40.00 & \bfseries 814.3
& 51.36 & 1987.9 \\
& AgentVerse
& 54.94 & 515.7
& 38.33 & 818.3
& 33.44 & \bfseries 1453.2 \\
& Reference
& \bfseries 61.11 & \bfseries 379.5
& \bfseries 55.00 & 904.7
& \bfseries 56.92 & 1890.9 \\
\midrule

\multirow{3}{*}{Qwen-Plus}
& CAMEL
& 57.16 & 482.4
& 21.67 & 654.2
& 26.11 & 2114.0 \\
& AgentVerse
& 58.59 & 407.7
& 23.33 & \bfseries 631.2
& 29.61 & \bfseries 2018.5 \\
& Reference
& \bfseries 60.46 & \bfseries 158.6
& \bfseries 58.33 & 1044.0
& \bfseries 57.32 & 2152.0 \\

\bottomrule
\end{tabular}%
}
\vspace{-0.5\baselineskip}
\end{table}

To assess the applicability of MASBench beyond mechanism-level analysis, we
evaluate CAMEL~\citep{li2023camel} and AgentVerse~\citep{chen2023agentverse}
across all three tasks and four LLM backbones.
All systems are evaluated on the same stratified 30\% subset under identical
task-specific observation constraints.
Our controlled scaffold serves as the reference system, using Protocol for
\textsc{Reasoning}, Memory+Protocol for \textsc{Scheduling}, and
Memory+Gate+Protocol for \textsc{Game}.
Further evaluation details are provided in Appendix~\ref{app:framework_setup}.

The results demonstrate that MASBench can serve as a common testbed for
evaluating heterogeneous multi-agent orchestration frameworks.
Performance differences become more pronounced as collaboration demands
increase: the gaps are generally modest in \textsc{Reasoning}, but substantially
larger in \textsc{Scheduling} and several \textsc{Game} settings.
This suggests that general-purpose orchestration does not automatically adapt
to the information asymmetry and coordination demands imposed by partial
observability.
The two frameworks also exhibit distinct task-dependent behaviors:
AgentVerse generally communicates more compactly and performs better in most
\textsc{Scheduling} settings, whereas CAMEL achieves stronger performance in
most \textsc{Game} settings.
These differences indicate that MASBench can distinguish how existing
frameworks manage distributed information and coordination across different
collaboration demands, complementing the controlled mechanism-level analysis.

\subsection{Further Analysis on Reasoning Category}
\label{sec:further-reason}

\begin{figure}[!h]
    \centering

    \begin{subfigure}[t]{0.55\linewidth}
        \centering
        \vspace{0pt}
        \includegraphics[height=3.25cm]{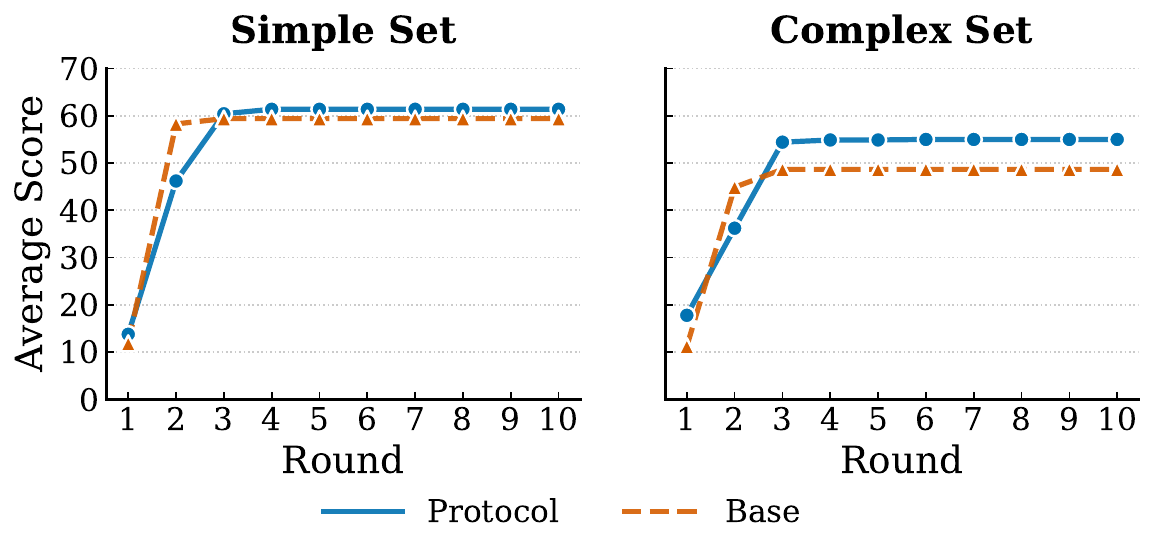}
        \caption{Round-wise score.}
        \label{fig:reasoning_roundcurve}
    \end{subfigure}
    \hfill
    \begin{subfigure}[t]{0.43\linewidth}
        \centering
        \vspace{0pt}
        \includegraphics[height=3.25cm]{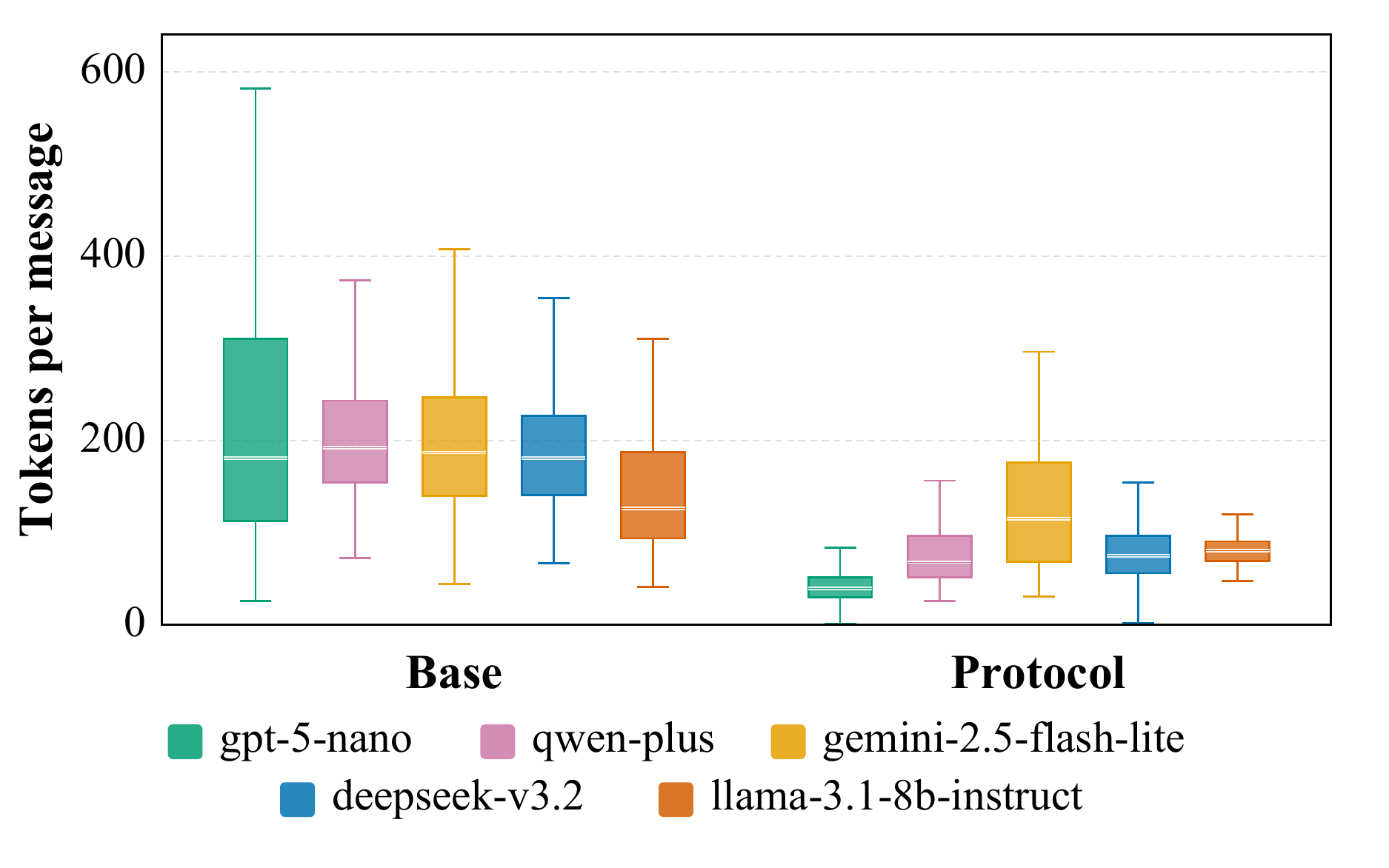}
        \caption{Message length distribution.}
        \label{fig:reasoning_msglen}
    \end{subfigure}

    \vspace{-1mm}
    \caption{\textbf{Further analysis on \textsc{Reasoning}.}
    (a) Round-wise score evolution on the Simple and Complex subsets.
    (b) Per-message token distributions across LLM backbones and settings.}
    \label{fig:reasoning_analysis}
\end{figure}

Figure~\ref{fig:reasoning_roundcurve} shows average score evolution over rounds. On the \textbf{Simple} split, both settings improve rapidly, while \textsc{Protocol} converges to a higher plateau. On the \textbf{Complex} split, the gap becomes clearer: \textsc{Protocol} attains a higher steady-state score, indicating that structured deliberation helps bridge information gaps. 

Figure~\ref{fig:reasoning_msglen} shows that, across backbones, protocolized communication yields shorter and more stable messages: the median message length significantly drops compared to \textsc{Base}. This suggests that \textsc{Protocol} reduces redundant wording by structuring clue requests and evidence sharing, which is crucial for reasoning under partial observability.


\subsection{Further Analysis on Scheduling Category}
\label{sec:further-schedule}

\begin{figure}[!ht]
    \centering

    \begin{subfigure}[t]{0.58\linewidth}
        \centering
        \vspace{0pt}
        \includegraphics[height=3.35cm]{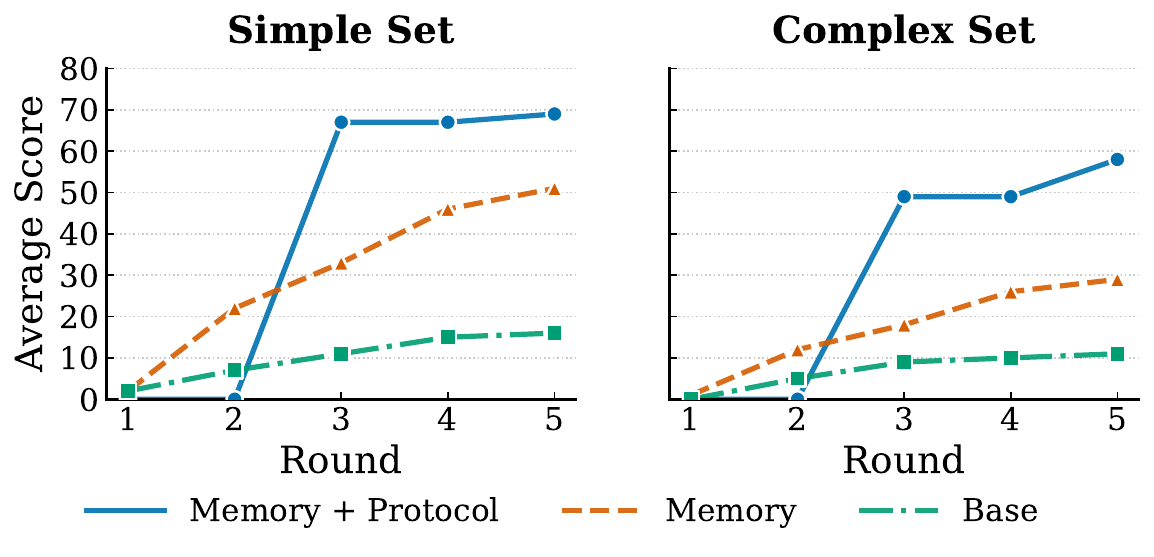}
        \caption{Round-wise score.}
        \label{fig:scheduling_roundcurve}
    \end{subfigure}
    \hfill
    \begin{subfigure}[t]{0.40\linewidth}
        \centering
        \vspace{0pt}
        \includegraphics[height=3.35cm]{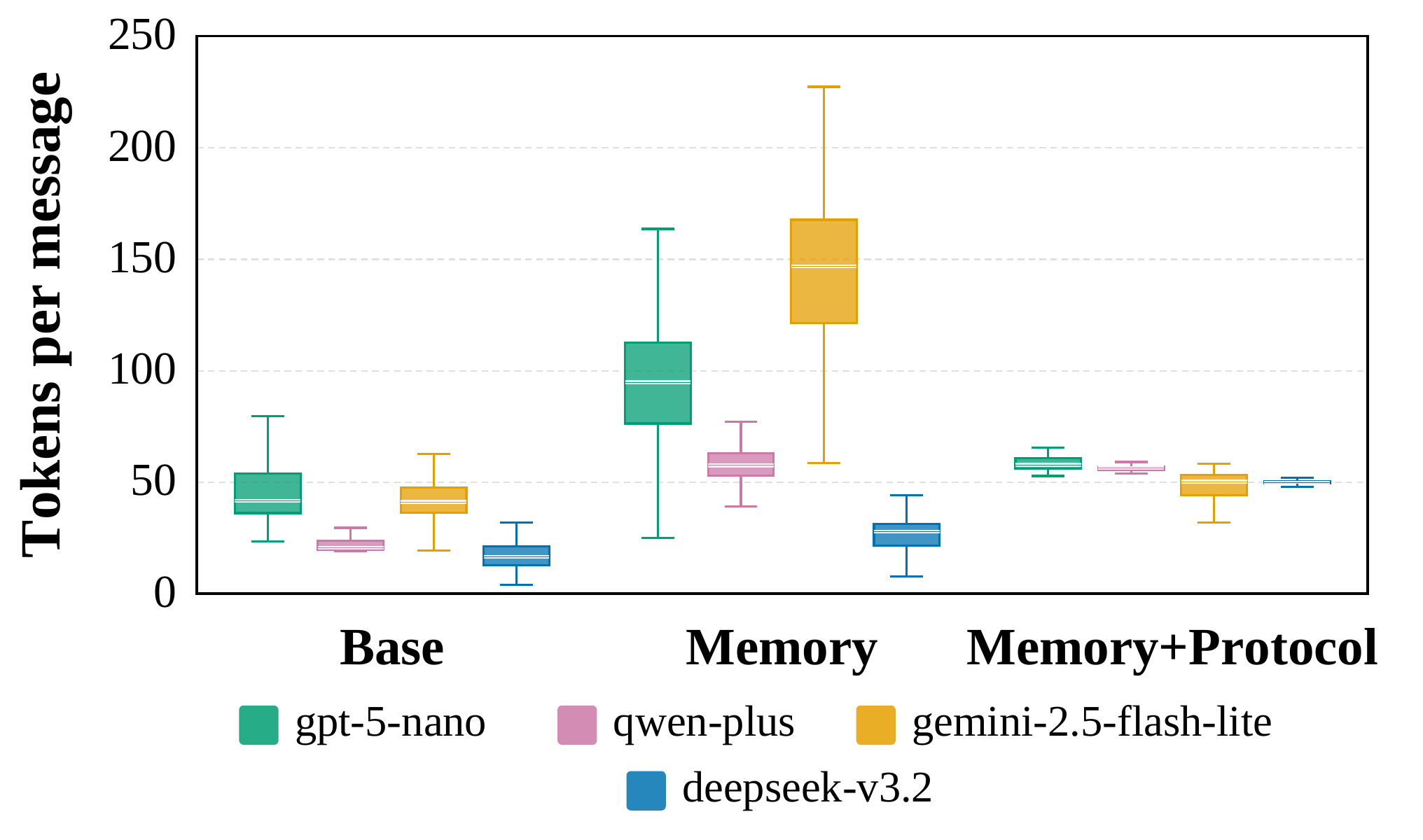}
        \caption{Message length distribution.}
        \label{fig:scheduling_msglen}
    \end{subfigure}

    \vspace{-1mm}
    \caption{\textbf{Further analysis on \textsc{Scheduling}.}
    (a) Round-wise score evolution on the Simple and Complex subsets.
    (b) Per-message token distributions across LLM backbones and settings.}
    \label{fig:scheduling_analysis}
\end{figure}

Figure~\ref{fig:scheduling_roundcurve} illustrates how performance score evolves over negotiation rounds. Compared with \textsc{Base},  \textsc{Memory} accelerates negotiation by grounding proposals in feasible intervals and avoiding rejected slots, while \textsc{Memory+Protocol} yields the fastest early improvement and the highest plateau on both Simple and Complex splits, indicating more efficient consensus.

As shown in Figure~\ref{fig:scheduling_msglen}, \textsc{Memory} does not consistently shorten messages because structured availability and shared state increase response length for some backbones. In contrast, \textsc{Memory+Protocol} constrains negotiation to typed actions (e.g., propose $\rightarrow$ vote $\rightarrow$ aggregate), reducing both message length and variance. Overall, \textsc{Memory} supports long-horizon constraint consistency, while \textsc{Protocol} turns that memory into efficient, convergent negotiation.

\vspace{-1mm}


\subsection{Further Analysis on Game Category}
\label{sec:further-game}

\begin{figure}[!ht]
    \centering

    \begin{subfigure}[t]{0.59\linewidth}
        \centering
        \vspace{0pt}
        \includegraphics[height=3.35cm]{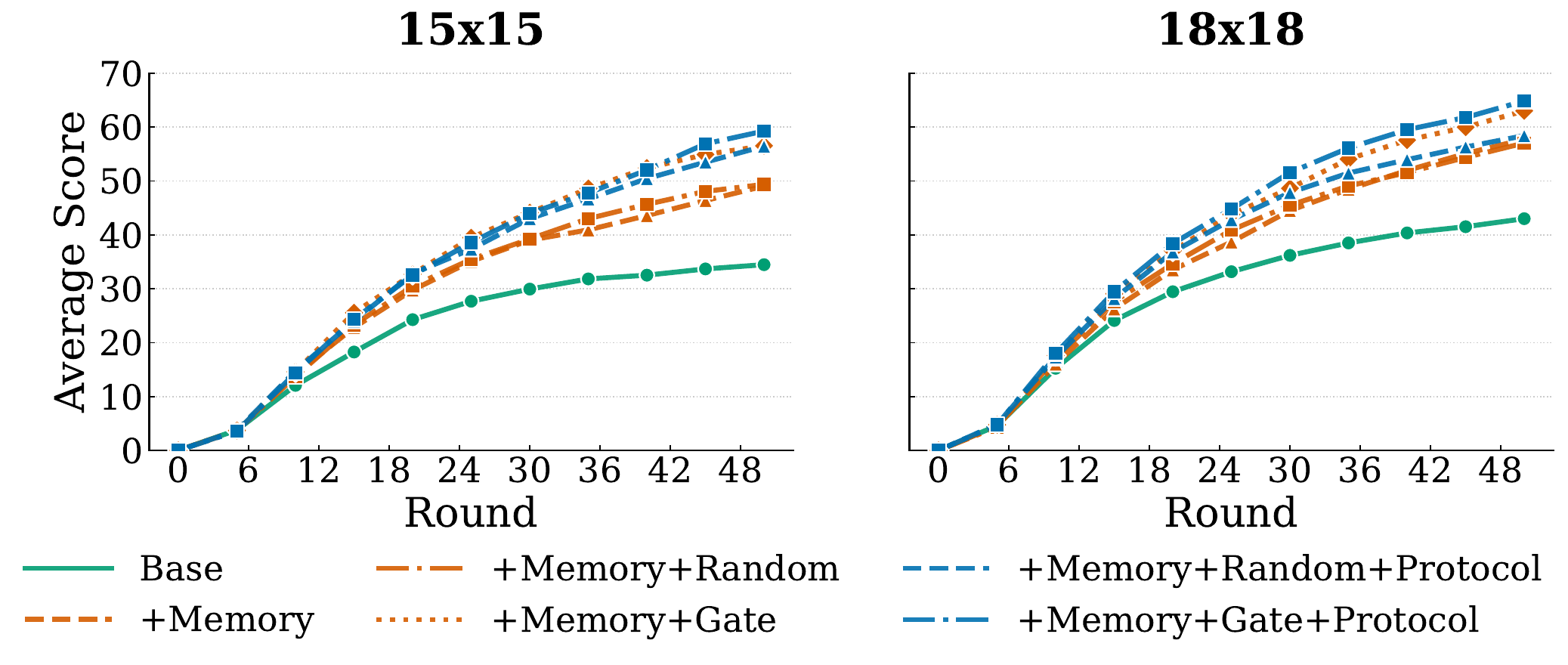}
        \caption{Round-wise score.}
        \label{fig:game_roundcurve}
    \end{subfigure}
    \hfill
    \begin{subfigure}[t]{0.39\linewidth}
        \centering
        \vspace{0pt}
        \includegraphics[height=3.35cm]{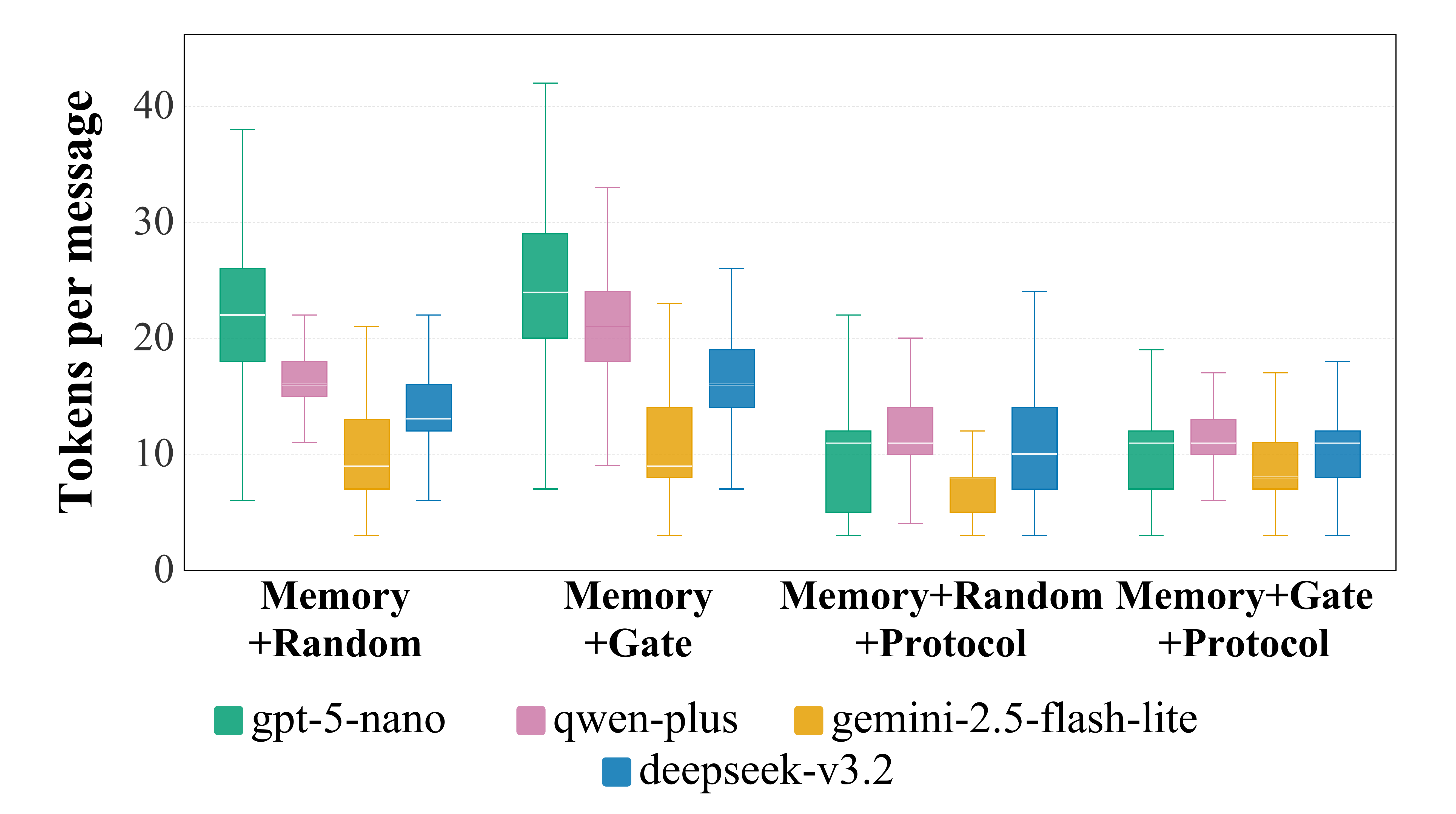}
        \caption{Message length distribution.}
        \label{fig:game_msglen}
    \end{subfigure}

    \vspace{-1mm}
    \caption{\textbf{Further analysis on \textsc{Game}.}
    (a) Round-wise score evolution on the $15{\times}15$ and $18{\times}18$ maps.
    (b) Per-message token distributions across LLM backbones and communication settings.}
    \label{fig:game_analysis}
\end{figure}

Figure~\ref{fig:game_roundcurve} shows that \textsc{Base} consistently underperforms memory-enabled variants, highlighting its limitations in long-horizon coordination. In terms of \textsc{routing}, \textsc{Memory+Gate} provides a robust uplift via selective routing. The best performance is achieved by \textsc{Memory+Gate+Protocol}, indicating that protocolized interactions effectively convert stored context into low-ambiguity coordination.

Figure~\ref{fig:game_msglen} shows that routing-only communication can still incur relatively long and variable messages. This high verbosity likely stems from the unstructured messages, where agents must repeatedly restate context to compensate for incoherent information flow. Adding \textsc{Protocol} substantially shortens messages and reduces variance under both \textsc{Random} and \textsc{Gate} routing, indicating that protocol mainly improves communication compactness and stability.

\section{Conclusion}
In this paper, we propose MASBench, a comprehensive benchmark designed to evaluate multi-agent collaboration under partially observable constraints. By structuring tasks into three progressive categories, our framework assesses key collaboration mechanisms of LLM-based agents. Through extensive experiments across diverse LLM backbones, we provide actionable empirical guidance, paving the way for the design of more efficient and effective multi-agent systems in the future.

\subsection*{AI use statement}

In this work, we used generative AI tools to assist with language editing and polishing of author-written text, including grammar correction, wording refinement, and readability improvements. Generative AI was also used as part of the synthetic data construction pipeline for the \textsc{Reasoning} task, including LLM-assisted generation and validation of benchmark instances. The \textsc{Scheduling} instances were generated programmatically, while the \textsc{Game} maps were manually specified. Generative AI tools were not used to replace the authors' judgment in experimental design, result interpretation, or the formulation of the paper's main conclusions. All AI-assisted content and data were reviewed by the authors, who take full responsibility for the final text, data, claims, and artifacts presented in
this work.


\bibliography{iclr2027_conference}
\bibliographystyle{iclr2027_conference}

\clearpage
\onecolumn
\appendix
\raggedbottom

\section{Extended Related Work on LLM-based Multi-Agent Systems}
\label{sec:extended_related_work}

Existing MAS broadly fall into two categories.
\textbf{(i) General orchestration frameworks} provide reusable abstractions for roles,
interaction patterns, and communication topologies~\citep{chen2023agentverse,wu2024autogen}.
For example, AgentVerse supports general-purpose task solving~\citep{chen2023agentverse}
with an iterative four-stage framework.
\textbf{(ii) Domain-specific frameworks} customize roles and workflows to meet the
requirements of specific domains.
In software engineering~\citep{hong2023metagpt,liu2024large,qian2024chatdev},
MetaGPT~\citep{hong2023metagpt} and ChatDev~\citep{qian2024chatdev} decompose the
development process into phased subtasks, where agents with distinct roles collaboratively
perform analysis, design, implementation, and testing.
In deep research~\citep{boiko2023autonomous,gridach2025agentic,ghafarollahi2025sciagents},
SciAgents leverages role-specialized agents, ontological knowledge graphs, and retrieval
tools to propose and critique scientific hypotheses~\citep{ghafarollahi2025sciagents}.

\section{Additional Experiments}


\subsection{Error Analysis}
\label{app:error_analysis}

We analyze $n{=}40$ sampled failure cases per task to characterize recurring
collaboration problems. We report their error distributions and examine
qualitative cases that illustrate how information is exchanged and used
during execution. The percentages describe the selected failures, not
failure rates over the full benchmark.

\subsubsection{Sampling and Annotation}

For each task, we construct a low-performing candidate pool in a stratified
manner. For every LLM backbone and mechanism setting, runs are ranked
separately within the Simple and Complex subsets by task score, and the
bottom 30\% from each subset are retained. We pool these candidates across
backbones, settings, and difficulty levels and randomly sample 40 cases for
manual analysis.

Each selected trajectory is manually inspected and assigned one primary
task-specific error category according to its dominant observed failure mode.
The percentages below characterize these diagnostic samples rather than
failure rates over the full benchmark.

\subsubsection{Error Distributions}
\label{app:error_distributions}

\begin{figure}[!htbp]
    \centering
    \includegraphics[width=\textwidth]
    {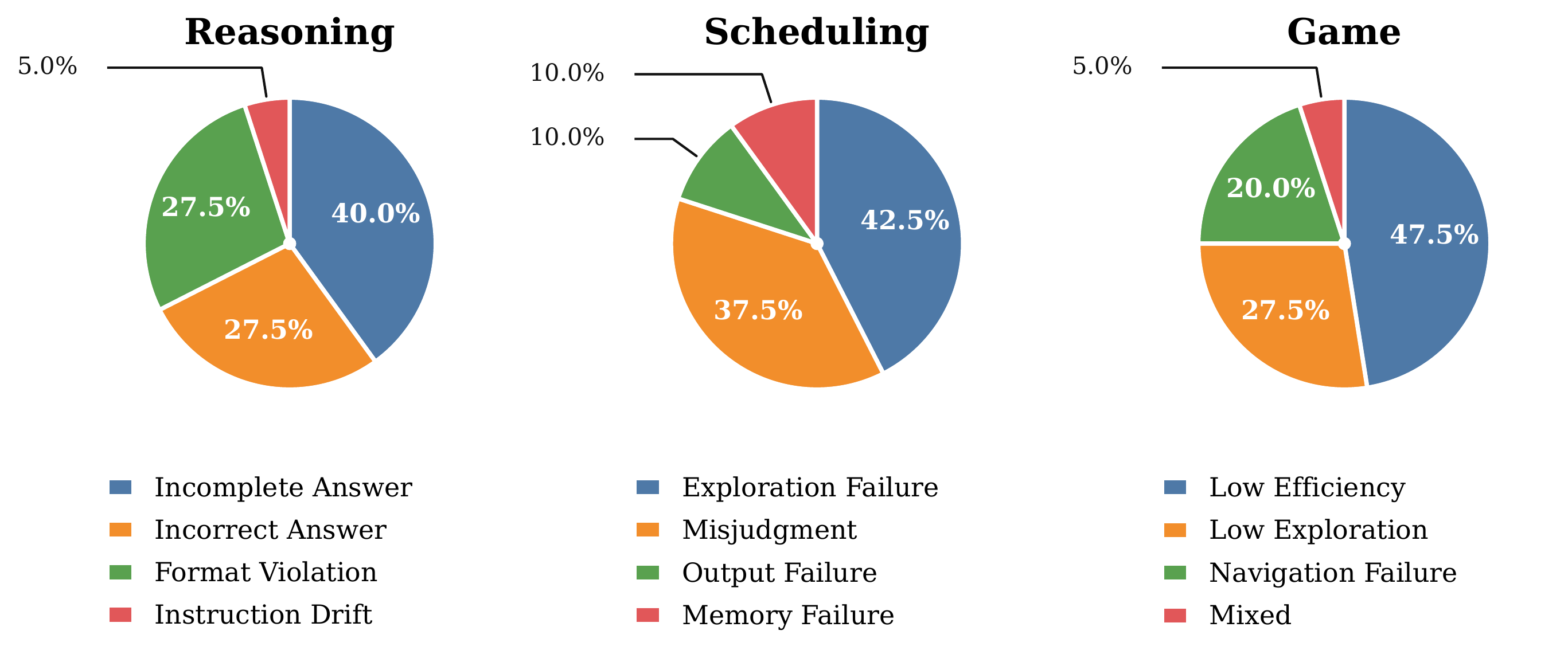}
    \caption{\textbf{Error type distributions across the three tasks.}
    Each task contains $n{=}40$ sampled failure cases; percentages are computed
    within each task's sample.}
    \label{fig:error_analysis}
\end{figure}

\paragraph{Reasoning.}
\textit{Incomplete Answer} is the most frequent category
(40.0\%, 16/40), followed by \textit{Incorrect Answer} and
\textit{Format Violation} (27.5\%, 11/40 each).
\textit{Instruction Drift} accounts for the remaining 5.0\% (2/40).
Incomplete answers omit required answer content and can lose recall under
the word-level $F_1$ metric even when part of the response is correct.
However, an incomplete final answer alone does not establish whether the
missing information was unavailable, inadequately exchanged, or omitted
during answer construction.
The sampled errors therefore motivate examining both information aggregation
and output compliance, rather than attributing every failure to incorrect
reasoning.

\paragraph{Scheduling.}
\textit{Exploration Failure} (42.5\%, 17/40) and \textit{Misjudgment}
(37.5\%, 15/40) together account for 80.0\% of the sample.
\textit{Output Failure} and \textit{Memory Failure} each account for 10.0\%
(4/40).
The dominant categories concern failing to surface feasible common windows
and incorrectly evaluating proposed candidates, respectively.
This distinction separates candidate search from candidate verification:
negotiation may fail either before a useful proposal is identified or when
agents assess its feasibility.
The distribution thus motivates inspecting proposal and verification
sequences, alongside the state maintained across rounds.

\paragraph{Game.}
\textit{Low Efficiency} accounts for 47.5\% (19/40), followed by
\textit{Low Exploration} at 27.5\% (11/40),
\textit{Navigation Failure} at 20.0\% (8/40), and
\textit{Mixed} at 5.0\% (2/40).
Efficiency and exploration issues together account for 75.0\% of the sample.
The reported problems include redundant or unproductive actions,
insufficient map coverage, and difficulties navigating toward task-relevant
locations.
These patterns motivate examining how agents turn local discoveries into
resource delivery, rather than treating navigation as the only source of
failure. The cases below distinguish useful information transfer from
incorrect use of received information.

\subsubsection{Qualitative Case Studies}
\label{app:error_cases}

We complement the distributions with matched comparisons in
\textsc{Reasoning} and \textsc{Scheduling}, and success and failure examples
in \textsc{Game}.
The cases illustrate interaction patterns rather than estimate their frequency.
The following summaries state the recorded outcomes; the marked fields are
reserved for the corresponding run metadata and original trace excerpts.

\paragraph{Reasoning: requesting missing evidence.}
In a matched case, \textsc{Protocol} requests missing evidence from the
partner and corrects the answer.
The diagnostic question is whether a request elicits information relevant
to the answer and how that information is subsequently used, not merely
whether the message follows the required syntax.
The current Protocol prompt requires a question and prohibits a final
answer in the first message. First-message behavior must therefore be
interpreted in the context of this explicit interaction rule.
The archived transcript below confirms the request--response sequence;
it does not independently establish which prompt revision was used.


\begin{quote}
\small
\textbf{Private information:}
The question asks whether Elara Vance or Rhys Calder has more combined
Nebula Quill and Crimson Scroll wins. Alice's evidence records two Nebula
Quills and one Crimson Scroll for Vance. Bob's evidence records one Nebula
Quill for Calder's \emph{Cygnus Prime} and three Crimson Scrolls for
\emph{Crown of Embers}, \emph{Scepter of Ash}, and \emph{Throne of Shadow}.
Thus, the totals are $3$ versus $4$; the reference answer is
``Rhys Calder.''\par

\textbf{Base trace:}
At turn 1, Alice states, ``Rhys Calder has no Nebula Quill or Crimson Scroll
wins shown in the texts'' and outputs
\texttt{<A>Elara Vance</A>}. The run terminates before Bob speaks.\par

\textbf{Protocol trace:}
At turn 1, Alice asks,
\texttt{<Q>}Do we have any other sources indicating Rhys Calder's wins
for Nebula Quill or Crimson Scroll beyond what's listed here?\texttt{</Q>}
At turn 2, Bob supplies
\texttt{<R>}Rhys Calder earned one Nebula Quill for 'Cygnus Prime'.\texttt{</R>}
and
\texttt{<R>}Rhys Calder won three Crimson Scroll awards for 'The Shardborne
Legacy' trilogy: 'Crown of Embers', 'Scepter of Ash', and 'Throne of Shadow'.\texttt{</R>}
At turn 3, Alice outputs \texttt{<A>Rhys Calder</A>}.\par

\textbf{Outcome:}
Base answers ``Elara Vance'' with word-level $F_1=0$;
Protocol answers ``Rhys Calder'' with $F_1=1$, matching the reference.
The failure is localized to treating evidence absent from Alice's private
context as absent from the full evidence set. The Protocol trajectory
obtains the missing award counts before committing to an answer.
\end{quote}

The case illustrates evidence-seeking interaction as a way to address an
information gap. It does not establish that the protocol eliminates
premature answers, which remain possible in the recorded diagnostic analysis.

\paragraph{Scheduling: state retention and coordinated verification.}
On the same calendars, \textsc{Base} and \textsc{Memory} fail to converge,
whereas \textsc{Memory+Protocol} reaches unanimous agreement on a feasible
slot in round 3.
The case records explicit tracking of pending and rejected slots with
Memory, and synchronized proposal and verification with Protocol.
It illustrates the distinction between retaining negotiation state and
organizing its use to reach a shared decision.


\begin{quote}
\small

\textbf{Relevant constraints:}
On Day 4, all participants finish breakfast at 08:00. Their next busy
intervals begin at 09:00, 10:00, 08:40, 10:10, and 09:30 for Members 1--5,
respectively, so 08:00--08:20 is feasible for everyone.
The stored valid-solution set independently lists this window as
solution 1 (indices $480$--$482$ in 10-minute units across five days).\par

\textbf{Base trace:}
Members 1 and 2 repeatedly propose Day 3 15:00--15:20 in rounds 3--5,
while Member 3 proposes other windows. In round 5, Member 3 explicitly
states, ``I cannot accept the 15:00-15:20 slot on Day3 due to conflicts.''
The run exhausts its budget and records ``No common meeting slot was found.''\par

\textbf{Memory trace:}
The state retains \texttt{candidate\_slots} and \texttt{rejected\_slots};
after round 3, the pending candidates include Day 4 08:00--08:20.
In rounds 4 and 5, Members 1--4 explicitly vote Yes for this slot, but
Member 5 replies only ``Day 4 08:00-08:20.'' The leader continues to
treat the slot as pending, stating in round 5,
``2 slots rejected. 2 pending slots awaiting votes.''
The run ends with \texttt{no\_slot} despite retaining a feasible candidate:
the unstructured reply is not converted into a final agreement.\par

\textbf{Memory+Protocol trace:}
Round 1 ranks Day 4 morning highest in the broad availability scan.
In round 2, Member 1 proposes Day 4 08:00--08:20; the leader includes
it as candidate index 5 among 15 candidates.
In round 3, all five recorded vote vectors have value 1 at index 5.
The leader declares, ``Winner found at Index 5 (5/5 votes).
Slot: Day 4 08:00-08:20,'' followed by ``Meeting Finalized.''\par

\textbf{Outcome:}
Base and Memory both record \texttt{no\_slot} after five rounds.
Memory+Protocol records \texttt{ok} in round 3 for Day 4 08:00--08:20,
matching solution 1 and all five calendars. The comparison separates
repeated incompatible proposals from failure to finalize a retained
candidate, and shows how explicit voting resolves the latter in this run.
\end{quote}

This comparison connects the search and verification difficulties in the
aggregate analysis to a concrete negotiation sequence.
Its interpretation concerns the evaluated workflow as a whole, including
state maintenance and phase-based coordination.

\paragraph{Game: sharing an unseen warehouse location.}
A successful communication case shows that routed messages support resource
delivery by sharing an otherwise unseen warehouse location.
The relevant link is between the information available before communication,
the transmitted location, and the subsequent deposit outcome.

\begin{quote}
\small

\textbf{Before communication:}
At round 3, \texttt{Agent\_0} at $(2,14)$ observes
warehouse $(3,15)$ with remaining capacities
\texttt{wood=0, sand=8, stone=20}.
\texttt{Agent\_1}, at $(15,14)$ and carrying four sand units,
sees only warehouse $(13,14)$, whose sand capacity is zero.
Warehouse $(3,15)$ is absent from its local view and recorded table memory.\par

\textbf{Message:}
\texttt{Agent\_0} sends to \texttt{Agent\_1} in round 3,
received in round 4:
``I'm moving right to (3,14) --- you're at (15,14), so we're on the same row.
There's a table at (3,15) with sand and stone if you need resources.''
The recipient's action prompt retains the location as
\texttt{proto discover fact=table@(3,15)}.\par

\textbf{Action and outcome:}
In round 4, \texttt{Agent\_1} explicitly identifies
$(3,15)$ as the newly reported target and chooses \texttt{UP},
then navigates leftward. After reaching $(3,15)$, it verifies
\texttt{sand=8} locally and selects \texttt{PUT\_DOWN}, \texttt{QTY: 4}
in round 19. The round-20 simulator state records an empty inventory,
four sand units contributed by this agent, and warehouse sand capacity
reduced from 8 to 4; the team score increases from 8 to 12.
\end{quote}

The case links communication to a task outcome, rather than to message
volume alone. It illustrates how sharing a local discovery can support
resource delivery under partial observability.

\paragraph{Game: relying on coarse capacity information.}
A separate failure case shows an agent relying on a coarse capacity message
despite a newer local observation.
Here, receiving information is not sufficient: the agent also needs to
reconcile it with directly observed constraints before acting.

\begin{quote}
\small

\textbf{Received message:}
\texttt{Agent\_2} sends in round 16, received in round 17:
``I'm moving down to (6,7); you're right next to the table at (3,8)
with sand capacity---let me know if you need help depositing.''
This coarse assurance follows its round-15 message, received in round 16,
that ``table at (3,8) has sand:4 available if needed'';
the latter appears in the recipient's prompt as
\texttt{proto discover fact=table@(3,8) total\_rem=4 can\_deposit=sand:4}.\par

\textbf{Newer observation:}
At round 18, \texttt{Agent\_3}'s current observation
places it on warehouse $(3,8)$ carrying four sand units.
\texttt{Current info} states a total placeable capacity of 28 but
\texttt{Per-type remaining at your current tile: wood=8, sand=0, stone=20}.
The simulator's warehouse state confirms these resource-specific values.\par

\textbf{Decision and outcome:}
The round-18 response acknowledges that sand capacity is zero,
but attributes the contradiction to ``a sync lag'' and chooses
\texttt{PUT\_DOWN}, \texttt{QTY: 4}, \texttt{TYPE: sand},
``trusting the team's verified report over stale per-type display''.
The action record has \texttt{action\_success=false};
the round-19 simulator state still shows four sand units carried,
unchanged warehouse capacities, and an unchanged team score of 8.
Thus, the attempted deposit produces no delivery or score gain.\par

\textbf{Failure localization:}
The response explicitly gives the earlier teammate reports
priority over a current, resource-specific observation when selecting the
deposit action. This supports a failure to reconcile conflicting evidence
at action selection. The records do not isolate whether the earlier
capacity discrepancy originated in memory updating or message interpretation.
\end{quote}

This case motivates resource-type-specific and recency-aware messages,
together with checking whether action selection respects current local
constraints. The summary alone does not establish which internal component
caused the inconsistency.

\paragraph{Cross-task implications.}
The cases highlight three complementary diagnostic questions: whether
missing evidence is requested and incorporated in \textsc{Reasoning},
whether proposals and feasibility checks are coordinated in
\textsc{Scheduling}, and whether received information is translated into
valid, productive actions in \textsc{Game}.
Together with the error distributions, these observations identify concrete
interaction patterns for further investigation without treating the selected
cases as population-level estimates of mechanism effects.

\begin{figure*}[!t]
    \centering
    \includegraphics[width=0.9\linewidth]{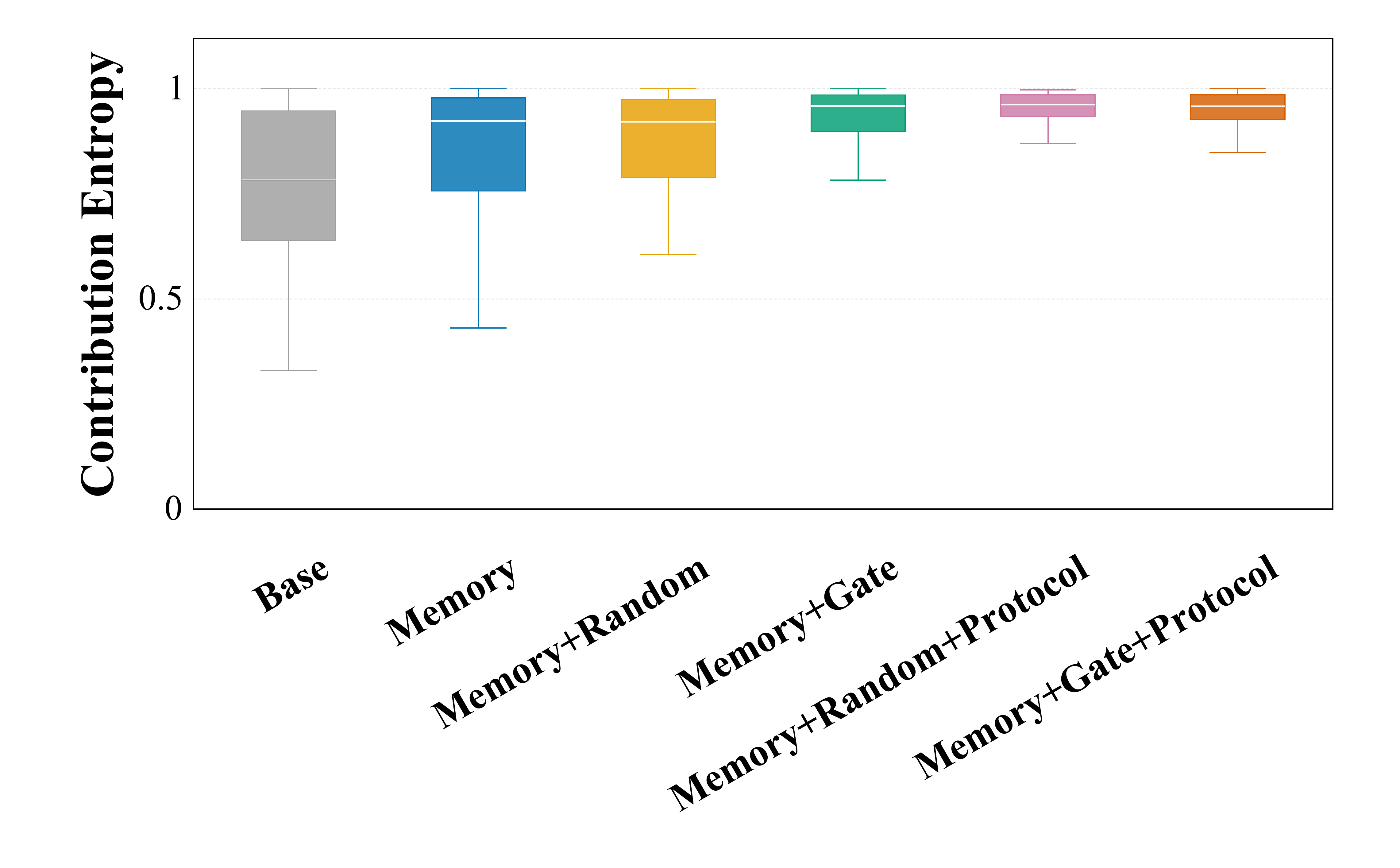}
    \caption{\textbf{Contribution Entropy in \textsc{Game}.} Higher and more stable entropy indicates balanced participation among agents.}
    \label{fig:game_entropy}
\end{figure*}

\subsection{Contribution Analysis on Game Category}
Figure~\ref{fig:game_entropy} evaluates participant fairness using Contribution Entropy. For each game instance, let $c_i$ denote the final cumulative contribution of agent $i$, computed from resources successfully deposited into the warehouses. We first normalize individual contributions into a distribution:
\begin{equation}
p_i = \frac{c_i}{\sum_{j=1}^{N} c_j}.
\end{equation}
The normalized contribution entropy is then computed as:
\begin{equation}
H_{\mathrm{norm}} = - \frac{1}{\log_2 N}\sum_{i=1}^{N} p_i \log_2 p_i,
\end{equation}
where $N$ is the number of agents and zero-contribution agents are included in the normalization. We exclude runs with no positive contribution from this fairness statistic, since such runs reflect zero task progress rather than an uneven contribution distribution. Runs in which only one agent has a positive contribution remain valid and receive entropy $0$. This scales the metric to $[0,1]$, with higher values indicating more evenly distributed contributions.

In the absence of protocol constraints (left four columns), the distribution is looser and entropy is generally lower, suggesting that neither memory nor routing logic alone guarantees equitable participation. While Gate routing improves stability over Random, both still show visible variance. In contrast, adding the Protocol yields higher normalized entropy and tighter distributions across backbones, suggesting that explicit interaction protocols regularize communication behavior and promote more balanced contributions.

\subsection{Additional Results of Round-wise Score Evolution}
Figures~\ref{fig:apx-reason}, ~\ref{fig:apx-schedule} and ~\ref{fig:apx-game} present the results of round-wise score evolution for the remaining evaluated backbones. The patterns are consistent with the Qwen backbone analyzed in Sec.~\ref{sec:further-reason}-\ref{sec:further-game}: the incorporation of the task-relevant collaboration mechanisms significantly contributes to the overall performance score, leading to a faster convergence during a multi-round collaboration.

The curves also reveal task-dependent convergence behaviors. 
In \textsc{Reasoning} and \textsc{Scheduling}, most gains emerge in early rounds, since agents mainly exchange missing clues or eliminate infeasible meeting slots. In contrast, \textsc{Game} exhibits more gradual score growth because agents must first explore the environment, accumulate spatial knowledge, and coordinate resource delivery over longer horizons. This further supports our progressive task design across the three categories.

\begin{figure*}[htbp]
  \centering

  \begin{subfigure}[b]{0.48\textwidth}
    \centering
    \includegraphics[width=\linewidth]{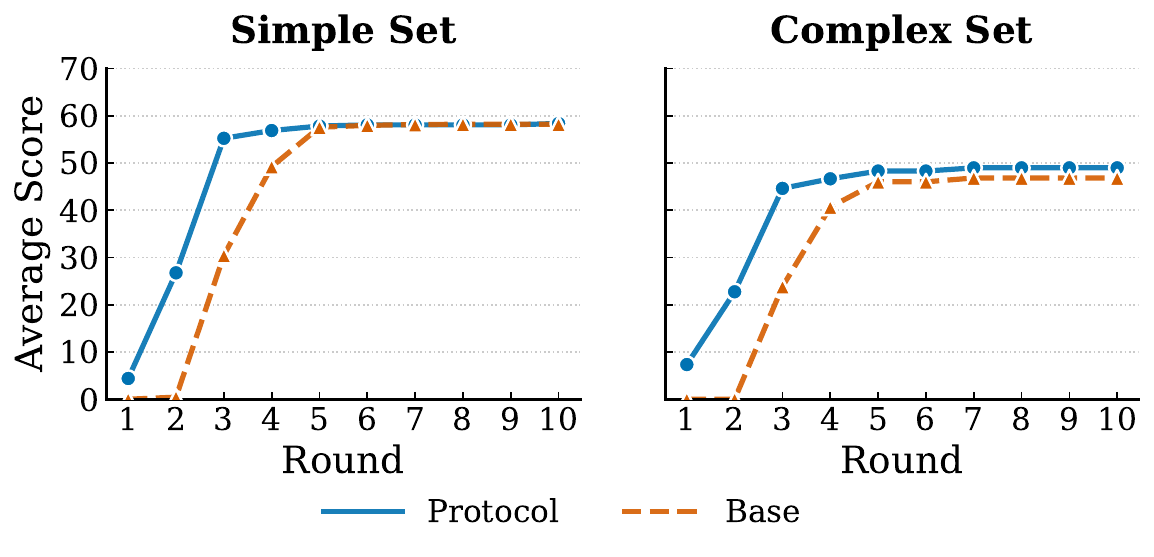}
    \caption{Round-wise score evolution for \texttt{DeepSeek}.}
  \end{subfigure}
  \hfill
  \begin{subfigure}[b]{0.48\textwidth}
    \centering
    \includegraphics[width=\linewidth]{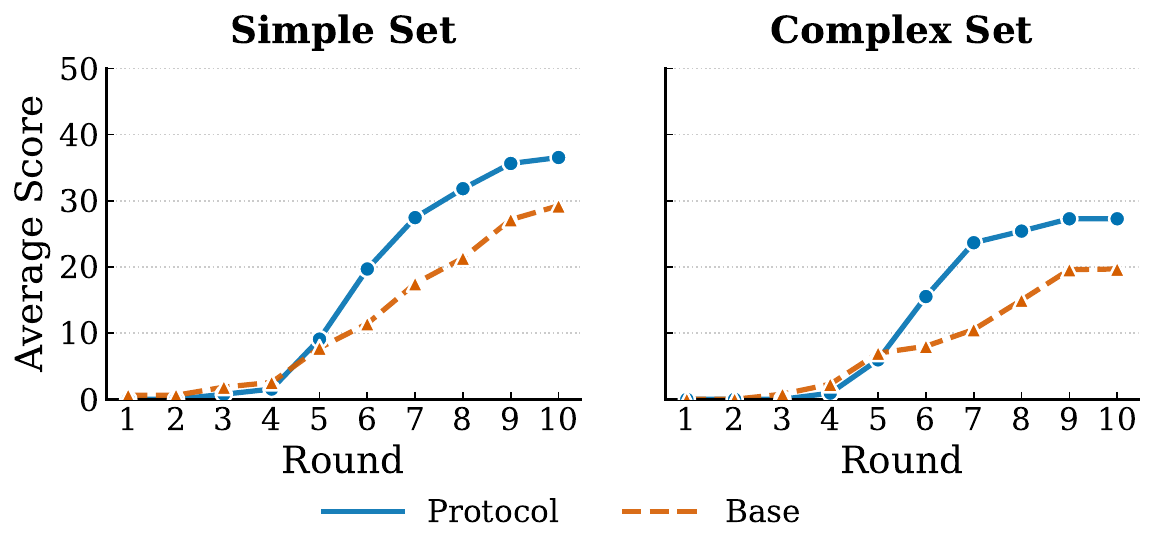}
    \caption{Round-wise score evolution for \texttt{Llama}.}
  \end{subfigure}

  \medskip

  \begin{subfigure}[b]{0.48\textwidth}
    \centering
    \includegraphics[width=\linewidth]{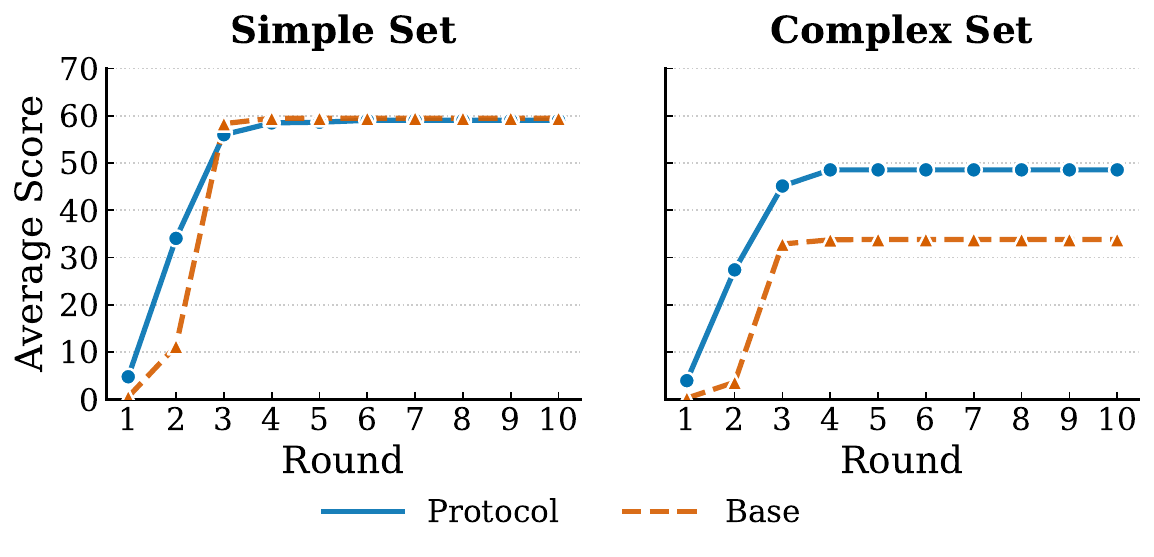}
    \caption{Round-wise score evolution for \texttt{Gemini}.}
  \end{subfigure}
  \hfill
  \begin{subfigure}[b]{0.48\textwidth}
    \centering
    \includegraphics[width=\linewidth]{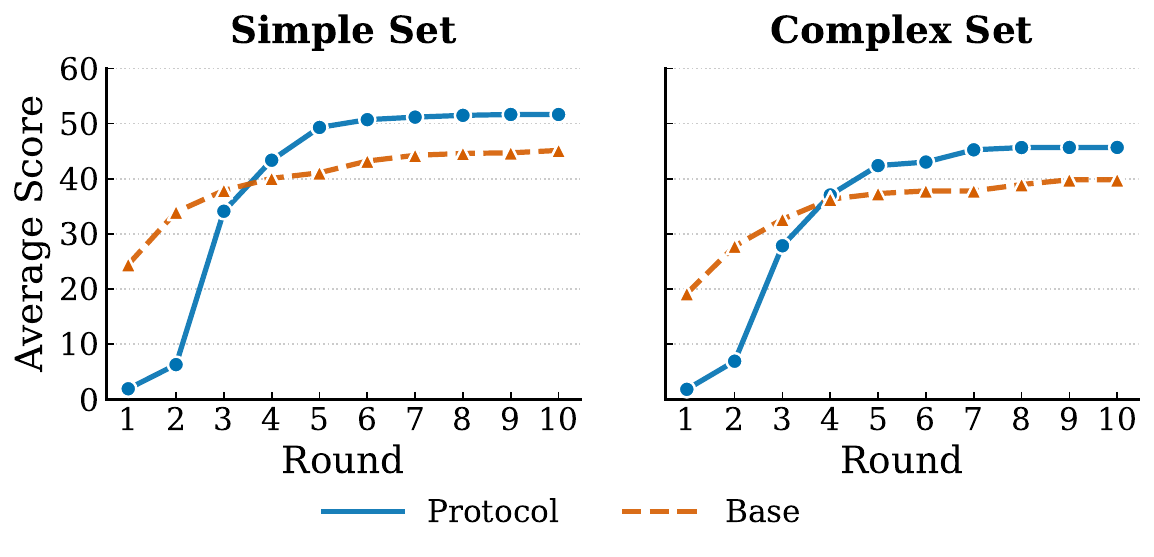}
    \caption{Round-wise score evolution for \texttt{GPT}.}
  \end{subfigure}

  \caption{Round-wise score evolution in \textsc{Reasoning}.}
  \label{fig:apx-reason}
\end{figure*}

\begin{figure*}[!t]
\centering
\begin{subfigure}[t]{0.32\textwidth}
  \centering
  \includegraphics[width=\linewidth]{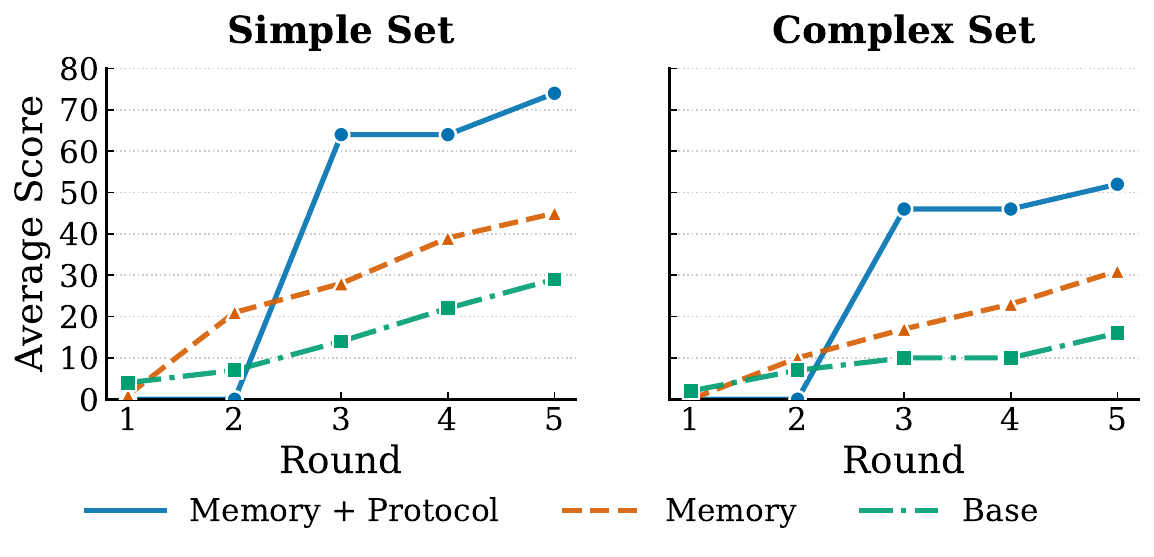}
  \caption{Round-wise score evolution for \texttt{DeepSeek}.}
  \label{fig:ex1}
\end{subfigure}\hfill
\begin{subfigure}[t]{0.32\textwidth}
  \centering
  \includegraphics[width=\linewidth]{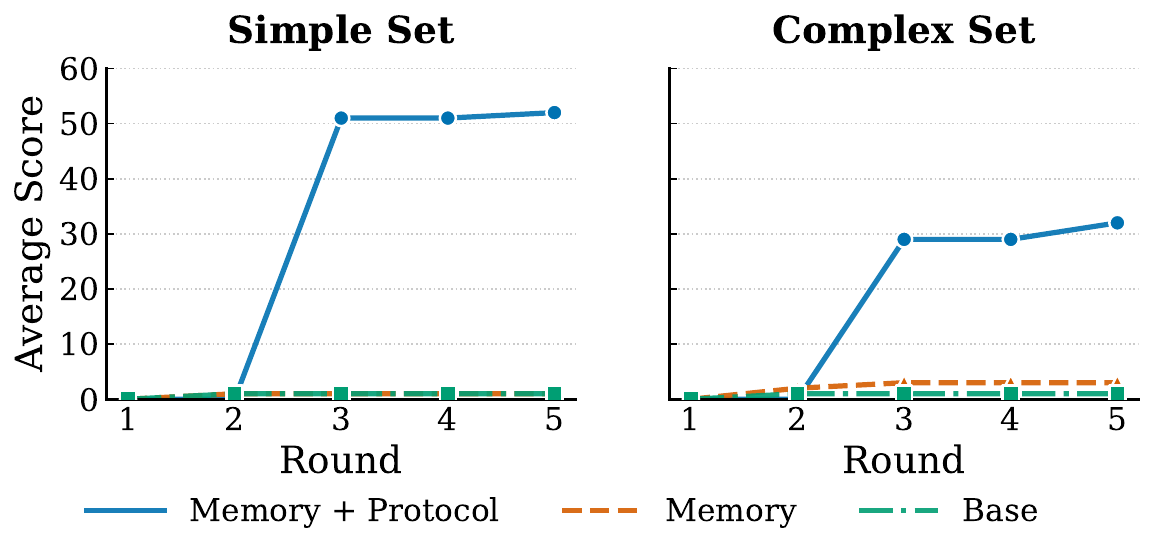}
  \caption{Round-wise score evolution for \texttt{GPT}.}
  \label{fig:ex2}
\end{subfigure}\hfill
\begin{subfigure}[t]{0.32\textwidth}
  \centering
  \includegraphics[width=\linewidth]{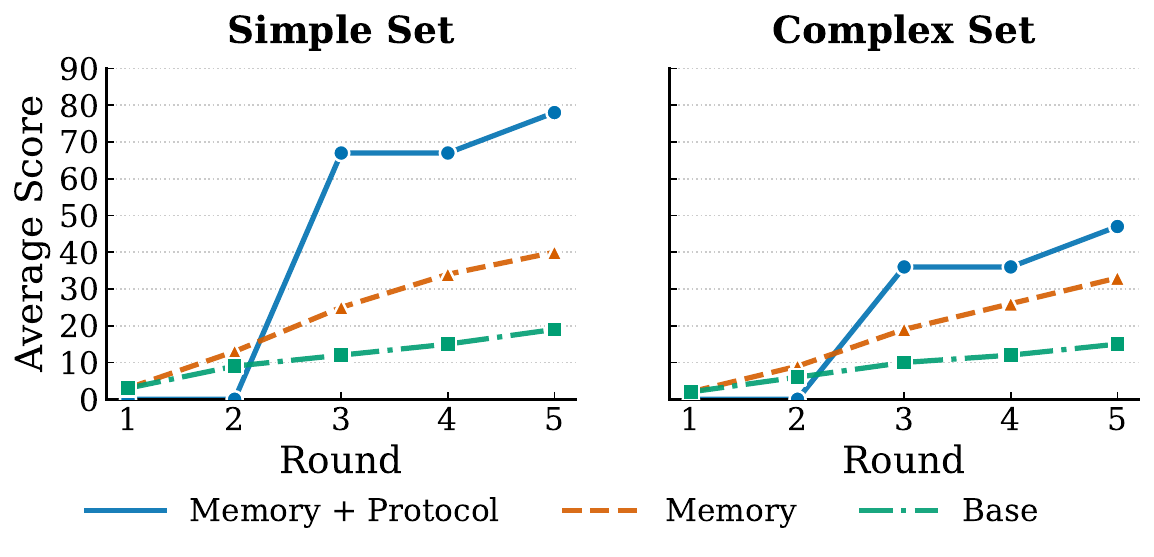}
  \caption{Round-wise score evolution for \texttt{Gemini}.}
  \label{fig:ex3}
\end{subfigure}

\caption{Round-wise score evolution in \textsc{Scheduling}.}
\label{fig:apx-schedule}
\end{figure*}

\begin{figure*}[!t]
\centering
\begin{subfigure}[t]{0.32\textwidth}
  \centering
  \includegraphics[width=\linewidth]{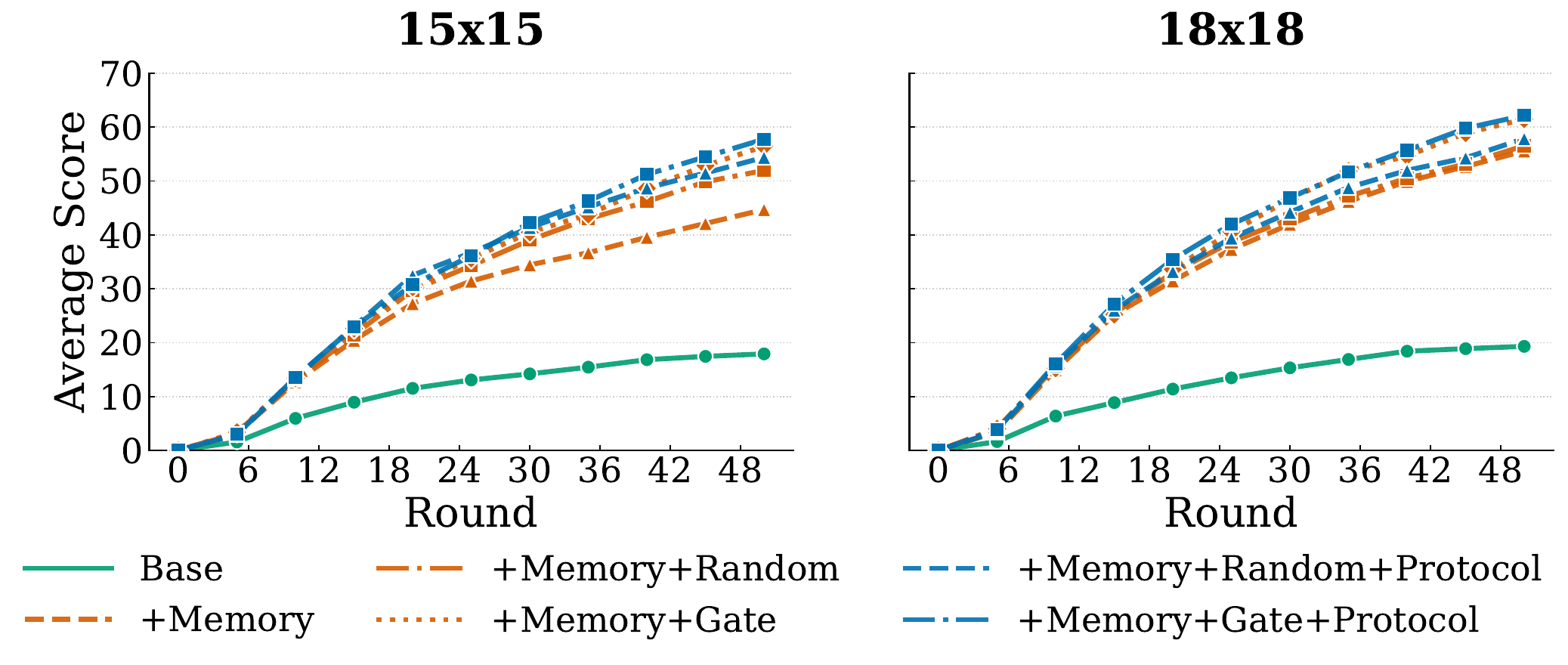}
  \caption{Round-wise score evolution for \texttt{DeepSeek}.}
  \label{fig:ex4}
\end{subfigure}\hfill
\begin{subfigure}[t]{0.32\textwidth}
  \centering
  \includegraphics[width=\linewidth]{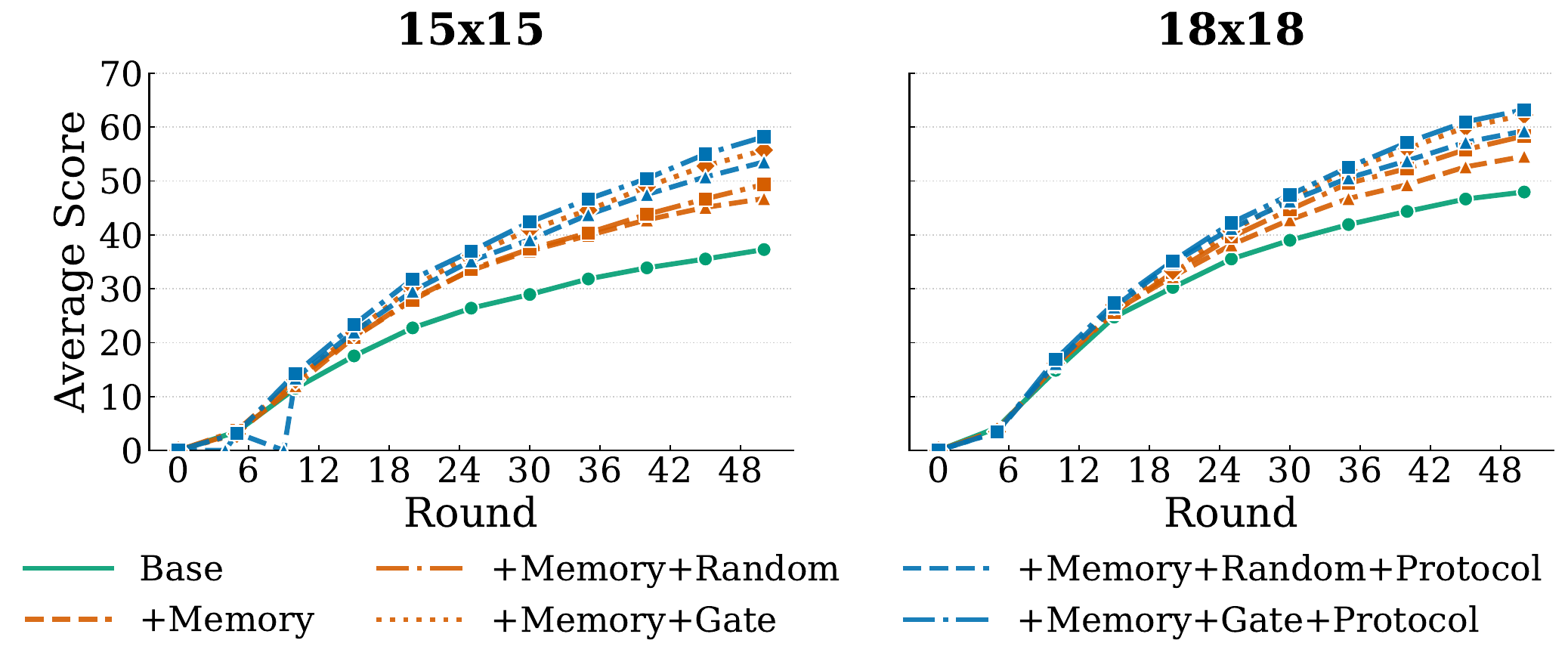}
  \caption{Round-wise score evolution for \texttt{GPT}.}
  \label{fig:ex5}
\end{subfigure}\hfill
\begin{subfigure}[t]{0.32\textwidth}
  \centering
  \includegraphics[width=\linewidth]{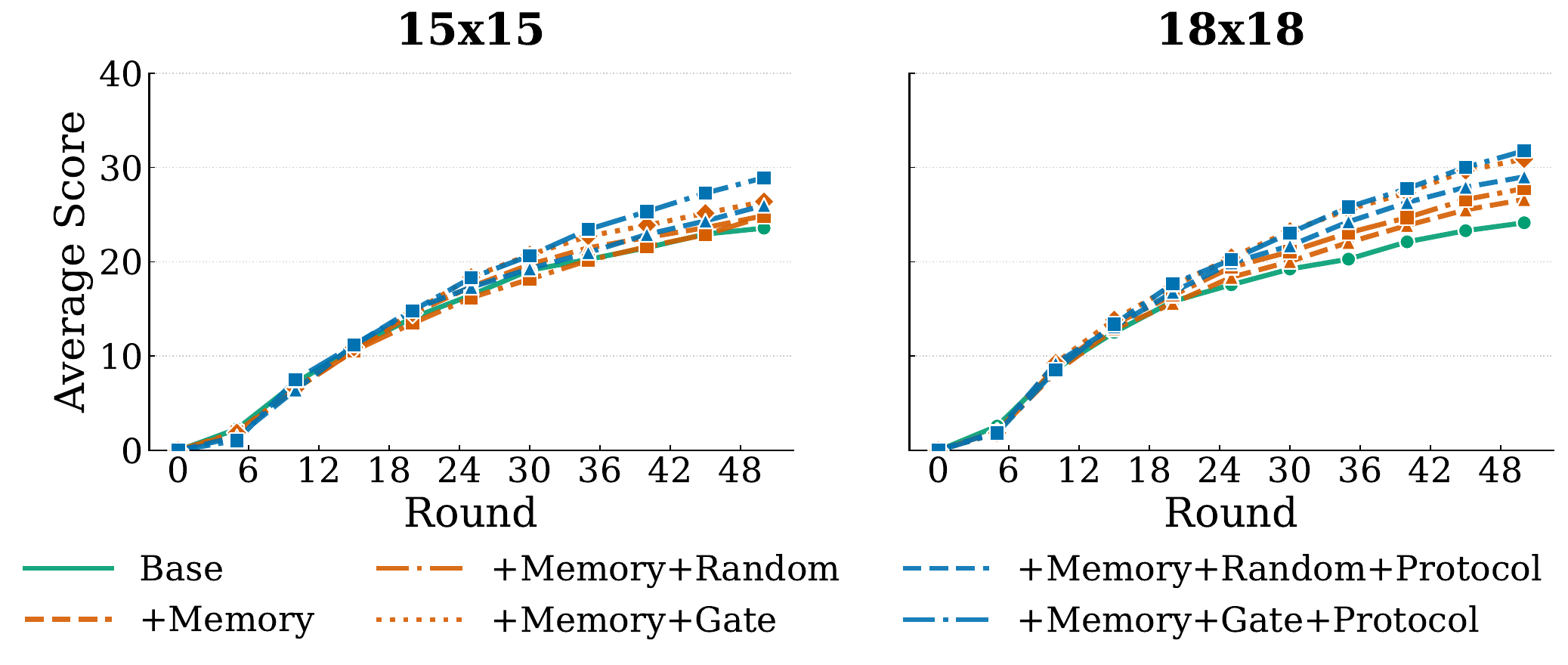}
  \caption{Round-wise score evolution for \texttt{Gemini}.}
  \label{fig:ex6}
\end{subfigure}

\caption{Round-wise score evolution in \textsc{Game}.}
\label{fig:apx-game}
\end{figure*}

\subsection{Additional Game Results with Full Message-Phase Token Cost}

\begin{table*}[!t]
\centering
\caption{\textbf{Experimental results on Game with full message-phase token accounting.} Score$\uparrow$, Tok$\downarrow$ and Eff$\uparrow$ denote performance score, message-phase completion cost, and communication efficiency. Unlike the main results, Tok counts all completion tokens generated during the message phase. \textbf{Bold} and \underline{underline} denote the best and second-best results per backbone.}
\label{tab:game_full_msg_cost}

\resizebox{\textwidth}{!}{%
    \begin{tabular}{l l
                    S[table-format=3.1] S[table-format=6.0] S[table-format=3.2]
                    S[table-format=3.1] S[table-format=6.0] S[table-format=3.2]
                    S[table-format=3.1] S[table-format=6.0] S[table-format=3.2]}
    \toprule
    \multirow{2}{*}{Model} & \multirow{2}{*}{Setting} &
    \multicolumn{3}{c}{Simple} &
    \multicolumn{3}{c}{Complex} &
    \multicolumn{3}{c}{Overall} \\
    \cmidrule(lr){3-5}\cmidrule(lr){6-8}\cmidrule(lr){9-11}
    & &
    {Score$\uparrow$} & {Tok$\downarrow$} & {Eff$\uparrow$} &
    {Score$\uparrow$} & {Tok$\downarrow$} & {Eff$\uparrow$} &
    {Score$\uparrow$} & {Tok$\downarrow$} & {Eff$\uparrow$} \\
    \midrule

     \multirow{6}{*}{GPT-5-nano} 
     & Base & {37.27} & \multicolumn{1}{c}{---} & \multicolumn{1}{c}{---} & {47.95} & \multicolumn{1}{c}{---} & \multicolumn{1}{c}{---} & {42.61} & \multicolumn{1}{c}{---} & \multicolumn{1}{c}{---} \\
     & Memory & {46.73} & \multicolumn{1}{c}{---} & \multicolumn{1}{c}{---} & {54.57} & \multicolumn{1}{c}{---} & \multicolumn{1}{c}{---} & {50.65} & \multicolumn{1}{c}{---} & \multicolumn{1}{c}{---} \\
     & Memory + Random & {49.36} & {7,607} & {6.49} & {58.35} & {9,740} & {5.99} & {53.85} & {8,674} & {6.21} \\
     & Memory + Gate & {\underline{55.72}} & {8,991} & {6.20} & {\underline{62.23}} & {11,041} & {5.64} & {\underline{58.98}} & {10,016} & {5.89} \\
     & Memory + Random + Protocol & {53.53} & {\underline{5,868}} & {\underline{9.12}} & {59.29} & {\underline{7,880}} & {\underline{7.52}} & {56.41} & {\underline{6,874}} & {\underline{8.21}} \\
     & Memory + Gate + Protocol & {\textbf{58.20}} & {\textbf{4,112}} & {\textbf{14.15}} & {\textbf{63.21}} & {\textbf{3,061}} & {\textbf{20.65}} & {\textbf{60.71}} & {\textbf{3,587}} & {\textbf{16.92}} \\
    \midrule

     \multirow{6}{*}{Gemini-2.5-Flash-Lite} 
     & Base & {23.57} & \multicolumn{1}{c}{---} & \multicolumn{1}{c}{---} & {24.14} & \multicolumn{1}{c}{---} & \multicolumn{1}{c}{---} & {23.85} & \multicolumn{1}{c}{---} & \multicolumn{1}{c}{---} \\
     & Memory & {24.89} & \multicolumn{1}{c}{---} & \multicolumn{1}{c}{---} & {26.61} & \multicolumn{1}{c}{---} & \multicolumn{1}{c}{---} & {25.75} & \multicolumn{1}{c}{---} & \multicolumn{1}{c}{---} \\
     & Memory + Random & {24.82} & {3,241} & {7.66} & {27.77} & {4,291} & {6.47} & {26.29} & {3,766} & {6.98} \\
     & Memory + Gate & {\underline{26.39}} & {\textbf{2,910}} & {\underline{9.07}} & {\underline{30.90}} & {\underline{3,999}} & {\underline{7.73}} & {\underline{28.64}} & {\underline{3,454}} & {\underline{8.29}} \\
     & Memory + Random + Protocol & {26.09} & {3,873} & {6.74} & {28.87} & {4,926} & {5.86} & {27.48} & {4,400} & {6.25} \\
     & Memory + Gate + Protocol & {\textbf{28.87}} & {\underline{2,995}} & {\textbf{9.64}} & {\textbf{31.95}} & {\textbf{3,726}} & {\textbf{8.57}} & {\textbf{30.41}} & {\textbf{3,360}} & {\textbf{9.05}} \\
    \midrule

     \multirow{6}{*}{Deepseek-v3.2} 
     & Base & {17.91} & \multicolumn{1}{c}{---} & \multicolumn{1}{c}{---} & {19.26} & \multicolumn{1}{c}{---} & \multicolumn{1}{c}{---} & {18.59} & \multicolumn{1}{c}{---} & \multicolumn{1}{c}{---} \\
     & Memory & {44.69} & \multicolumn{1}{c}{---} & \multicolumn{1}{c}{---} & {55.56} & \multicolumn{1}{c}{---} & \multicolumn{1}{c}{---} & {50.12} & \multicolumn{1}{c}{---} & \multicolumn{1}{c}{---} \\
     & Memory + Random & {51.93} & {\textbf{3,606}} & {14.40} & {56.53} & {\underline{4,512}} & {12.53} & {54.23} & {\textbf{4,059}} & {13.36} \\
     & Memory + Gate & {\underline{56.44}} & {\underline{3,624}} & {\underline{15.58}} & {\underline{61.47}} & {4,549} & {\underline{13.51}} & {\underline{58.95}} & {4,086} & {\underline{14.43}} \\
     & Memory + Random + Protocol & {54.49} & {4,482} & {12.16} & {58.32} & {5,584} & {10.44} & {56.40} & {5,033} & {11.21} \\
     & Memory + Gate + Protocol & {\textbf{57.67}} & {3,702} & {\textbf{15.58}} & {\textbf{62.47}} & {\textbf{4,424}} & {\textbf{14.12}} & {\textbf{60.07}} & {\underline{4,063}} & {\textbf{14.79}} \\
    \midrule

     \multirow{6}{*}{Qwen-Plus} 
     & Base & {34.47} & \multicolumn{1}{c}{---} & \multicolumn{1}{c}{---} & {43.00} & \multicolumn{1}{c}{---} & \multicolumn{1}{c}{---} & {38.73} & \multicolumn{1}{c}{---} & \multicolumn{1}{c}{---} \\
     & Memory & {49.05} & \multicolumn{1}{c}{---} & \multicolumn{1}{c}{---} & {57.73} & \multicolumn{1}{c}{---} & \multicolumn{1}{c}{---} & {53.39} & \multicolumn{1}{c}{---} & \multicolumn{1}{c}{---} \\
     & Memory + Random & {49.37} & {4,967} & {9.94} & {57.03} & {6,333} & {9.01} & {53.20} & {5,650} & {9.42} \\
     & Memory + Gate & {\underline{56.52}} & {4,783} & {11.82} & {\underline{63.01}} & {\underline{5,902}} & {\underline{10.68}} & {\underline{59.77}} & {\underline{5,342}} & {\underline{11.19}} \\
     & Memory + Random + Protocol & {56.45} & {\underline{4,740}} & {\underline{11.91}} & {58.41} & {6,057} & {9.64} & {57.43} & {5,399} & {10.64} \\
     & Memory + Gate + Protocol & {\textbf{59.26}} & {\textbf{3,970}} & {\textbf{14.93}} & {\textbf{64.85}} & {\textbf{4,902}} & {\textbf{13.23}} & {\textbf{62.05}} & {\textbf{4,436}} & {\textbf{13.99}} \\

    \bottomrule
    \end{tabular}%
} 

\end{table*}

Table~\ref{tab:game_full_msg_cost} reports an alternative cost accounting for the \textsc{Game} category. 
Unlike the main results, where Tok includes only tokens generated for inter-agent messages, here Tok includes all completion tokens generated during the message phase. This provides a more conservative estimate of communication overhead, especially for routing methods that require additional send/recipient decisions.

The results show that lower transmitted communication does not always imply lower message-phase generation cost. In particular, \textsc{Memory+Gate} uses fewer transmitted tokens than \textsc{Memory+Random} for every backbone,
but its full message-phase token cost is higher for GPT-5-nano and slightly higher for DeepSeek-v3.2 because of routing-decision overhead. In contrast, adding \textsc{Protocol} to \textsc{Gate} improves overall message-phase
efficiency across backbones, although the reduction in generation cost varies substantially across models. These results distinguish communication volume from the additional generation overhead required to decide when and where to communicate.

\subsection{Cross-Round Memory Ablation on Scheduling}
\label{app:cross_round_memory}

To disentangle cross-round state retention from the availability preprocessing
used in the full \textsc{Memory} configuration, we introduce an intermediate
\textsc{Cross-round Memory} variant for \textsc{Scheduling}.
This variant retains the raw busy-calendar representation used by
\textsc{Base} and does not perform gap extraction or construct the verified
availability whitelist.
Its only modification is to preserve negotiation state and interaction history
across rounds.
The full \textsc{Memory} configuration additionally uses the precomputed
availability representation described in Appendix~\ref{app:mechanisms}.
Table~\ref{tab:scheduling_cross_round_memory} compares these three settings.

\begin{table*}[!ht]
\centering
\caption{\textbf{Cross-round memory ablation on Scheduling.}
Score$\uparrow$, Tok$\downarrow$, and Eff$\uparrow$ denote performance score,
communication cost, and communication efficiency.
\textbf{Bold} and \underline{underline} indicate the best and second-best
values among the three settings for each backbone and split; ties share the
same marking.}
\label{tab:scheduling_cross_round_memory}
\vspace{-0.05\baselineskip}

\setlength{\tabcolsep}{5pt}
\renewcommand{\arraystretch}{1.07}

\resizebox{0.85\textwidth}{!}{%
\begin{tabular}{l l *{3}{S[table-format=2.2] S[table-format=5.0] S[table-format=3.2]}}
\toprule
\multirow{2}{*}{Model} & \multirow{2}{*}{Setting} &
\multicolumn{3}{c}{Simple} &
\multicolumn{3}{c}{Complex} &
\multicolumn{3}{c}{Overall} \\
\cmidrule(lr){3-5}\cmidrule(lr){6-8}\cmidrule(lr){9-11}
& &
{Score$\uparrow$} & {Tok$\downarrow$} & {Eff$\uparrow$} &
{Score$\uparrow$} & {Tok$\downarrow$} & {Eff$\uparrow$} &
{Score$\uparrow$} & {Tok$\downarrow$} & {Eff$\uparrow$} \\
\midrule

\multirow{3}{*}{GPT-5-nano}
& Base
& {\underline{1}} & {\textbf{1,131}} & {\underline{0.88}}
& {1} & {\textbf{1,101}} & {0.91}
& {\underline{1}} & {\textbf{1,116}} & {0.90} \\

& Cross-round Memory
& {\textbf{2}} & {\underline{1,259}} & {\textbf{1.59}}
& {\underline{2}} & {\underline{1,185}} & {\underline{1.69}}
& {\textbf{2}} & {\underline{1,222}} & {\textbf{1.64}} \\

& Memory
& {\underline{1}} & {1,842} & {0.54}
& {\textbf{3}} & {1,683} & {\textbf{1.78}}
& {\textbf{2}} & {1,763} & {\underline{1.13}} \\
\midrule

\multirow{3}{*}{Gemini-2.5-Flash-Lite}
& Base
& {19} & {\textbf{996}} & {19.07}
& {15} & {\textbf{1,001}} & {14.98}
& {17} & {\textbf{999}} & {17.02} \\

& Cross-round Memory
& {\underline{31}} & {\underline{1,310}} & {\textbf{23.67}}
& {\underline{23}} & {\underline{1,378}} & {\underline{16.69}}
& {\underline{27}} & {\underline{1,344}} & {\underline{20.09}} \\

& Memory
& {\textbf{40}} & {1,816} & {\underline{22.03}}
& {\textbf{33}} & {1,795} & {\textbf{18.39}}
& {\textbf{36.5}} & {1,805} & {\textbf{20.22}} \\
\midrule

\multirow{3}{*}{DeepSeek-v3.2}
& Base
& {29} & {\textbf{388}} & {\underline{74.66}}
& {16} & {\textbf{448}} & {35.74}
& {22.5} & {\textbf{418}} & {53.82} \\

& Cross-round Memory
& {\underline{43}} & {652} & {65.95}
& {\underline{30}} & {\underline{623}} & {\underline{48.15}}
& {\underline{36.5}} & {638} & {\underline{57.21}} \\

& Memory
& {\textbf{45}} & {\underline{507}} & {\textbf{88.77}}
& {\textbf{31}} & {638} & {\textbf{48.61}}
& {\textbf{38}} & {\underline{572}} & {\textbf{66.39}} \\
\midrule

\multirow{3}{*}{Qwen-Plus}
& Base
& {16} & {\textbf{506}} & {31.65}
& {11} & {\textbf{525}} & {20.94}
& {13.5} & {\textbf{515}} & {26.19} \\

& Cross-round Memory
& {\underline{49}} & {\underline{984}} & {\textbf{49.78}}
& {\textbf{30}} & {\underline{1,153}} & {\textbf{26.02}}
& {\underline{39.5}} & {\underline{1,069}} & {\textbf{36.96}} \\

& Memory
& {\textbf{51}} & {1,063} & {\underline{47.96}}
& {\underline{29}} & {1,273} & {\underline{22.77}}
& {\textbf{40}} & {1,168} & {\underline{34.23}} \\

\bottomrule
\end{tabular}%
}

\vspace{-0.1\baselineskip}
\end{table*}

\paragraph{Analysis.}
Cross-round state retention already yields a consistent performance improvement
over \textsc{Base}: \textsc{Cross-round Memory} achieves a higher Score in all
12 backbone--split comparisons.
At the Overall level, Score increases from 1.0 to 2.0 for GPT-5-nano,
17.0 to 27.0 for Gemini-2.5-Flash-Lite, 22.5 to 36.5 for DeepSeek-v3.2,
and 13.5 to 39.5 for Qwen-Plus.
Because this variant keeps the raw calendar representation and does not use
availability preprocessing, these results show that the gains associated with
the full \textsc{Memory} configuration cannot be attributed solely to the
precomputed availability whitelist.
Persistent information across negotiation rounds contributes independently
to the observed improvement.

The full \textsc{Memory} configuration provides additional gains in most
settings.
Across the 12 backbone--split Score comparisons, it outperforms
\textsc{Cross-round Memory} in nine cases, ties in one, and is lower in two.
At the Overall level, the two variants tie for GPT-5-nano, while full
\textsc{Memory} increases Score from 27.0 to 36.5 for Gemini,
36.5 to 38.0 for DeepSeek, and 39.5 to 40.0 for Qwen.
This suggests that availability preprocessing and structured negotiation state
provide additional benefits beyond cross-round retention, although their
incremental effect is model- and split-dependent.

The ablation also reveals a communication trade-off.
\textsc{Cross-round Memory} uses more communication tokens than \textsc{Base}
in all 12 comparisons, but fewer tokens than the full \textsc{Memory}
configuration in 10 of 12.
Despite this additional communication relative to \textsc{Base},
its communication efficiency is higher in 11 of 12 comparisons, with the
DeepSeek-v3.2 Simple split as the only exception.
Thus, the results separate two effects within the Scheduling memory design:
cross-round retention provides a broad performance benefit, while the full
structured Memory configuration usually provides further task-performance
gains at the cost of additional communication.

\section{Framework-Level Evaluation Details}
\label{app:framework_setup}

This section describes the framework-level evaluation in
Sec.~\ref{sec:framework_eval}.
We evaluate implementations built with
CAMEL~\citep{li2023camel} and AgentVerse~\citep{chen2023agentverse}
on the three MASBench tasks.
Task adapters enforce the benchmark's observation boundaries and
interaction rules, while the frameworks provide persistent agents and
conversation management.
The reference uses \textsc{Protocol} for \textsc{Reasoning},
\textsc{Memory+Protocol} for \textsc{Scheduling}, and
\textsc{Memory+Gate+Protocol} for \textsc{Game}.

\subsection{Evaluation Subsets and Information Boundaries}

We sample 30\% of the instances without replacement within each task's
difficulty strata, using sampling seed 42.
The selected subsets contain 90 \textsc{Reasoning} instances
(60 Simple and 30 Complex), 60 \textsc{Scheduling} instances
(30 Simple and 30 Complex), and 30 \textsc{Game} maps
(15 Simple and 15 Complex).
The same instance manifests are used across frameworks and backbones,
and reference records are matched using the corresponding instance
identifiers.
All performance and communication-cost values in
Table~\ref{tab:framework_results}, including those of the reference,
are computed on these selected subsets rather than on the full benchmark.

The evaluated backbones are GPT-5-nano,
\texttt{gemini-2.5-flash-lite}, DeepSeek-v3.2, and \texttt{qwen-plus}.
The task-specific information boundaries are preserved throughout
evaluation: each reasoning agent receives only its own evidence partition,
each scheduling member receives only its own availability, and each game
agent receives its local observation and messages addressed to it.
Other agents' private inputs and the evaluator's ground truth are not
provided to the collaborating agents.
Information is shared only through the communication channels permitted
by the task.

\subsection{Framework Integration}

Each task participant is represented by a persistent framework agent.
CAMEL uses \texttt{ChatAgent} with its native conversation memory.
For \textsc{Reasoning} and \textsc{Scheduling}, AgentVerse uses
\texttt{ConversationAgent} within \texttt{BasicEnvironment}, with sequential
speaking order and public-message visibility for all task participants.
Its basic memory updater stores public messages in
\texttt{ChatHistoryMemory}.
Private inputs remain in the corresponding agent's role description and
are not broadcast.

The adapters deliver public utterances through the frameworks' memory
interfaces while retaining their native prompt templates and history
serialization.
Thus, shared task inputs and observation boundaries do not imply identical
serialized model requests.

For \textsc{Game}, communication follows the benchmark's restricted
point-to-point channel rather than public broadcast.
Action responses remain private, and an explicitly generated message is
delivered to its selected eligible recipient in the next round, subject
to the environment's delivery rules.
The framework variants disable the reference scaffold's additional view,
action, response, and message memories.
To bound conversation history, CAMEL uses a native window of 32 message
entries, while AgentVerse includes the most recent 16 model outputs from
its chat history.
Under their respective storage formats, these windows cover approximately
eight action--message rounds; they are not defined as equal token budgets.

\subsection{Task-Specific Interaction and Termination}

\paragraph{Reasoning.}
Alice and Bob receive the shared question and their respective private
context blocks.
Both framework implementations use the plain Base task prompt.
Alice speaks first, and the agents alternate for at most ten individual
turns, rather than ten pairs of turns.
Execution stops at the first answer recognized by the benchmark parser,
using an answer tag or a boxed answer, or at the turn limit if no answer
is recognized.
Conversation history is maintained by the framework throughout the run.

The framework runs use temperature zero and a requested output limit of
4,096 tokens per turn.
For GPT-5-nano, the adapter raises the output-token limit to 8,192 to
accommodate reasoning-token usage.

\paragraph{Scheduling.}
Each instance contains five members with private schedules over five days.
The objective is to agree on a common 20-minute meeting window.
The adapter deterministically converts each member's schedule into maximal
free intervals lasting at least 20 minutes and supplies these intervals
only to the corresponding agent.
This preprocessing replaces the reference implementation's LLM-based
availability extraction.

The framework implementations use the plain task prompt and native
conversation memory, without the reference's explicit candidate-state
controller or an additional team-memory summarizer.
Each round consists of one utterance from each of the five members in
sequence.
After a complete round, a separate stateless checker examines the public
transcript for explicit agreement on a single slot.
The checker has no access to private availability and does not propose
meeting slots.
The conversation ends when the checker reports completion or after
five rounds.

A deterministic evaluator checks the proposed window against all five
private schedules.
An infeasible proposal is scored as a failure even if the checker reports
consensus; the evaluator does not repair the proposal.

\paragraph{Game.}
Each framework agent interacts with the environment for 50 rounds,
making one action call and one message-decision call per round.
The message decision specifies whether to send, the eligible recipient,
and the message content.
The environment retains the benchmark's action semantics, observation
and communication ranges, resource constraints, and scoring rules.
Episodes run for the full horizon without an early-stop heuristic.

The Game runner is configured with environment and model seed 42,
temperature zero, and top-$p$ equal to one.
The reported \textsc{Game} results are averaged over five repeated
executions.
Only averaged results have been retained, so run-to-run standard
deviations cannot be recovered from the available records.
We therefore report the means without repeated-run error bars or
claims of statistical significance.

\subsection{Evaluation Metrics and Aggregation}

\paragraph{Performance.}
Performance follows Sec.~\ref{sec:metrics}.
For \textsc{Reasoning}, we compute word-level $F_1$ using the shared
scoring routine, which lowercases text, splits at non-alphanumeric
boundaries, and retains articles.
Instances without a recognized answer receive zero.
For \textsc{Scheduling}, success requires a final meeting window that
is feasible for all participants.
For \textsc{Game}, the raw resource score is normalized by the
map-specific reference score before aggregation.
Scores are scaled by 100 for reporting.

Overall performance weights instances equally rather than averaging
the Simple and Complex subset means with equal weight.
For \textsc{Game}, repetition-level results are averaged for each map
before the map-level means are aggregated.

\paragraph{Communication cost.}
Tok measures the total tokens generated for inter-agent messages, averaged
over evaluation instances. Prompt tokens, private reasoning, private action-generation outputs, and
non-message control outputs are excluded.
The same accounting scope is applied to the reference and framework
implementations, including message-format tokens and structured fields
that are actually transmitted.

In \textsc{Reasoning}, the counted communication consists of the
partner-visible \texttt{MESSAGE} blocks for the reference and public
utterances for the framework implementations.
In \textsc{Scheduling}, it consists of the messages exchanged through
the shared conversation.
In \textsc{Game}, it includes the complete transmitted payload, including
any transmitted \texttt{STATE}, \texttt{OBS}, and \texttt{MSG} fields.

Prompt tokens, private reasoning, private action-generation outputs,
and control outputs that are not transmitted to other agents are excluded.
Accordingly, Tok measures communication volume rather than end-to-end
inference cost.
Differences in Tok should not be interpreted directly as differences in
total generated tokens, API expenditure, or execution time.

\subsection{Scope of Interpretation}

The comparison evaluates complete system configurations under common
task instances and observation boundaries.
The reference includes task-specific collaboration mechanisms, whereas
the framework implementations use their native agent histories and
message serialization together with the task adapters described above.
Differences may therefore reflect multiple implementation choices,
including preprocessing, conversation management, interaction control,
and message construction.

These results demonstrate how MASBench can evaluate framework-backed
multi-agent systems and distinguish their task-dependent performance
and communication behavior.
They complement the within-scaffold mechanism comparisons, but do not
isolate the causal effect of replacing only the underlying framework
library.

\section{Benchmark Design Details}
\subsection{Reasoning}
\label{app:reasoning_details}
\subsubsection{Data Generation based on Multi-hop Question Answering}
We adapt multi-hop instances from HotpotQA~\cite{yang2018hotpotqa} to form a collaborative inference task under partial observability. For each problem, we decompose its evidence chain and distribute it across two agents so that no single agent has sufficient information to solve the task independently.

\subsubsection{Quality Assurance and Scoring}
We implement a multi-stage validation pipeline to filter low-quality generations. Each entry is assigned a composite score $s \in [0, 4]$, defined as:
$$s = s_{\text{fmt}} + \mathbb{I}_{\text{q}} + \mathbb{I}_{\text{sup}} + \mathbb{I}_{\text{irr}}$$
where $\mathbb{I}_{(\cdot)}$ denotes a binary indicator function. The components are defined as follows:
\begin{itemize}
    \item \textbf{Format Check ($s_{\text{fmt}}$):} Validates JSON schema integrity and softly scores supporting-fact coverage. For each agent, the coverage score is the number of assigned supporting facts divided by 5 and capped at 1; $s_{\mathrm{fmt}}$ is the average of the two agents' scores. 

    \item \textbf{Question Validity ($\mathbb{I}_{\text{q}}$):} Verified via \texttt{Gemini-Flash}; the question must be unanswerable without the provided context.
    \item \textbf{Sufficiency ($\mathbb{I}_{\text{sup}}$):} The referenced supporting-fact sentences alone must be sufficient to recover the ground-truth answer.
    \item \textbf{Irrelevance ($\mathbb{I}_{\text{irr}}$):} After removing all supporting facts, the remaining context must be insufficient to recover the answer, ensuring no ``shortcuts'' exist.
\end{itemize}
We employ a \textbf{Retry-and-Select} strategy, attempting generation up to 5 times per seed until a score ($s>=3.5$) is achieved. This LLM-assisted validation is used for dataset construction and filtering.

\subsubsection{Difficulty Stratification}
For each constructed instance, we compute a heuristic difficulty score based on the number of supporting facts $N_f$, the total context length $L_c$, and the ground-truth answer length $L_a$:
\begin{equation}
\mathrm{Difficulty} = 10 \cdot N_f + \frac{L_c}{500} + \frac{L_a}{5}.
\end{equation}
We use this score only for dataset stratification, not for evaluation. Instances are split into \textit{Simple} and \textit{Complex} subsets according to this score, with the threshold selected to match the target dataset statistics.

\subsubsection{Task Specification and Evaluation}
\paragraph{Input.}
Each agent is provided with a personalized, partitioned view of the context and the global question. Neither agent possesses sufficient information to solve the query independently.
$$ \text{Input}_i = \{ \text{Question}, \mathcal{C}_i, \mathcal{S}_i \} $$
where $\mathcal{C}_i$ is the masked context and $\mathcal{S}_i$ is the partial set of supporting facts.

\paragraph{Output and Evaluation.}
Agents interact via a multi-turn dialogue (up to 10 rounds). The system evaluates the final answer extracted from specific tags (e.g., \texttt{<A>answer</A>}).
\begin{itemize}
    \item \textbf{Reasoning Accuracy:} Measured by word-level F1 score between the predicted and reference answers after normalization:
    $$ F1 = \frac{2 \cdot \text{Precision} \cdot \text{Recall}}{\text{Precision} + \text{Recall}}. $$
    \item \textbf{Communication Cost:} Measured by the total message tokens, excluding internal thought processes and final answer tokens.
\end{itemize}

\subsubsection{Example}
To illustrate the data structure and constraints of the reasoning task, we provide a concrete example of a constructed instance below.
\begin{tcolorbox}[colback=gray!5!white, colframe=gray!75!black, title=\textbf{Example Instance: Reasoning Task}]
\fontsize{9pt}{11pt}\selectfont

\textbf{Task Definition:}
\begin{itemize}[leftmargin=*]
    \item \textbf{Question:} Evaluate the adaptability of the Krellnari and Zydonians to extreme environments and identify the fundamental difference in their interstellar propulsion.
    \item \textbf{Type:} Comparative Analysis
    \item \textbf{Difficulty:} Complex
\end{itemize}

\textbf{Private Observations (Excerpt):}
Information is distributed between two agents. Each agent possesses unique records regarding the physiology, home worlds, and technologies of the two species.

\begin{itemize}[leftmargin=*, nosep]
    \item \textbf{Agent A:}
    \begin{itemize}[leftmargin=*, nosep]
        \item \textit{Krellnari Home (Xylos):} Subterranean cities powered by superheated geothermal vents.
        \item \textit{Social Dynamics:} Operate as a vast hive-mind via the \textbf{Nexus Mind}.
        \item \textit{Physiology:} Silicon-based; withstands $T > 500$ K and high pressure, but highly specialized and limited in adaptability.
        \item \textit{Navigation:} Uses a network of \textbf{Fold-Space Anchors} for instantaneous transit between fixed points.
    \end{itemize}

    \item \textbf{Agent B:}
    \begin{itemize}[leftmargin=*, nosep]
        \item \textit{Zydonian Home (Volantis):} Buoyant metropolises in the upper atmosphere of an ammonia gas giant.
        \item \textit{Communication:} Bio-luminescent patterns on epidermis augmented by pheromonal signals.
        \item \textit{Resilience:} Unique cellular regeneration allows survival in a vacuum and protection against cosmic radiation and diverse atmospheric compositions.
        \item \textit{Propulsion:} \textbf{Warp-Bubble engines} that distort spacetime for continuous travel to any destination without fixed infrastructure.
    \end{itemize}
\end{itemize}

\textbf{Ground Truth:}
The correct answer requires comparing environmental adaptability and propulsion dependency.
\begin{itemize}[nosep]
    \item \textbf{Adaptability:} \textbf{Zydonians} (due to cellular regeneration vs. Krellnari specialization).
    \item \textbf{Travel Difference:} \textbf{Fixed vs. Mobile} (Fold-Space Anchors require pre-established points; Warp-Bubbles allow free navigation).
\end{itemize}

\textbf{Reference-style Output:}
A comparative response: \textit{``The Zydonians exhibit superior adaptability... Krellnari rely on stationary Fold-Space Anchors... Zydonians utilize mobile Warp-Bubble engines.''}
\end{tcolorbox}

\subsection{Scheduling}
\label{app:scheduling_details}

To ensure the \texttt{Scheduling} benchmark serves as a rigorous testbed for multi-agent coordination, we implement a controllable generation pipeline that manages schedule density, solvability, and solution diversity.

\subsubsection{Calendar Generation}
The timeline is modeled as a set of discrete time units. Let $m$ be the granularity parameter (\textit{i.e.,} $m=10$ minutes). The total time horizon is defined as $T = D \times 24 \times 60 / m$ units, where $D=5$ is the number of days. A valid solution is a continuous meeting window of duration $\delta=20$ minutes whose start and end times are aligned to the discretized timeline and feasible for all agents.

\paragraph{Initialization and Fixed Constraints.}
For each agent, the schedule is initialized with fixed ``Rest'' blocks (e.g., sleep, meals) based on daily routines (e.g., Sleep: 00:00--07:00, Lunch: 12:00--14:00). These intervals are strictly marked as \texttt{BUSY} and removed from the available time slots $\mathcal{T}_{avail}$.

\paragraph{Stochastic Busy Filling.}
To simulate realistic constraints,  we fill the remaining non-fixed slots with random activities. A busy ratio $\rho_i\in[0.3,0.6]$ is sampled for each agent. The generator iteratively samples a duration $d \in [10, 60]$ and an activity type (e.g., ``study'', ``work'') to insert a \texttt{BUSY} block, until the schedule density reaches $\rho_i$.

After stochastic busy filling, we enumerate all common feasible meeting windows across agents. Instances with too few valid windows are discarded and regenerated, while instances with sufficient valid windows are further processed by candidate pruning.

\subsubsection{Candidate Solution Pruning}
To strictly control the difficulty level,  we implement a \textbf{Candidate Pruning} algorithm that controls the number of valid meeting windows. This ensures that the final dataset contains exactly $K$ valid meeting slots specified by \texttt{candidate\_num}.

Let $\mathcal{V}$ be the set of all valid meeting windows initially found in the generated schedules. If $|\mathcal{V}|<K$, the instance is discarded and regenerated. Otherwise, the algorithm proceeds as follows:

\begin{enumerate}
    \item \textbf{Protection Set ($\mathcal{P}$):} We randomly sample $K$ solutions from $\mathcal{V}$ to form the protected set $\mathcal{P} \subset \mathcal{V}$. These are the intended ground truths we wish to preserve.
    \item \textbf{Greedy Obstruction:} We iteratively select a time point and an agent to insert a minimum-unit \texttt{BUSY} block. The goal is to invalidate non-protected solutions ($\mathcal{V} \setminus \mathcal{P}$) without affecting $\mathcal{P}$.
    \item \textbf{Termination:} The process repeats until $|\mathcal{V}| = K$. The schedules are then rewritten to include these artificial obstructions.
\end{enumerate}

\subsubsection{Difficulty Stratification}
After candidate pruning, the number of valid meeting windows $K$ is used as the difficulty control variable. 
Instances with more feasible windows are considered easier because agents have more opportunities to reach a valid consensus. 
Accordingly, we define \textit{Simple} instances with $K\in\{4,5\}$ and \textit{Complex} instances with $K\in\{2,3\}$. 
This stratification controls negotiation difficulty while ensuring that every retained instance remains solvable.

\subsubsection{Task Specification and Evaluation}

\paragraph{Input.}
Agents are provided with their individual schedules in a human-readable format (e.g., \texttt{YYYY-MM-DD HH:MM - HH:MM: Activity}). They must coordinate to find a common time slot of duration $\delta$ (i.e., 20 minutes).

\paragraph{Output and Evaluation.}
The system parses and normalizes the final proposal $\hat{s}=(t_{\mathrm{start}}, t_{\mathrm{end}})$ into the standard time format. Let $\mathcal{S}_{\mathrm{GT}}$ be the set of valid solutions derived from the final pruned schedules. The evaluation metric is strict Exact Match:

\[
\mathbb{I}_{\mathrm{succ}} =
\begin{cases}
1, & \hat{s} \in \mathcal{S}_{\mathrm{GT}},\\
0, & \text{otherwise}.
\end{cases}
\]
Partial overlaps are considered failures to enforce precise time calculation and duration satisfaction.

\subsubsection{Example}
To illustrate the data structure and constraints of the \texttt{Scheduling} task, we provide a concrete example of a generated instance below. 

\begin{tcolorbox}[colback=gray!5!white, colframe=gray!75!black, title=\textbf{Example Instance: Scheduling Task}]
\fontsize{10pt}{11pt}\selectfont

\textbf{Task Definition:}
\begin{itemize}[leftmargin=*]
    \item \textbf{Task Type:} \texttt{meeting}
    \item \textbf{Duration Constraint:} 20 minutes (2 units)
    \item \textbf{Horizon:} 5 days
    \item \textbf{Granularity:} $m=10$ minutes
\end{itemize}

\textbf{Private Observations (Excerpt):}
Each agent receives only their own local schedule (\texttt{Member*\_final\_schedule.json}). They must infer availability without seeing others' data. Below are snippets from \textbf{Day 5} for five agents.

\begin{itemize}[leftmargin=*, nosep]
    \item \textbf{Member 1:} \dots \\
    \texttt{Day 5 07:00-08:00 Rest: breakfast} \\
    \texttt{Day 5 08:20-08:30 Busy: study} \\
    \texttt{Day 5 09:20-11:50 Busy: watch movie} \textit{...}

    \item \textbf{Member 2:} \dots \\
    \texttt{Day 5 07:00-08:00 Rest: breakfast} \\
    \texttt{Day 5 10:40-10:50 Busy: meeting} \\
    \texttt{Day 5 11:10-11:50 Busy: shopping} \textit{...}

    \item \textbf{Member 3:} \dots \\
    \texttt{Day 5 07:00-08:00 Rest: breakfast} \\
    \texttt{Day 5 09:40-09:50 Busy: shopping} \\
    \texttt{Day 5 09:50-10:00 Busy: meeting} \textit{...}

    \item \textbf{Member 4:} \dots \\
    \texttt{Day 5 07:00-08:00 Rest: breakfast} \\
    \texttt{Day 5 08:30-08:40 Busy: work} \\
    \texttt{Day 5 09:20-10:30 Busy: work} \textit{...}

    \item \textbf{Member 5:} \dots \\
    \texttt{Day 5 07:00-08:00 Rest: breakfast} \\
    \texttt{Day 5 09:50-10:00 Busy: shopping} \\
    \texttt{Day 5 14:30-15:00 Busy: watch movie}  \textit{...}
\end{itemize}

\textbf{Ground Truth Solutions:}
The system identifies all valid time slots where \textit{all} members are free for at least 20 minutes.
\begin{itemize}[nosep]
    \item \texttt{S0: Day 5 08:40-09:00}
    \item \texttt{S1: Day 5 08:50-09:10}
    \item \texttt{S2: Day 5 09:00-09:20}
    \item \texttt{S3: Day 5 14:10-14:30}
\end{itemize}

\textbf{Required Output:}
The agents must negotiate and output a single string: \texttt{Day <d> HH:MM-HH:MM} (e.g., \texttt{Day 5 08:40-09:00}).
\end{tcolorbox}

\subsection{Game}
\label{app:game_details}

\subsubsection{Environment}
The Game task is a cooperative resource-collection game in a partially observable 2D grid. Agents must discover resources, move them to compatible warehouses, and coordinate under a fixed episode budget.

\begin{itemize}
    \item \textbf{Coordinates:} The simulator uses Cartesian coordinates $(x,y)$, with $x$ increasing from left to right and $y$ increasing from bottom to top. The local ASCII view marks the acting agent with \texttt{Y}, other agents with \texttt{a}, resources with \texttt{w/s/S}, and warehouses with \texttt{T}.
    \item \textbf{Two-layer cells:} Each cell has a terrain layer and an item layer. Terrain types include empty, mountain, river, wall, warehouse, forest, and grass. Mountains, rivers, and walls are impassable; warehouses are passable deposit tiles. The item layer stores stackable wood, sand, and stone resources.
    \item \textbf{Warehouses:} Warehouses are deposit-only. Once resources are placed on a warehouse, they contribute to the team score and cannot be picked up again. Each warehouse has a total capacity and per-resource capacity limits; deposited resources count toward the score only within these limits.
    \item \textbf{Partial observability:} Each agent observes only a square local window centered on itself. Out-of-map cells are rendered as \texttt{*}. The current implementation does not use line-of-sight occlusion; visibility is determined by window distance.
    \item \textbf{Observation dictionary:} The agent observation includes the current round, local ASCII view, global position, inventory status, current-tile information, visible resources, visible warehouses with remaining capacity, visible teammates, and messages delivered from the previous round.
    \item \textbf{Map families:} Simple instances use $15\times15$ maps, while complex instances use $18\times18$ maps with more agents and resources. Maps are manually specified and converted into machine-readable simulator files.
\end{itemize}

\subsubsection{Objectives and Scoring}

The objective is to maximize the amount of correctly deposited resources within $50$ rounds. A resource can contribute to the score only after it is successfully deposited on a warehouse cell. Each warehouse $i$ has two levels of storage constraints: a total capacity $C_i$ and a per-type capacity $L_{i,k}$ for each resource type $k$. A deposit action is accepted only if it does not violate either constraint, i.e., the total number of resources stored in warehouse $i$ cannot exceed $C_i$, and the number of type-$k$ resources stored in warehouse $i$ cannot exceed $L_{i,k}$. 

Let $n_{i,k}$ denote the amount of resource type $k$ deposited on warehouse $i$, and $T_k$ the global target quota for type $k$. Let $\mathcal{K}=\{\mathrm{wood}, \mathrm{sand}, \mathrm{stone}\}$. 
For each resource type $k$, its effective deposited amount is
\[
c_k = \sum_i \min(n_{i,k}, L_{i,k}).
\]
The final game score is then
\[
S = \sum_{k\in\mathcal{K}} \min(c_k, T_k).
\]
Resources left on non-warehouse cells, or deposited beyond the warehouse/type limits, do not increase the score.

\paragraph{Score Normalization.}
To make scores comparable across maps with different resource layouts and
capacity constraints, we compute a map-specific heuristic reference score
$U_m$ offline for each map $m$.
The reference planner has access to the complete map and does not use an LLM.

Specifically, it reads agent initial positions, obstacles, resource quantities,
inventory capacities, warehouse capacities, and task quotas.
Shortest-path distances are computed with BFS.
For each candidate delivery, the estimated transportation cost is the distance
from the agent's current position to the resource, plus the distance from the
resource to a compatible warehouse, together with one round for pickup and one
round for deposit.
Subject to the 50-round horizon and capacity constraints, the planner evaluates
multiple greedy task-allocation heuristics based on factors such as
reward-to-time ratio, travel distance, and unmet resource demand, under both
global and round-robin allocation schemes.
We use the highest score obtained across these heuristic plans as $U_m$.

For a run on map $m$ with raw score $S_m$, the normalized score is
\[
s_m =
\min\left(
100,\;
100\cdot\frac{S_m}{U_m}
\right),
\]
and the reported Game score is the mean of $s_m$ over evaluated maps.

Importantly, $U_m$ is a heuristic reference score rather than a guaranteed
global optimum or strict upper bound.
The planner assumes no inter-agent collisions or path conflicts, and only when
it reaches the total task quota can the resulting score be identified with the
map's theoretical maximum. In the reported experiments, we did not observe any run with
$S_m > U_m$; thus, the clipping operation did not affect the reported results.

\paragraph{Action Space.}
Agents output one discrete action per round:
\begin{itemize}
    \item \textbf{Movement:} \texttt{UP}, \texttt{DOWN}, \texttt{LEFT}, \texttt{RIGHT}, and \texttt{STAY} move to adjacent cells or remain in place.
    \item \textbf{Pickup:} \texttt{PICK\_UP [QTY] [TYPE]} collects resources from the current tile only. An agent can carry multiple units but only one resource type at a time.
    \item \textbf{Deposit:} \texttt{PUT\_DOWN [QTY]} deposits the carried resource on the current tile only. To score, the current tile must be a compatible warehouse with positive remaining capacity for the carried type.
\end{itemize}
If an explicit quantity violates inventory, tile, or warehouse capacity constraints, the action fails rather than being silently repaired.

\paragraph{Per-round Simulation Process.}
The current Game runtime uses a synchronous loop:
\begin{enumerate}
    \item Messages generated in round $t-1$ are delivered to their recipients.
    \item Each agent receives its local observation and generates an action.
    \item If communication is enabled, each agent may generate at most one point-to-point message using the same pre-update observation context.
    \item The environment resolves movement conflicts, executes \texttt{PICK\_UP}/\texttt{PUT\_DOWN}, updates warehouse inventories and score, and advances the round counter.
\end{enumerate}

\paragraph{Repeated Evaluation.}
For each backbone and mechanism configuration, every map is executed five
times. Performance and communication statistics are first averaged across
the five runs for each map and then aggregated across maps.

\subsubsection{Map Examples}
We provide two concrete examples to illustrate the resource distribution and warehouse constraints.

\begin{tcolorbox}[
  colback=gray!5,
  colframe=gray!60!black,
  boxrule=0.8pt,
  arc=1.5mm,
  left=1mm,right=1mm,top=1mm,bottom=1mm,
  title=\textbf{Example 1: Standard Map (15$\times$15)},
  fonttitle=\bfseries
]
\small
\RaggedRight
\noindent\textbf{File:} \texttt{15x15\_map/15x15\_v1.json}

\textbf{Map Layout:}\par\vspace{1mm}
\begin{center}
    \includegraphics[width=0.75\linewidth]{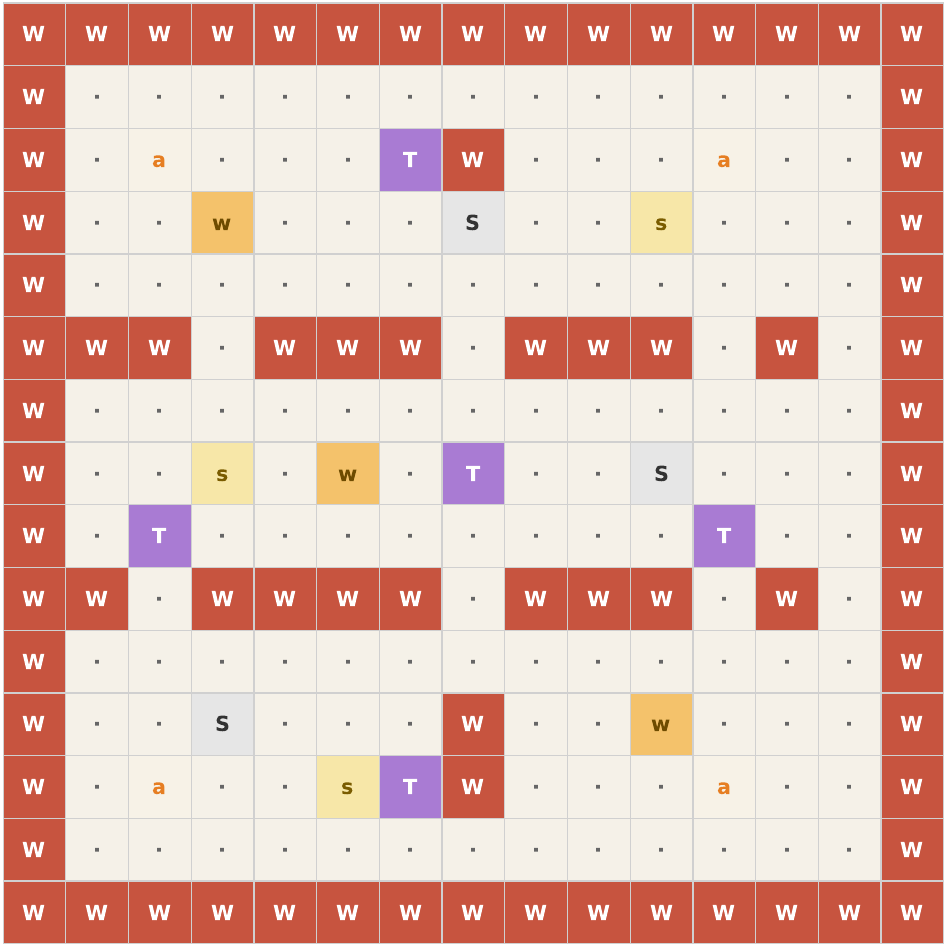}
\end{center}

\textbf{Configuration:}
\begin{itemize}[leftmargin=*,itemsep=1pt,topsep=2pt]
  \item Size: $15\times15$ with 4 agents.
  \item View: $4\times4$, Inventory capacity: 4.
  \item Target quotas: stone 36, sand 36, wood 36.
  \item Legend: \texttt{T}=warehouse, \texttt{W}=wall,\\
  \hspace*{1.5em}\texttt{a}=agent, \texttt{w}/\texttt{s}/\texttt{S}=wood/sand/stone.
  \item Warehouse @ [7,13]: capacity 28;\\
  \hspace*{1.5em}type caps \{wood: 18, sand: 4, stone: 6\}.
\end{itemize}
\end{tcolorbox}

\begin{tcolorbox}[
  colback=gray!5,
  colframe=gray!60!black,
  boxrule=0.8pt,
  arc=1.5mm,
  left=1mm,right=1mm,top=1mm,bottom=1mm,
  title=\textbf{Example 2: Complex Terrain (18$\times$18)},
  fonttitle=\bfseries
]
\small
\RaggedRight
\noindent\textbf{File:} \texttt{18x18\_map/18x18\_v1.json}

\textbf{Map Layout:}\par\vspace{1mm}
\begin{center}
    \includegraphics[width=0.75\linewidth]{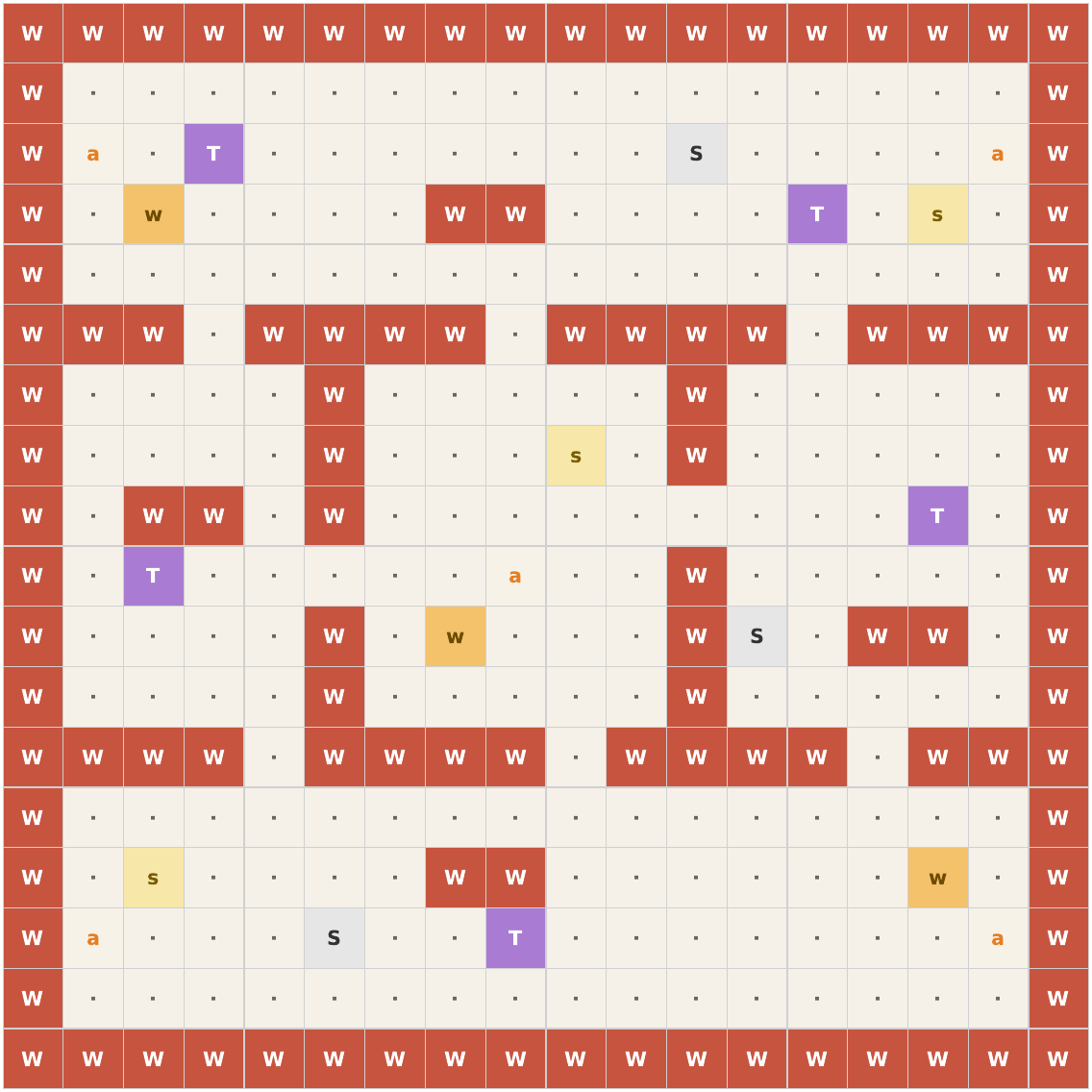}
\end{center}

\textbf{Configuration:}
\begin{itemize}[leftmargin=*,itemsep=1pt,topsep=2pt]
  \item Size: $18\times18$ with 5 agents.
  \item View: $5\times5$, Inventory capacity: 4.
  \item Target quotas: stone 42, sand 42, wood 42.
  \item Legend: \texttt{T}=warehouse, \texttt{W}=wall,\\
  \hspace*{1.5em}\texttt{a}=agent, \texttt{w}/\texttt{s}/\texttt{S}=wood/sand/stone.
  \item Warehouse @ [4,16]: capacity 30;\\
  \hspace*{1.5em}type caps \{wood: 4, sand: 0, stone: 18\}.
\end{itemize}
\end{tcolorbox}

\section{Mechanisms}
\label{app:mechanisms}

\subsection{Reasoning}
\label{app:reasoning_mechanisms}

The Reasoning task uses a two-agent turn-based dialogue workflow under partial observability. We first describe the Base workflow, and then introduce the Protocol.

\subsubsection{Base Workflow}
The Base workflow defines the default interaction process for the Reasoning task and serves as the baseline setting without additional mechanisms. Each instance involves two agents, where each agent receives the question and a private subset of the evidence. Since neither agent has sufficient information to solve the task independently, agents must exchange information through a turn-based dialogue.

\begin{itemize}
    \item \textbf{Observation:} Each agent observes its private context, the question, and the full dialogue history generated so far.  Each turn, the complete utterance of the current speaker is appended to a shared history visible to both participants.
    \item \textbf{Interaction:} Communication is entirely in natural language. Agents may ask targeted questions, share relevant evidence, and provide brief rationales when combining the partitioned evidence, without following a predefined message schema.
    \item \textbf{Termination:} The system monitors the dialogue for specific termination triggers. A valid answer is identified by matching either the \texttt{<A>answer</A>} tags (taking the last occurrence) or the \texttt{\textbackslash boxed\{answer\}} format (taking the first occurrence).
\end{itemize}

\subsubsection{Protocol}

Protocol combines structured public messaging with evidence-seeking instructions and protocol-specific message extraction and termination. It preserves the private evidence partitions, speaker alternation, and task objective, while changing how information is requested, shared, and committed to a final answer.
\begin{itemize}
    \item \textbf{Message Encapsulation:} Agents distinguish their public message from internal reasoning using a designated \texttt{MESSAGE} block. Only the extracted public block is transmitted to the partner.
    \item \textbf{Functional Tagging:} Within the \texttt{<MESSAGE>} block, agents must utilize structured tags to define their communicative intent:
    \begin{itemize}
        \item \texttt{<Q>...</Q>}: To solicit specific missing information from the partner.
        \item \texttt{<R>...</R>}: To provide evidence or responses to previously asked questions.
        \item \texttt{<A>...</A>}: To propose a final answer.
    \end{itemize}
    \item \textbf{Deterministic Termination:} The dialogue terminates immediately upon the detection of an \texttt{<A>} tag within a \texttt{<MESSAGE>} block. If no answer is proposed within the \texttt{max\_rounds} (default 10), the session is marked as a failure.
\end{itemize}

\subsubsection{Prompt Templates}

\noindent\textbf{Base}\par
\noindent

\begin{tcolorbox}[title={Base: Prompt Template}, breakable,  fonttitle=\bfseries\sffamily]
\begin{lstlisting}[breaklines=true, basicstyle=\ttfamily, columns=fullflexible]
You are ${name}, collaborating with ${partner} to answer a multi-hop question using the private information given below.

Question: ${question}
Information (only visible to you):
${information}

GUIDELINES:
- You may ask and answer questions freely in natural language.
- When you are confident you have a complete final answer, output it ONCE as <A>answer</A>.
- The <A>...</A> must contain the full answer needed to solve the question.
- Your private information is incomplete by design; you only see part of the evidence.
- You MUST actively use your partner's information before deciding on a final <A>answer</A>; answering only from your own view is likely to be wrong.
- Always start your reply with "${name}:".
\end{lstlisting}
\end{tcolorbox}

\noindent\textbf{Protocol}\par
\noindent

\begin{tcolorbox}[title={Protocol: Prompt Template}, breakable,  fonttitle=\bfseries\sffamily]
\begin{lstlisting}[breaklines=true, basicstyle=\ttfamily, columns=fullflexible]
You are ${name}, collaborating with ${partner} to answer a multi-hop question using the private information given below.

Question: ${question}

Information (only visible to you):
${information}

TASK:
- Work with ${partner} to find the correct answer.
- Think step by step about what is known and what is still missing.
- Your private information is incomplete by design; you only see part of the evidence.
- You MUST actively use your partner's information before deciding on a final <A>answer</A>; answering only from your own view is likely to be wrong.
- When you are confident, give ONE final answer wrapped as <A>answer</A>.

MESSAGE PROTOCOL (what your partner sees):
- In each turn, include a MESSAGE block that your partner can see.
- Start the MESSAGE block with the line: "${name}: MESSAGE"
- After that line, write 1-4 bullet lines. Each bullet starts with "- " and uses exactly ONE operator:
  * <Q>...</Q>      : ask a concrete question to your partner about missing or uncertain information.
  * <R>...</R>      : respond briefly to a previous <Q> from your partner.
  * <A>...</A>      : propose the complete final answer (ONLY place where <A>...</A> may appear).
- If your partner says something that conflicts with your evidence, use <Q> to ask a targeted question about that point, and then use <R> to give a corrected answer if they ask you.
- Only the MESSAGE block (from "${name}: MESSAGE" down to the bullets) is visible to ${partner} and counts as communication.
- In practice, you can just output the MESSAGE block and nothing else.

STYLE:
- Keep each operator very concise; avoid restating the full question.
- Put any information you want ${partner} to see inside these operators in the MESSAGE block.
\end{lstlisting}
\end{tcolorbox}

\subsection{Scheduling}
\label{app:scheduling_mechanisms}

The Scheduling task models privacy-constrained group negotiation, where each agent only observes its own calendar and must coordinate with others to find a common meeting slot. We describe the Base workflow first, and then introduce Memory and Protocol.

\subsubsection{Base Workflow}
The Base workflow defines the default negotiation process and serves as the baseline setting without additional memory or protocol constraints. Agents negotiate in natural language based on their private schedules and the public conversation history.
\begin{itemize}
    \item \textbf{Observation:} Each agent receives its own private schedule, represented as a list of ``Busy Activities'' (e.g., \textit{14:00-15:00 Meeting}). Agents must infer their free time by identifying gaps between these busy intervals.
    \item \textbf{Interaction:} Communication follows a turn-based broadcast workflow. Agents speak sequentially, and each message is appended to a shared conversation history visible to all participants.
    \item \textbf{Termination:} A designated agent acts solely as a neutral termination detector. It has no privileged access to private schedules or hidden agent states. Its only function is to analyze the public conversation log at the end of each round to determine if a consensus has been reached.
\end{itemize}

\subsubsection{Memory} 
Memory further equips each agent with a precomputed availability whitelist by extracting feasible free intervals from its calendar, and accumulates \texttt{candidate\_slots} and \texttt{rejected\_slots} across rounds to reduce repeated reasoning. 

\begin{itemize}
    \item \textbf{Initialization:} Before the negotiation begins, each agent's calendar is converted into a verified availability list through two steps:
    \begin{enumerate}
        \item \textbf{Gap Extraction:}  identifies all continuous free intervals from the daily schedule.
        \item \textbf{Duration Filter:}  filters these intervals, discarding any slot shorter than the target meeting duration.
    \end{enumerate}
    The resulting availability list is stored in the agent's local memory.
    \item \textbf{Memory Structure:} The memory mechanism consists of two parts. The local memory stores each agent's verified availability list, while a structured \texttt{team\_memory} object tracks the public negotiation state, including \texttt{candidate\_slots} and \texttt{rejected\_slots}.
    \item \textbf{Observation:} In each round, agents are prompted with their local availability memory instead of the raw busy schedule. They are instructed to propose or accept only slots that fit within their verified availability list. The current \texttt{team\_memory} is also provided as public context to avoid repeatedly proposing rejected slots.
    \item \textbf{Stateful Update:} At the end of each round, the public conversation log is parsed to update the \texttt{team\_memory}. Newly proposed slots are added to \texttt{candidate\_slots}, while slots rejected by any agent are moved to \texttt{rejected\_slots}.
\end{itemize}

\subsubsection{Protocol}
Protocol makes the scheduling negotiation more structured by replacing open-ended discussion with a phase-based procedure and typed outputs. This reduces ambiguity in candidate collection and voting, and makes the final consensus easier to verify.

\begin{itemize}
    \item \textbf{Structured Phases:} The negotiation is controlled by a finite-state protocol with three phases:
    \begin{itemize}
        \item \textbf{Phase 0: Broad Scan.} Agents report coarse availability windows, such as \textit{Day 1 Morning}.
        \item \textbf{Phase 1: Propose.} Agents propose concrete meeting slots (e.g., ``10:00-10:20'') within the selected broad windows.
        \item \textbf{Phase 2: Vote.} Agents vote (0/1) on consolidated candidates.
    \end{itemize}
    \item \textbf{JSON-Only communication:} Agents output a phase-specific JSON object rather than free-form natural language. For example, proposals use \texttt{\{"p": [[day, start, end]]\}}, while votes use \texttt{\{"v": [1, 0, ...]\}}.
\end{itemize}

\subsubsection{Prompt Templates}

\noindent\textbf{Base}\par
\noindent

\begingroup
\lstdefinestyle{schedulingprompt}{
  basicstyle=\ttfamily\footnotesize,
  columns=flexible,
  breaklines=true,
  breakatwhitespace=false,
  keepspaces=false,
  showstringspaces=false,
  aboveskip=0pt,
  belowskip=0pt
}
\tcbset{
  schedpromptbox/.style={
    breakable,
    fonttitle=\bfseries\sffamily\small,
    left=0.8mm,
    right=0.8mm,
    top=0.8mm,
    bottom=0.8mm,
    boxsep=0.6mm,
    before skip=2pt,
    after skip=4pt
  }
}

\begin{tcolorbox}[title={Base: Task Initiation}, schedpromptbox]
\begin{lstlisting}[style=schedulingprompt]
Goal: Find one common time slot of exactly ${duration_minutes}
minutes for participants: ${member_names}.

Common unavailable times for all participants:
- 00:00-07:00 (Sleep)
- 07:00-08:00 (Breakfast)
- 12:00-14:00 (Lunch)
- 18:00-20:00 (Dinner)

Constraints:
1. Privacy: individual schedules are private. Participants must
   propose specific slots without revealing full daily agendas.
2. Efficiency: reach final agreement within ${max_rounds} turns.
   Proposals outside Day 1-${days} are INVALID.
\end{lstlisting}
\end{tcolorbox}

\begin{tcolorbox}[title={Base: Member Negotiation}, schedpromptbox]
\begin{lstlisting}[style=schedulingprompt]
You are Agent #${agent_id} ("${agent_name}"), participating
in a collaborative meeting scheduling task.
Your goal is to negotiate with your team to find ONE common
time slot that fits everyone's schedule.

### [SYSTEM STATUS]
Rounds: ${current_round}/${max_rounds}.

### [TASK DIRECTIVE]
${task_prompt}

### [YOUR PRIVATE KNOWLEDGE]
**Your SCHEDULED ACTIVITIES (Busy/Black List):**
${private_schedule}
*(You are BUSY during these times. Implicitly free otherwise.)*
(NOTE: This schedule is PRIVATE. Do not copy-paste it into
the chat. Only you can see this.)

### [TEAM CONTEXT]
Conversation History:
${conversation_history}

### [INSTRUCTIONS]
1. **CONFLICT CHECK (No Overlap):**
   - You see a list of Busy Activities.
   - To accept a slot, ensure it does NOT overlap, even by
     1 minute, with any activity in your schedule.
   - Also ensure it does not violate Sleep/Meal constraints.

2. **INTERACTION:**
   - Check Conversation History for proposed slots.
   - If someone proposed a slot, check your schedule and reply.
   - If there is no active slot, or it was rejected, propose a
     new random slot that works for you.

3. **PRIVACY:**
   - Do NOT reveal your full agenda. Only share specific
     decisions, e.g., "I am free at 10:00".

### [OUTPUT FORMAT]
Output your response in plain text.
If the group has reached a final agreement, strictly follow
the format defined in the Task Directive.
\end{lstlisting}
\end{tcolorbox}

\begin{tcolorbox}[title={Base: Completion Check}, schedpromptbox]
\begin{lstlisting}[style=schedulingprompt]
You are the **Meeting Adjudicator**.
Role: You are the meeting Status Board. Do NOT make suggestions.
ONLY report the status of slots.

### [INPUT DATA]
Total rounds: ${max_rounds}. Current round: ${current_round}.
Prefer decisions that move quickly toward a final slot.

Conversation Log (Current Round):
${turn_log}

Rejected History:
${team_memory}

### [STATUS CLASSIFICATION LOGIC]
1. IDENTIFY REJECTED (Red List)
- Scan the turn log for any "No" or "Reject".
- Identify which slot was rejected, often from the previous
  instruction or a specific new proposal.
- Add these slots to the Rejected List.

2. IDENTIFY ACTIVE/PENDING (Green List)
- Scan the current round for NEW PROPOSALS.
- If a new proposal received a "No" in this round, reject it.
- Otherwise mark it as PENDING. Agents discuss these next.

3. CHECK CONSENSUS
- If any Pending slot received "Yes" from ALL 5 agents, set
  is_complete=true and extract it as the final agreed time.
- The conclusion must specify the slot, e.g.,
  "Final slot agreed: Day 3 10:00-10:20".

### [OUTPUT FORMAT]
Return JSON ONLY.
{
  "is_complete": boolean,
  "conclusion": "Brief status; if final, name the slot.",
  "next_instruction": "REJECTED: [...]\nPENDING: [...]\nACTION: Vote on PENDING slots or propose NEW ones."
}

Constraint for next_instruction:
- You MUST list the PENDING slots found in the current log.
- Example: "REJECTED: Day 5 14:40. PENDING: Day 1 14:00.
  ACTION: Agents, please vote on the Pending slots."
\end{lstlisting}
\end{tcolorbox}

\begin{tcolorbox}[title={Base: Final Slot Extraction}, schedpromptbox]
\begin{lstlisting}[style=schedulingprompt]
You are a strict information extractor for a MEETING scheduling
benchmark.

Input text:
${final_conclusion}

Task:
- Extract the FINAL agreed meeting slot iff there is exactly ONE
  clear slot.
- Slot format: Day <day> <HH:MM>-<HH:MM>.

Rules:
- If no clear final slot, or multiple slots are mentioned,
  return found=false.
- Accept separators such as "-", "--", "~", and optional comma
  after Day number.
- Times must be 24-hour and strictly "HH:MM".
- Meeting duration is exactly ${duration_minutes} minutes.
- If only a start time is present, infer end_time as
  start_time + ${duration_minutes} minutes.

Output JSON only:
{
  "found": boolean,
  "day": integer|null,
  "start_time": string|null,
  "end_time": string|null
}
\end{lstlisting}
\end{tcolorbox}

\noindent\textbf{Memory}\par
\noindent
The Memory setting reuses the Base task initiation, completion check,
and final-slot extraction templates. 

\begin{tcolorbox}[title={Memory: Gap Extraction}, schedpromptbox]
\begin{lstlisting}[style=schedulingprompt]
You are a **Raw Schedule Gap Extractor**.
Task: Identify ALL free time gaps between busy blocks for
Day ${target_day}.

### 1. CONFIGURATION
- Target Day: ${target_day}
- Output Mode: RAW DUMP. Do NOT filter. Do NOT judge duration.
- Continuity: return the FULL continuous gap, e.g.,
  "14:00-16:00", NOT "14:00-14:20".
- Strict Blocking: ANY activity listed in the inputs is 100% BUSY.

### 2. INPUT DATA
Global Blocked Rules:
${global_rules}

Personal Busy Schedule (Day ${target_day}):
${private_schedule_day}

CRITICAL INSTRUCTION:
- Every entry above is a FORBIDDEN ZONE.
- It does NOT matter if the label is "Sleep", "Lunch",
  "Personal", or "Meeting".
- If it is listed, it is BLOCKED. You cannot define a gap
  overlapping with it.

### 3. EXECUTION STEPS
1. Treat Global Rules and Personal Schedule items as solid walls.
2. Merge overlapping busy slots into larger blocks.
   Example: "Lunch 12:00-14:00" + "Meeting 14:00-14:20"
   = Busy 12:00-14:20.
3. Find empty space strictly between Forbidden Zones.
4. Output EVERY gap found, even if it is only 1 minute long.

### 4. OUTPUT FORMAT
Return JSON ONLY.
{
  "day": "${target_day}",
  "all_gaps": [
    "08:50-09:10",
    "15:10-15:30"
  ]
}
\end{lstlisting}
\end{tcolorbox}

\begin{tcolorbox}[title={Memory: Duration Filter}, schedpromptbox]
\begin{lstlisting}[style=schedulingprompt]
You are a **Strict Duration Auditor**.
Task: Filter Candidate Slots and remove ANY slot strictly shorter
than ${duration_minutes} minutes.

### 1. CONFIGURATION
- Minimum Required Duration: ${duration_minutes} minutes.
- Input Data: raw time gaps identified from a schedule.
- Objective: keep valid slots; discard invalid slots.

### 2. INPUT DATA
${candidate_slots}

### 3. VALIDATION LOGIC
For EACH slot:
1. Calculate Duration_Minutes = End_Time - Start_Time.
2. If Duration_Minutes < ${duration_minutes}: REJECT.
3. If Duration_Minutes >= ${duration_minutes}: KEEP.

### 4. OUTPUT FORMAT
Return JSON ONLY.
Populate audit_log first to prove the math was calculated.
{
  "audit_log": [
    "09:00-09:10 (10m) -> REJECT",
    "14:00-16:00 (120m) -> KEEP"
  ],
  "valid_slots": [
    "14:00-16:00"
  ]
}
\end{lstlisting}
\end{tcolorbox}

\begin{tcolorbox}[title={Memory: Round-1 Instruction}, schedpromptbox]
\begin{lstlisting}[style=schedulingprompt]
Meeting negotiation started. The Team Memory is currently empty.
Constraint: Slots must be exactly ${duration_minutes} minutes long.
Strategy: Diversify. Do not focus only on the first day. Propose
slots across different days or different times to avoid crowding.
\end{lstlisting}
\end{tcolorbox}

\begin{tcolorbox}[title={Memory: Member Negotiation}, schedpromptbox]
\begin{lstlisting}[style=schedulingprompt]
You are Agent #${agent_id} ("${agent_name}"), participating in
a collaborative meeting scheduling task.
Your goal is to negotiate with your team to find ONE common time
slot that fits everyone's schedule.

### [SYSTEM STATUS]
Rounds: ${current_round}/${max_rounds}.

### [TASK DIRECTIVE]
${task_prompt}

### [YOUR PRIVATE KNOWLEDGE]
**Your VALID FREE INTERVALS (White List):**
${availability_list}
*(You are ONLY free inside these specific intervals.)*
(NOTE: This schedule is PRIVATE. Do not copy-paste it into
the chat. Only you can see this.)

### [TEAM CONTEXT]
Public Team Memory, for reference only:
${team_memory}

Conversation History:
${conversation_history}

### [INSTRUCTIONS: CHECK & STRATEGIZE]
1. IDENTIFY TARGETS (One-Strike Filter)
- Look at the Leader's PENDING list and any new proposals from
  agents who spoke before you in this round.
- If ANY agent rejected a slot, that slot is DEAD. Do not vote
  Yes on it or propose it again.
- Any proposed slot with ZERO rejections is ALIVE.
- You MUST address every ALIVE slot first.

2. STRICT INTERNAL CHECK (White List)
- Compare each ALIVE slot against your White List.
- The slot must fit ENTIRELY inside one free interval.
- Do not approximate. If a slot ends at 10:05 and your interval
  ends at 10:00, you MUST say No.

3. PRIVACY
- Only share specific decisions. Do not reveal your full agenda.

### [OUTPUT FORMAT]
Output your response in plain text.
If the group has reached a final agreement, strictly follow
the format defined in the Task Directive.
\end{lstlisting}
\end{tcolorbox}

\begin{tcolorbox}[title={Memory: Team Memory Update}, schedpromptbox]
\begin{lstlisting}[style=schedulingprompt]
You are the **Meeting Memory Scribe**.
Goal: Categorize every time slot in the log as Candidate (Alive)
or Rejected (Dead).

### 1. INPUTS
Task:
${task_prompt}

Leader's Summary:
${leader_instruction}

Conversation Log:
${turn_log}

Previous Memory:
${team_memory}

### 2. CLASSIFICATION LOGIC
STEP A: IDENTIFY REJECTIONS
- Scan for keywords: "No", "Busy", "Conflict", "Cannot do".
- If ANY agent rejects a slot, mark it REJECTED.
- If Leader says "Abandon Day 1", all Day 1 slots are REJECTED.

STEP B: IDENTIFY CANDIDATES
- Scan for slots currently being proposed or verified.
- A candidate must be proposed and NOT rejected by anyone yet.
- If Leader says "Verify Day 2 14:00", it is a CANDIDATE.

STEP C: MERGE & CLEAN
- Add new rejections to rejected_slots.
- Remove any slot from candidate_slots if it appears in
  rejected_slots.
- A slot cannot be both. Rejected wins.

### 3. OUTPUT FORMAT
Return JSON ONLY.
{
  "candidate_slots": [ [day, "HH:MM", "HH:MM"] ],
  "rejected_slots": [ [day, "HH:MM", "HH:MM"] ]
}
\end{lstlisting}
\end{tcolorbox}

\noindent\textbf{Memory+Protocol}\par
\noindent
The Memory+Protocol setting reuses the Memory gap extraction and
duration filter.

\begin{tcolorbox}[title={Memory+Protocol: Phase-0 Leader Instruction}, schedpromptbox]
\begin{lstlisting}[style=schedulingprompt]
PHASE: BROAD AVAILABILITY SCAN (Phase 0).
Action: Identify your Top 2-3 free windows from STANDARD BLOCKS:
1. Morning: 08:00-12:00
2. Afternoon: 14:00-18:00
3. Evening: 20:00-24:00
Constraint: Select a block ONLY if you have at least
${duration_minutes} minutes of CONTINUOUS free time inside it.
Output JSON ONLY: {"broad": [ [day, "time"] ]}
CRITICAL: Use block names Morning, Afternoon, or Evening in JSON,
NOT your specific free times.
\end{lstlisting}
\end{tcolorbox}

\begin{tcolorbox}[title={Memory+Protocol: Phase 0 Broad Scan}, schedpromptbox]
\begin{lstlisting}[style=schedulingprompt]
You are Agent #${agent_id} ("${agent_name}").
TASK: BROAD AVAILABILITY SCAN (PHASE 0)

### 1. GOAL
Identify your TOP 3 best Standard Blocks where you have the
MOST TOTAL FREE TIME.

### 2. YOUR TRUTH (WHITE LIST)
You are ONLY free strictly inside these intervals:
${availability_list}

### 3. SCANNING LOGIC
STEP A: DEFINE STANDARD BLOCKS
Evaluate these blocks for each day (Day 1-5):
1. Morning: 08:00-12:00
2. Afternoon: 14:00-18:00
3. Evening: 20:00-24:00

STEP B: CALCULATE SCORE
- Add up the duration of ALL free intervals strictly inside
  each block.
- If max continuous segment < ${duration_minutes} minutes,
  Score = 0.

STEP C: RANK & SELECT
1. Sort blocks by Score, high to low.
2. Pick the Top 3 winners.
3. Map each winner back to its Day ID.

DATA TYPE WARNING:
- Correct first element: Day ID, e.g., 1, 2, 3, 4, 5.
- Wrong first element: calculated minutes, e.g., 590 or 120.
  NEVER output the score.
- Privacy: do not reveal your full availability list.

### 4. OUTPUT FORMAT
Return JSON ONLY.
{
  "broad": [
    [2, "Afternoon"],
    [1, "Morning"],
    [3, "Evening"]
  ]
}
\end{lstlisting}
\end{tcolorbox}

\begin{tcolorbox}[title={Memory+Protocol: Broad Scan Aggregation}, schedpromptbox]
\begin{lstlisting}[style=schedulingprompt]
You are the **Meeting Strategist** (Phase: BROAD SCAN).
Goal: accurately count votes from the log and identify the TOP 2
overlaps.

### 1. INPUTS
Agents' Broad Availability:
${turn_log}
Data format: [Day_Index, "Block_Name"], e.g., [3, "Afternoon"].

### 2. STRICT AGGREGATION LOGIC
1. Normalize [3, "Afternoon"] as key "Day 3 Afternoon".
2. Tally every agent response; count exact frequency per block.
3. Select the highest-count and second-highest-count blocks.
4. If votes tie, choose the earlier Day/Time.

### 3. OUTPUT FORMAT
Return JSON ONLY.
Populate vote_debug with real counts.
{
  "vote_debug": {"Day 1 Morning": 0},
  "is_complete": false,
  "conclusion": "Ranking: [Winner] (N votes), [Runner-up] (M votes).",
  "next_instruction": "PHASE: COLLECT. Strategy and quotas."
}

next_instruction must include:
- Priority #1 [Winner] and Priority #2 [Runner-up].
- Priority #1: propose 2 slots inside this zone.
- Priority #2: propose 1 slot inside this zone.
- Total: 3 slots per agent.
\end{lstlisting}
\end{tcolorbox}

\begin{tcolorbox}[title={Memory+Protocol: Team Memory Update}, schedpromptbox]
\begin{lstlisting}[style=schedulingprompt]
You are the **Protocol State Machine**.
Goal: Update Team Memory JSON from the Leader's latest command.

### 1. INPUTS
Current Memory:
${team_memory}

Leader's Command:
${leader_instruction}

### 2. TRANSITION LOGIC
Check triggers in STRICT ORDER.

CASE A: BROAD STRATEGY SET
Trigger: "FOCUS ON".
1. Extract time blocks from the command, e.g., "Day 1 Morning".
2. Map standard times:
   - Morning -> "08:00", "12:00"
   - Afternoon -> "14:00", "18:00"
   - Evening -> "20:00", "24:00"
3. Save as broad_time_blocks, e.g., [ [1, "08:00", "12:00"] ].
4. Set phase="COLLECT" and candidate_slots=[].

CASE B: START VOTING
Trigger: "PHASE: VOTE".
1. Extract numbered slots into candidate_slots.
2. Keep rejected_slots and broad_time_blocks unchanged.
3. Set phase="VOTE".

CASE C: VOTE FAILED / RESET
Trigger: "PHASE: COLLECT" and no "FOCUS ON".
1. Move all current candidate_slots into rejected_slots.
2. Deduplicate rejected_slots.
3. Clear candidate_slots.
4. Keep broad_time_blocks unchanged.
5. Set phase="COLLECT".

CASE D: DONE
Trigger: "Finalized" or "Success".
Return memory as is and set phase="FINALIZED".

### 3. OUTPUT FORMAT
Return JSON ONLY.
{
  "candidate_slots": [ [day, "HH:MM", "HH:MM"] ],
  "rejected_slots": [ [day, "HH:MM", "HH:MM"] ],
  "broad_time_blocks": [ [day, "HH:MM", "HH:MM"] ],
  "phase": "VOTE" | "COLLECT" | "FINALIZED"
}
\end{lstlisting}
\end{tcolorbox}

\begin{tcolorbox}[title={Memory+Protocol: Phase 1 Propose Slots}, schedpromptbox]
\begin{lstlisting}[style=schedulingprompt]
You are Agent #${agent_id} ("${agent_name}").
TASK: PROPOSE 2-4 ANCHOR SLOTS (PHASE 1)

### 1. CONTEXT
Target Duration: ${duration_minutes} minutes (EXACTLY).
Leader Instruction:
${leader_instruction}

Your Personal Memory (available slots):
${availability_list}

Dead Zones (Rejected):
${team_memory}

Peers' Recent Proposals:
${conversation_history}

### 2. STANDARD TIME DEFINITIONS
1. Morning: 08:00-12:00
2. Afternoon: 14:00-18:00
3. Evening: 20:00-24:00

### 3. EXECUTION ALGORITHM
STEP A: DEFINE SEARCH SCOPE
- Strictly follow Leader Priority Zones, e.g.,
  Day 3 Afternoon = 14:00-18:00.
- If asked for 2 slots in Priority #1, provide 2 distinct slots.

STEP B: GENERATE CANDIDATES
- You must be 100% FREE.
- Slice availability into ${duration_minutes}-minute chunks.

STEP C: GAP ANALYSIS
- Look at Peers' Recent Proposals.
- Categorize existing proposals into hour buckets, e.g.,
  14h, 15h, 16h, 17h.
- Identify hours with ZERO or FEW proposals.
- If no proposals exist, assume all hours are gaps.

STEP D: STRATEGIC SELECTION
- Select slots from gaps identified in Step C.
- If 15:xx is empty, prioritize a slot in 15:xx rather than
  piling onto 14:xx.
- If all hours are covered, try a time between existing slots,
  e.g., if 14:00 and 14:40 are taken, try 14:20.

STEP E: SANITY CHECK
1. Am I free? Yes.
2. Is duration exactly ${duration_minutes}? Yes.
3. Is it inside the requested zone? Yes.
4. Privacy preserved? Yes. Do not reveal your full availability.

### 4. OUTPUT FORMAT
Return JSON ONLY. Exactly 2-4 slots.
{
  "p": [ [day_int, "HH:MM", "HH:MM"] ]
}
\end{lstlisting}
\end{tcolorbox}

\begin{tcolorbox}[title={Memory+Protocol: COLLECT Slots}, schedpromptbox]
\begin{lstlisting}[style=schedulingprompt]
You are the **Meeting Protocol Coordinator** (Phase: COLLECT).
Goal: extract, clean, sort, and list valid slot proposals for voting.

### 1. INPUTS
Raw Proposals:
${turn_log}

Blacklist (Previously Rejected):
${team_memory}

Target Duration:
${duration_minutes} minutes.

### 2. CLEANING RULES
1. Extract all proposed slots [Day, Start, End] from the log.
2. Discard slots where (End - Start) != ${duration_minutes}.
3. Discard any slot that exactly matches the Blacklist.
4. Remove exact duplicates only: same Day + Start + End.
5. KEEP overlaps. 09:00-09:20 and 09:10-09:30 are different.
6. Sort by Day ascending, then Start Time ascending.

### 3. OUTPUT FORMAT
Return JSON ONLY.
Format next_instruction as a clear numbered list for voting.
{
  "is_complete": false,
  "conclusion": "Collected X unique valid proposals (sorted).",
  "next_instruction": "PHASE: VOTE. Numbered candidate list."
}

next_instruction must list candidates as:
0: Day 1 09:00-09:20
1: Day 1 09:10-09:30
\end{lstlisting}
\end{tcolorbox}

\begin{tcolorbox}[title={Memory+Protocol: Phase 2 Vote}, schedpromptbox]
\begin{lstlisting}[style=schedulingprompt]
You are Agent #${agent_id} ("${agent_name}").
TASK: BATCH AVAILABILITY CHECK (PHASE 2)

### 1. INPUT
The Leader needs you to vote (1=Yes, 0=No) on these slots:
${leader_instruction}

### 2. YOUR TRUTH (WHITE LIST)
You are ONLY free strictly inside these intervals:
${availability_list}

### 3. VERIFICATION LOGIC
CRITICAL: Forget Broad Scan preferences. Judge ONLY by the math.

For each slot S (Start-End), find a Free Interval F (Start-End).
Formula: F_Start <= S_Start AND S_End <= F_End.

Trap warnings:
1. Exact start: candidate 09:20-09:40, free 09:20-10:00
   fits, so vote 1.
2. Misalignment: candidate 11:30-11:50, free 11:40-12:00
   starts too early, so vote 0.
3. Partial: candidate 10:00-10:20, free 09:00-10:10
   ends too late, so vote 0.
Privacy: do not reveal your full availability list.

### 4. OUTPUT FORMAT
Return JSON ONLY. Set math_verified=true to confirm that
you calculated the boundaries.
{
  "math_verified": true,
  "v": [1, 0, 1, 0]
}
\end{lstlisting}
\end{tcolorbox}

\begin{tcolorbox}[title={Memory+Protocol: VOTE Check}, schedpromptbox]
\begin{lstlisting}[style=schedulingprompt]
You are the **Meeting Adjudicator** (Phase: VOTE).
Goal: strict verification of UNANIMOUS CONSENSUS (100% agreement).

### 1. INPUTS
The Candidates: use the numbered list in the Leader's last message,
e.g., "0: Day 1 10:00-10:20".

The Votes:
${turn_log}

### 2. STRICT JUDGMENT LOGIC
1. For each candidate index i, sum votes from ALL 5 agents:
   Sum[i] = A1[i] + A2[i] + A3[i] + A4[i] + A5[i].
2. WIN only if Sum[i] == 5.
3. FAIL if Sum[i] < 5. Four votes is a FAIL.
4. If winner index k is found, look back at the candidate list
   and extract the exact slot string.

### 3. OUTPUT FORMAT
Return JSON ONLY.
Always output vote_tally. If a winner exists, conclusion MUST
include the specific slot string.

Case A: Winner Found
{
  "vote_tally": [3, 5, 2, 4],
  "is_complete": true,
  "conclusion": "Winner found at Index X (5/5 votes). Slot: Day <d> <HH:MM>-<HH:MM>",
  "next_instruction": "Meeting Finalized."
}

Case B: No Winner
{
  "vote_tally": [3, 4, 2, 4],
  "is_complete": false,
  "conclusion": "No consensus. Best candidate received max X/5 votes.",
  "next_instruction": "PHASE: COLLECT. Previous candidates rejected. Propose NEW slots."
}
\end{lstlisting}
\end{tcolorbox}
\endgroup

\subsection{Game}
\label{app:game_mechanisms}

The Game task is a cooperative grid-world resource collection task under
partial observability. Agents collect resources and deposit them on
warehouse tiles within a fixed number of rounds. The Base workflow defines
the default environment interaction loop. Memory, Routing, and Protocol are
then added as modular extensions to this workflow. 

\subsubsection{Base Workflow}
\label{sec:game_base_workflow}

The Base workflow defines the default execution loop for the Game task, where agents act solely based on their local observations. At each round, an agent receives a local view centered on itself, together with its global grid position, inventory state, current-tile information, visible terrain, visible resources, visible warehouse tiles, visible teammates, and task-specific fields such as the remaining capacity of the current tile. 
Coordinates follow the implementation convention: \texttt{x} increases from left to right, and \texttt{y} increases from bottom to top.

Each agent generates one environment action per round. The action space includes movement actions \texttt{UP}, \texttt{DOWN}, \texttt{LEFT}, \texttt{RIGHT}, and \texttt{STAY}, as well as interaction actions \texttt{PICK\_UP} and \texttt{PUT\_DOWN}. For resource interactions, the parser also reads optional \texttt{QTY} and \texttt{TYPE} fields. 
An agent can carry only one resource type at a time. The \texttt{PICK\_UP} action is valid only on the agent's current tile and is constrained by the available resource quantity on that tile and the agent's remaining carrying capacity. The \texttt{PUT\_DOWN} action is also executed on the current tile and is constrained by the carried quantity, the remaining tile capacity, and, for warehouse tiles, the remaining per-type warehouse capacity.

The environment resolves all agents' actions synchronously. It first computes movement intentions, blocks moves into impassable terrain, resolves contested target cells and collisions, and then applies pickup/deposit interactions at the agents' resulting positions. Deposited resources contribute to the score only when placed on compatible warehouse tiles, and each resource type is capped by its target quota. 
After all interactions are resolved, the environment updates the score, feedback, and round state. In the Base workflow, communication is inactive: \texttt{messages\_enabled} is set to false, no message phase is executed, and agents do not condition their actions on inter-agent messages.

\subsubsection{Memory}

Memory enables agents to accumulate discovered world knowledge across rounds, helping them reuse previously observed resources, warehouse tiles, and teammate information under partial observability. 

\begin{itemize}
    \item \textbf{Memory structure} 
    Each memory-enabled agent maintains a bounded \texttt{known\_map} as its persistent world state. Each entry records the location, terrain type, resource counts, warehouse capacity, per-type warehouse capacity, timestamp, information source, and staleness metadata. In addition, the agent keeps compact buffers for recent responses, sent and received messages, and a team-state record when structured communication is available.

    \item \textbf{Memory update} 
    After each local observation, visible resources and warehouse-capacity information are merged into \texttt{known\_map}. Directly observed facts have priority over older message-derived facts. When communication is enabled, delivered messages are also parsed into structured facts about resources, warehouse tiles, goals, requests, and teammate states. The implementation combines deterministic parsing for compact message patterns with a strict JSON extractor for free-form resource or warehouse facts.

    \item \textbf{Observation} 
    In each action step, the prompt is augmented with memory-derived hints, including visible resource and warehouse summaries, relevant \texttt{known\_map} entries outside the current view, recent communication facts, and protocol team-state summaries when available. These hints help agents reuse previously discovered information under partial observability.

    \item \textbf{Bounded context.} 
    To keep the prompt compact, memory entries are bounded and pruned by recency and relevance. Only a small number of nearby or task-relevant resource and warehouse hints are injected into the prompt. Memory is used only as advisory context for action selection; the simulator still deterministically checks all movement, pickup, and deposit constraints.
\end{itemize}

\subsubsection{Routing}

Routing enables point-to-point communication among agents, allowing local discoveries to be selectively shared under partial observability. It only controls the communication decision, i.e., whether to send a message and which teammate should receive it, while the environment action is still generated and executed by the Base workflow.

\begin{itemize}
    \item \textbf{Communication} 
    In each communication round, an agent can send at most one message to one teammate. 
    Candidate recipients are teammates currently within communication range, excluding the sender itself. A queued message is addressed to a single recipient and becomes visible to that recipient in the next round if the recipient remains within communication range at delivery time.

    \item \textbf{Random routing} 
    In the Random routing variant, the receiver is sampled from the candidate recipient set using the configured deterministic random source. The sender does not decide whether to send or whom to send to; the LLM only generates the message content for the selected recipient.

    \item \textbf{Gate-LLM routing} 
    In the Gate-LLM routing variant, the routing decision is decomposed into two steps. 
    The agent first decides whether to send a message through a \texttt{SEND} prompt. 
    If sending is enabled, a second \texttt{TO} prompt selects one recipient from the candidate set. The routing step records the selected recipient and compact routing evidence for subsequent message generation, but it does not directly determine the final message body.
\end{itemize}

\subsubsection{Protocol}
Protocol standardizes the content of inter-agent messages so that communication becomes more compact, grounded, and easier to parse. It constrains how a message is formatted after Routing has determined whether and to whom the agent sends a message.

\begin{itemize}
    \item \textbf{Scope} 
    Routing is responsible for the communication decision, including whether to send a message and which teammate receives it. Protocol modifies message construction and validation, not the environment's action space or execution rules. Changes in received messages may nevertheless affect subsequent action choices.

    \item \textbf{Message format} 
    When Protocol is enabled, this free-form payload is replaced by a structured message consisting of three parts:
    \begin{itemize}
        \item \texttt{STATE}: a deterministic line containing the sender's identity, position, carrying state, current goal, and action intent;
        \item \texttt{OBS}: a grounded observation line selected from available evidence, such as resource information, warehouse-tile capacity, carry state, request, or status;
        \item \texttt{MSG}: one short natural-language sentence generated by the LLM for a brief plan, request, or comment.
    \end{itemize}

    \item \textbf{Grounding constraints} 
    Runtime guards are used to prevent unsupported or contradictory message content. 
    In particular, the implementation rejects unsupported coordinates, invented quantities, unsupported capacity claims, and inconsistent warehouse-capacity statements.
\end{itemize}

\subsubsection{Prompt Templates}

\begingroup
\lstdefinestyle{gameprompt}{
  basicstyle=\ttfamily\footnotesize,
  columns=flexible,
  breaklines=true,
  breakatwhitespace=false,
  keepspaces=false,
  showstringspaces=false,
  aboveskip=0pt,
  belowskip=0pt
}
\tcbset{
  gamepromptbox/.style={
    breakable,
    fonttitle=\bfseries\sffamily\small,
    left=0.8mm,
    right=0.8mm,
    top=0.8mm,
    bottom=0.8mm,
    boxsep=0.6mm,
    before skip=2pt,
    after skip=4pt
  }
}

\noindent\textbf{System Prompt}\par
\noindent
\begin{tcolorbox}[title=\textbf{System Prompt}, gamepromptbox]
\begin{lstlisting}[style=gameprompt]
You are an agent. You need to cooperate with other agents and finish a given task.
\end{lstlisting}
\end{tcolorbox}

\noindent\textbf{Common action prompt.}\par
\noindent
This prompt is used by all Game settings. Memory-enabled settings inject
additional memory blocks at \texttt{\$\{memory\_hints\}} and
\texttt{\$\{memory\_sections\}}.

\begin{tcolorbox}[title={Common Action Prompt}, gamepromptbox]
\begin{lstlisting}[style=gameprompt]
You are ${agent_name}, in a sandbox where walls(W) block movement.
Your goal is to complete the task through exploration and collaboration.

### Task
- Collect resources and deposit them on warehouse tiles (T).
- Target items: ${target_items}
- Scoring counts each item type up to its target amount.

### Rules
- Terrain and item layers are separate.
- Items: wood (w), sand (s), stone (S).
- Warehouse tile (T): deposit-only; deposited resources can score.
- You cannot PICK_UP resources from a warehouse tile.
- Passable: empty (.), grass (G), forest (F), warehouse tile (T).
- Impassable: mountain (M), wall (W), river (R).
- Warehouse tiles and other placeable tiles have limited capacity.
- PICK_UP takes resources only from your current tile.
- PUT_DOWN places resources only on your current tile.
- To score, first move onto a compatible warehouse tile, then PUT_DOWN.
- Warehouse deposits are limited by carried quantity, free capacity,
  and remaining warehouse per-type capacity.
- Carry at most one resource type at a time.
- Coordinates: x increases left-to-right; y increases bottom-to-top.
- Current view: Y is you; a is another visible agent.
- Messaging is handled in a separate message phase and arrives next round.

Round: ${round}

${response_memory}

${vision_block}

Your current observation:
${observation}

Last action feedback:
${last_feedback}

Received messages:
${messages}

Recent history:
${history}

${memory_hints}
${memory_sections}

### Output Requirement
Give a brief analysis and plan, then strictly output:
ACTION: [UP|DOWN|LEFT|RIGHT|STAY|PICK_UP|PUT_DOWN]
QTY: [integer]
TYPE: [wood|sand|stone]

For movement or STAY, omit QTY and TYPE.
For PICK_UP, include QTY and TYPE when multiple resource types are present.
For PUT_DOWN, include QTY.
\end{lstlisting}
\end{tcolorbox}

\noindent\textbf{Memory prompt}\par
\noindent
Memory reuses the common action prompt and adds structured view summaries,
known-map facts, and message-derived hints.

\begin{tcolorbox}[title={Memory: Action-Prompt Injection}, gamepromptbox]
\begin{lstlisting}[style=gameprompt]
### Memory Hints
- Use memory (resources and warehouse tiles) to augment limited view.
- Coordinate using recent received messages when available.
- If current view has little actionable information, use memory to
  choose directions, resources, or warehouse-tile targets.
- Always integrate the current grid with memory before deciding.

Visible Resources Summary:
${visible_resources}

Visible Warehouse-Tile Summary:
${visible_warehouse_tiles}

Memory - resources and warehouse tiles (not in current view):
- warehouse@(${x}, ${y}): remaining=${capacity};
  per_type=${per_type_capacity}; seen=${seen_round}; src=${source}
- resource@(${x}, ${y}): items=${items};
  seen=${seen_round}; src=${source}

Recent communication facts:
${team_memory}
\end{lstlisting}
\end{tcolorbox}

\begin{tcolorbox}[title={Memory: Message-to-KB Extractor}, gamepromptbox]
\begin{lstlisting}[style=gameprompt]
You are an information extraction assistant.
Given an agent message about warehouse tiles or resources,
extract facts into STRICT JSON.
Output exactly one JSON object with key "entries" (array). No commentary.
Each entry:
- x (int), y (int)
- terrain ("warehouse" or "resource")
- optional items (object: type -> int)
- optional remaining_capacity (int)

If no facts can be extracted, output {"entries": []}.

Message:
${raw_message}
\end{lstlisting}
\end{tcolorbox}

\noindent\textbf{Routing prompts}\par
\noindent
Routing reuses the common action prompt for environment actions and adds a
separate message phase. Random routing selects the recipient in the mechanism;
Gate routing asks the agent for \texttt{SEND}, \texttt{TO}, and \texttt{MSG}.

\begin{tcolorbox}[title={Routing: Gate Send Prompt}, gamepromptbox]
\begin{lstlisting}[style=gameprompt]
You are ${self_name}. This is MESSAGE PHASE for round ${round}.
Single-message-per-round mode is enabled.
Your position: (${x},${y}).

Candidates in communication range:
${candidate_list}

Payload check:
- Identify one concrete payload before deciding SEND.
- Useful payloads answer or ask a concrete need, share a compatible
  warehouse tile, or share a fresh resource or warehouse-tile fact.
- Avoid repeated facts, generic status, and movement-only updates.

Output EXACTLY one line:
SEND: yes|no
\end{lstlisting}
\end{tcolorbox}

\begin{tcolorbox}[title={Routing: Gate Recipient Prompt}, gamepromptbox]
\begin{lstlisting}[style=gameprompt]
You are ${self_name}. This is MESSAGE PHASE for round ${round}.
Single-message-per-round mode is enabled.
Your position: (${x},${y}).

Candidates in communication range:
${candidate_list}

A message will be sent this round.
Choose the teammate most likely to use or answer the payload.
For fresh map coverage, prefer someone less recently contacted
or likely missing that fact.

Output EXACTLY one line:
TO: <one candidate name>
\end{lstlisting}
\end{tcolorbox}

\begin{tcolorbox}[title=\textbf{Routing: Gate Message Prompt}, gamepromptbox]
\begin{lstlisting}[style=gameprompt]
You are {SELF_NAME}. This is MESSAGE PHASE for round {ROUND_NUM}.
Your position: ({X},{Y}). Target: {TARGET_NAME}.
Target position: ({TX},{TY}) | In view: {IN_VIEW} | Carrying: {CARRYING_OR_UNKNOWN}
Task: Collaborate with others to achieve the goal.

Selected teammate context:
- selected teammate fact: {RESOURCE_OR_WAREHOUSE_FACT}
- current load: {TYPE}:{QTY}

Output EXACTLY one line:
MSG: <your message text>
\end{lstlisting}
\end{tcolorbox}

\begin{tcolorbox}[title={Routing: Random Message Prompt}, gamepromptbox]
\begin{lstlisting}[style=gameprompt]
You are ${self_name}. This is MESSAGE PHASE for round ${round}.
Your position: (${x},${y}). Target: ${target_name}.
Target hints:
in_view=${in_view}; asked_last_round=${asked_last_round};
recently_contacted=${recently_contacted}

Task: Collaborate with others to achieve the goal.
The recipient was selected from communication-range candidates.

Output EXACTLY one line:
MSG: <your message text>
\end{lstlisting}
\end{tcolorbox}

\noindent\textbf{Protocol prompts}\par
\noindent
Protocol reuses the selected recipient from Routing. It replaces or
constrains the message content; repeated Routing templates are not duplicated.

\begin{tcolorbox}[title={Protocol: Message Prompt}, gamepromptbox]
\begin{lstlisting}[style=gameprompt]
You are ${self_name}. Send one useful message to ${recipient}.
Protocol will attach STATE/OBS from evidence that supports your message.

Output EXACTLY two lines:
LANE: <RESOURCE|WAREHOUSE|CARRY|REQUEST|STATUS>
MSG: <one short grounded message under ${max_chars} characters>

Deterministic STATE header:
STATE: id=${self_name}; loc=(${x},${y});
  carry=${carry}; goal=${goal}; action=${action}

Recipient context:
${recipient_context}

Routing brief:
${routing_context}

Content candidates:
${content_candidates}

Evidence:
${evidence_facts}

Guidance:
- Choose one compact update from Content candidates.
- If sharing RESOURCE or WAREHOUSE, include type and coordinate.
- For warehouse-tile facts, mention capacity or resource compatibility.
- Do not invent coordinates, quantities, capacities, or carried items.
- Do not add explanation outside the two output lines.
\end{lstlisting}
\end{tcolorbox}

\begin{tcolorbox}[title={Protocol: Wire Format}, gamepromptbox]
\begin{lstlisting}[style=gameprompt]
STATE: id=${self_name}; loc=(${x},${y});
  carry=${carry}; goal=${goal}; action=${action}
OBS: ${grounded_fact}
MSG: ${short_note}
\end{lstlisting}
\end{tcolorbox}
\endgroup

\end{document}